\documentclass[conference]{IEEEtran}
\IEEEoverridecommandlockouts
\usepackage{amsmath,amssymb,amsfonts}
\usepackage{algorithmic}
\usepackage{graphicx}
\usepackage{textcomp}
\usepackage{xcolor}
\usepackage[utf8]{inputenc}
\usepackage{url}  
\usepackage{pifont}
\usepackage[sort]{natbib} 
\usepackage{hyperref}
\usepackage{multirow}
\usepackage{booktabs}
\usepackage{arydshln}
\usepackage{pdfpages}
\usepackage[compatibility=false]{caption}
\usepackage{array}
\usepackage{adjustbox}
\usepackage{makecell}
\usepackage{subcaption}

\def\BibTeX{{\rm B\kern-.05em{\sc i\kern-.025em b}\kern-.08em
    T\kern-.1667em\lower.7ex\hbox{E}\kern-.125emX}}
\begin{document}

% \nocite{*}

\title{Motion Generation: A Survey of Generative Approaches and Benchmarks}

\makeatletter
\newcommand{\linebreakand}{%
  \end{@IEEEauthorhalign}
  \hfill\mbox{}\par
  \mbox{}\hfill\begin{@IEEEauthorhalign}
}
\makeatother

\author{
    \IEEEauthorblockN{Aliasghar Khani}
    \IEEEauthorblockA{
        \textit{Autodesk Research} \\
        aliasghar.khani@autodesk.com
    }
    \and
    \IEEEauthorblockN{Arianna Rampini}
    \IEEEauthorblockA{
        \textit{Autodesk Research}
        % arianna.rampini@autodesk.com
    }
    \and
    \IEEEauthorblockN{Bruno Roy}
    \IEEEauthorblockA{
        \textit{Autodesk Research}
        % bruno.roy@autodesk.com
    }
    \and
    \IEEEauthorblockN{Larasika Nadela}
    \IEEEauthorblockA{
        \textit{Autodesk Research}
        % bruno.roy@autodesk.com
    }
    \linebreakand 
    \IEEEauthorblockN{Noa Kaplan}
    \IEEEauthorblockA{
        \textit{Autodesk Research}
        % noa.kaplan@autodesk.com
    }
    \and
    \IEEEauthorblockN{Evan Atherton}
    \IEEEauthorblockA{
        \textit{Autodesk Research}
        % bruno.roy@autodesk.com
    }
    \and
    \IEEEauthorblockN{Derek Cheung}
    \IEEEauthorblockA{
        \textit{Autodesk Research}
        % bruno.roy@autodesk.com
    }
    \and
    \IEEEauthorblockN{Jacky Bibliowicz}
    \IEEEauthorblockA{
        \textit{Autodesk Research}
        % bruno.roy@autodesk.com
    }
}

\maketitle

\begin{abstract}
Motion generation, the task of synthesizing realistic motion sequences from various conditioning inputs, has become a central problem in computer vision, computer graphics, and robotics, with applications ranging from animation and virtual agents to human-robot interaction. As the field has rapidly progressed with the introduction of diverse modeling paradigms including GANs, autoencoders, autoregressive models, and diffusion-based techniques, each approach brings its own advantages and limitations. This growing diversity has created a need for a comprehensive and structured review that specifically examines recent developments from the perspective of the generative approach employed. 

In this survey, we provide an in-depth categorization of motion generation methods based on their underlying generative strategies. Our main focus is on papers published in top-tier venues since 2023, reflecting the most recent advancements in the field. In addition, we analyze architectural principles, conditioning mechanisms, and generation settings, and compile a detailed overview of the evaluation metrics and datasets used across the literature. Our objective is to enable clearer comparisons and identify open challenges, thereby offering a timely and foundational reference for researchers and practitioners navigating the rapidly evolving landscape of motion generation.

\end{abstract}


\section{\textbf{Introduction}}
Motion generation refers to the task of synthesizing realistic and coherent sequences of human or character motion, typically represented as 3D joint trajectories or poses, conditioned on various inputs such as action labels, textual prompts, or environmental cues~\citep{tevethuman, khani2025unimogen, guo2020action2motion}. This capability plays a fundamental role in computer vision, computer graphics, robotics, and human-computer interaction, with applications ranging from virtual avatar animation and video games to human-robot collaboration and embodied agents~\citep{li2018convolutional, petrovich2021action, jiang2023motiongpt}. Effective motion generation automates traditionally manual and time-consuming animation workflows, while also supporting behavioral modeling, simulation, and data augmentation.

Given the rapid growth and diversification of research in motion generation, spanning modeling paradigms such as GANs, autoencoders, autoregressive models, and diffusion-based approaches, there is a growing need for a comprehensive and structured survey to organize and clarify this rapidly evolving field. In addition to reviewing modeling approaches, compiling the evaluation metrics and datasets used in recent works makes it easier to compare methods, assess progress, identify evaluation gaps, and guide the development of standardized benchmarks for future research.

While prior surveys have primarily focused on broad topics like general human motion generation~\citep{zhu2023human, lyu20223d, ye2022human}, or more specialized areas such as human interaction motion generation~\citep{sui2025survey} and text-driven motion generation~\citep{sahili2025text}, they are either old or lack a detailed and systematic categorization of methods based on the underlying generation approach. However, the choice of generation approach significantly affects various aspects of a method, including motion quality, inference speed, and computational requirements. Therefore, a systematic review that emphasizes generation strategies is critical for understanding and comparing methods along this dimension. Such a survey not only helps researchers identify trends and gaps in the literature but also supports reproducibility and informs the development of future benchmarks and datasets. Ultimately, we aim to create a reference that helps both knowledgeable readers and those new to the topic navigate the field of motion generation.

In this survey, we focus specifically on the generation approaches employed in recent motion generation methods. Our primary emphasis is on works published in top-tier conferences since 2023, ensuring coverage of the latest advancements in the field. We categorize and analyze a wide range of models such as autoregressive models, variational autoencoders, generative adversarial networks, and diffusion-based techniques, highlighting their architectural designs, conditioning mechanisms, and application contexts. Additionally, we provide a comprehensive overview of the evaluation metrics and datasets used across the surveyed works. This enables a clearer understanding of how different models are assessed, facilitates fair comparisons, and helps identify emerging trends and existing gaps in current evaluation practices.

\textbf{Paper Structure.} The remainder of this survey is organized as follows. Section~\ref{sec:background} provides background on motion data and a high-level overview of motion generation paradigms. Section~\ref{sec:methods} presents a detailed categorization and analysis of motion generation methods based on their underlying generation approaches. Section~\ref{sec:datasets} summarizes the commonly used datasets in recent literature. Section~\ref{sec:evaluation_metrics} reviews the evaluation protocols and metrics employed across studies. Section~\ref{sec:statistical_insights} provides statistical insights into publication trends, the distribution of methods across different categories, dataset usage frequency, and the most common conditioning signals, offering a quantitative perspective on the current research landscape. Finally, Section~\ref{sec:conclusion} discusses current challenges and future research directions.

\textbf{Contributions.} The main contributions of this survey are as follows:
% \begin{itemize}
%     \item We provide a focused categorization of recent human motion generation methods based on their underlying generation approaches, including feed-forward based models, autoencoders, variational autoencoders, vector quantized variational autoencoders, continuous autoregressive models, discrete autoregressive models, generative adversarial networks, diffusion-based models, latent diffusion-based models, flow matching-based models, and others such as physics-based and reinforcement-based models.
    
%     \item We analyze each category in terms of model architecture, conditioning mechanisms, and generation settings, offering insights into the design choices and capabilities of different approaches.
    
%     \item We compile a comprehensive list of datasets and evaluation metrics used in the literature, facilitating standardized comparison and reproducibility across studies.
    
%     \item We identify current trends, limitations, and open challenges in the field, highlighting directions for future research in motion generation.

%     \item We present statistical insights into the surveyed papers, including publication trends, distribution of methods by category, dataset usage frequencies, and the prevalence of different conditioning signals, offering a quantitative view of the current research landscape.
    
%     \item We aim to support both newcomers and experienced researchers by providing a structured and up-to-date reference for navigating the rapidly evolving landscape of motion generation research.
% \end{itemize}

\begin{itemize}
    \item We provide a focused categorization of recent motion generation methods based on their underlying generation approaches, including feed-forward-based models, autoencoders, variational autoencoders, vector quantized variational autoencoders, continuous autoregressive models, discrete autoregressive models, generative adversarial networks, diffusion-based models, latent diffusion-based models, flow matching-based models, and others such as physics-based and reinforcement-based models.
    
    \item We analyze each category in terms of model architecture, conditioning mechanisms, and generation settings, offering insights into the design choices and capabilities of different approaches.
    
    \item We compile a comprehensive list of datasets and evaluation metrics used in the literature, facilitating standardized comparison and reproducibility across studies.
    
    % \item We identify current \textbf{trends} limitations, and **open challenges** in the field, highlighting directions for future research in motion generation.
    
    \item We present statistical insights into the surveyed papers, including publication trends, distribution of methods by category, dataset usage frequencies, and the prevalence of different conditioning signals, offering a quantitative view of the current research landscape.
    
    \item We aim to support both newcomers and experienced researchers by providing a structured and up-to-date reference for navigating the rapidly evolving landscape of motion generation research.
\end{itemize}

% \begin{itemize}
%     \item Definition and importance of motion generation (AI)
%     \item Importance of having a survey paper for motion generation (Us)
%     \item although there is a previous survey, there are a lot of papers since 2023 and there is a need to systematic review of these papers to see the trends, advancements. (US)
%     \item what is our focus in this survey? approaches used in the methods. We also provide a comprehensive list of metrics and datasets used in these papers (US)
%     % \item In our review we noticed some key challenges which is common in papers (e.g., physical plausibility like foot sliding, foot penetration, limited to skeleton and human data~\citep{tevethuman, zhao2024dart, li2024lamp, guo2023back, ahuja2019language2pose, petrovich2021action, wan2024tlcontrol, zhang2018mode, ahn2018text2action, zhou2023ude, cervantes2022implicit}, lack of study of specific problems of motion) (US)
%     \item Contributions of the survey (AI)
% \end{itemize}
\section{\textbf{Background}}
\label{sec:background}
\subsection{\textbf{Motion Data}}

Human motion data captures the dynamic configuration of the human body over time and serves as the foundation for learning-based motion generation systems. The structure and quality of this data critically influence the design, performance, and generalization ability of generative models. Motion data can be categorized along two major axes: how the motion is represented (\textit{Motion Data Representation}) and how it is acquired or constructed (\textit{Motion Data Collection}). Understanding these dimensions is essential for selecting appropriate modeling techniques and for benchmarking across datasets and tasks.

\subsubsection{\textbf{Motion Data Representation}}

Human motion can be represented in several formats, each offering a different trade-off between accuracy, efficiency, physical realism, and modeling complexity. The choice of representation directly influences the design of generative models, the selection of loss functions, and the fidelity of synthesized motion. Common representations include 2D/3D keypoint, rotation, and mesh formats, as illustrated in Fig.~\ref{fig:data_representation}.

\begin{figure}[t]
  \centering
  \begin{subfigure}[t]{0.24\textwidth}
    \includegraphics[trim=20pt 20pt 20pt 20pt, clip, width=\linewidth]{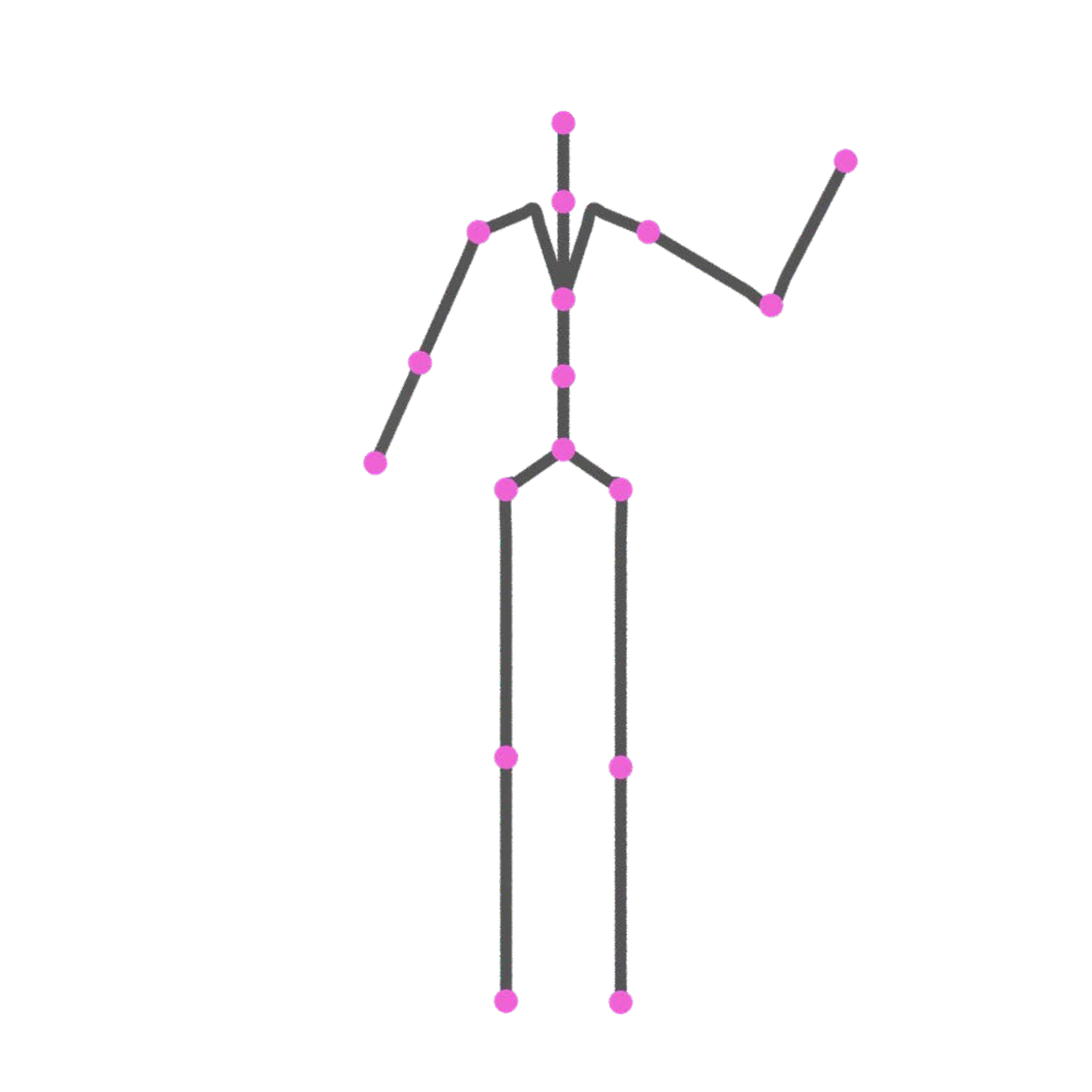}
    \caption*{(a)}
  \end{subfigure}
  \begin{subfigure}[t]{0.24\textwidth}
    \includegraphics[trim=20pt 20pt 20pt 20pt, clip, width=\linewidth]{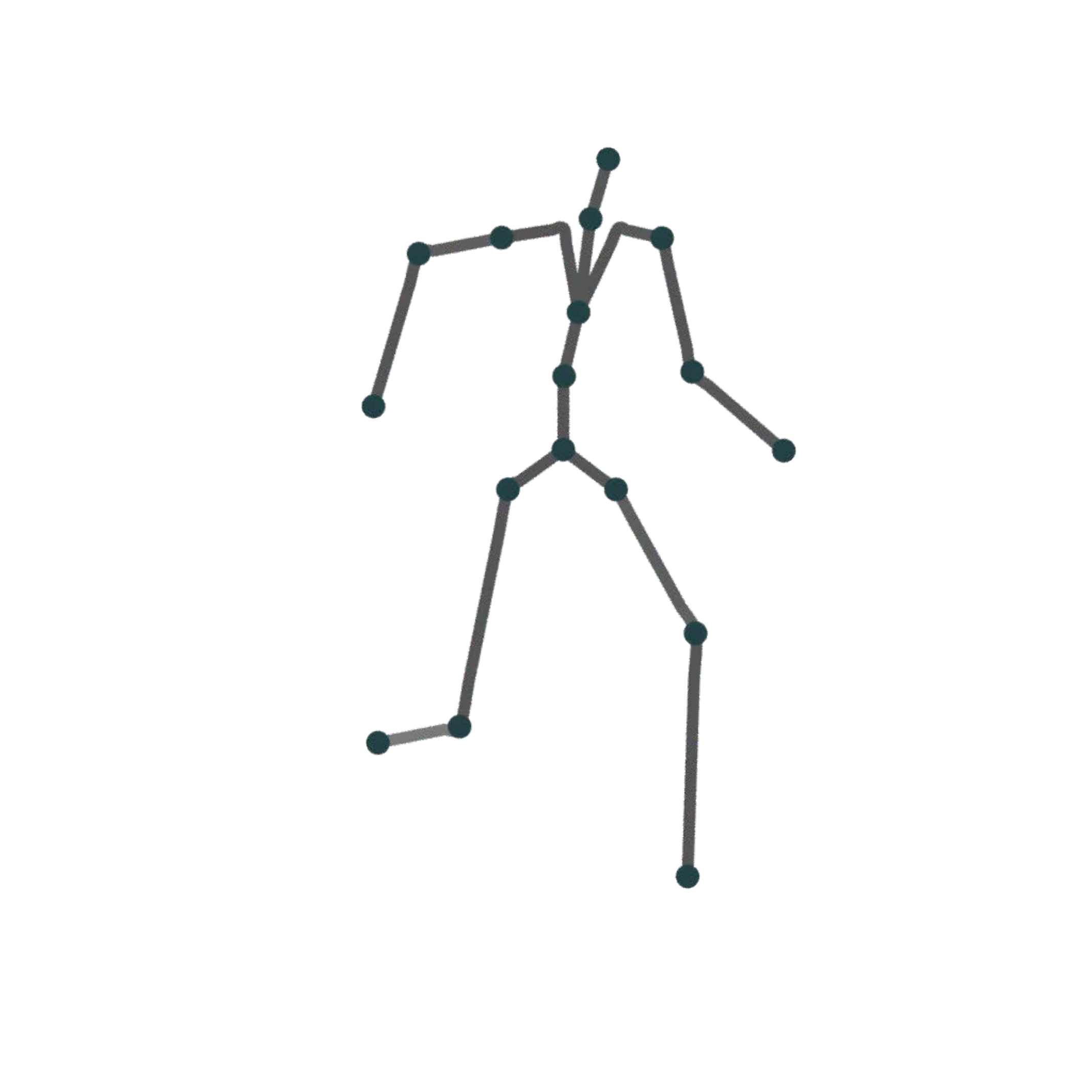}
    \caption*{(b)}
  \end{subfigure}
  \begin{subfigure}[t]{0.24\textwidth}
    \includegraphics[trim=20pt 20pt 20pt 20pt, clip, width=\linewidth]{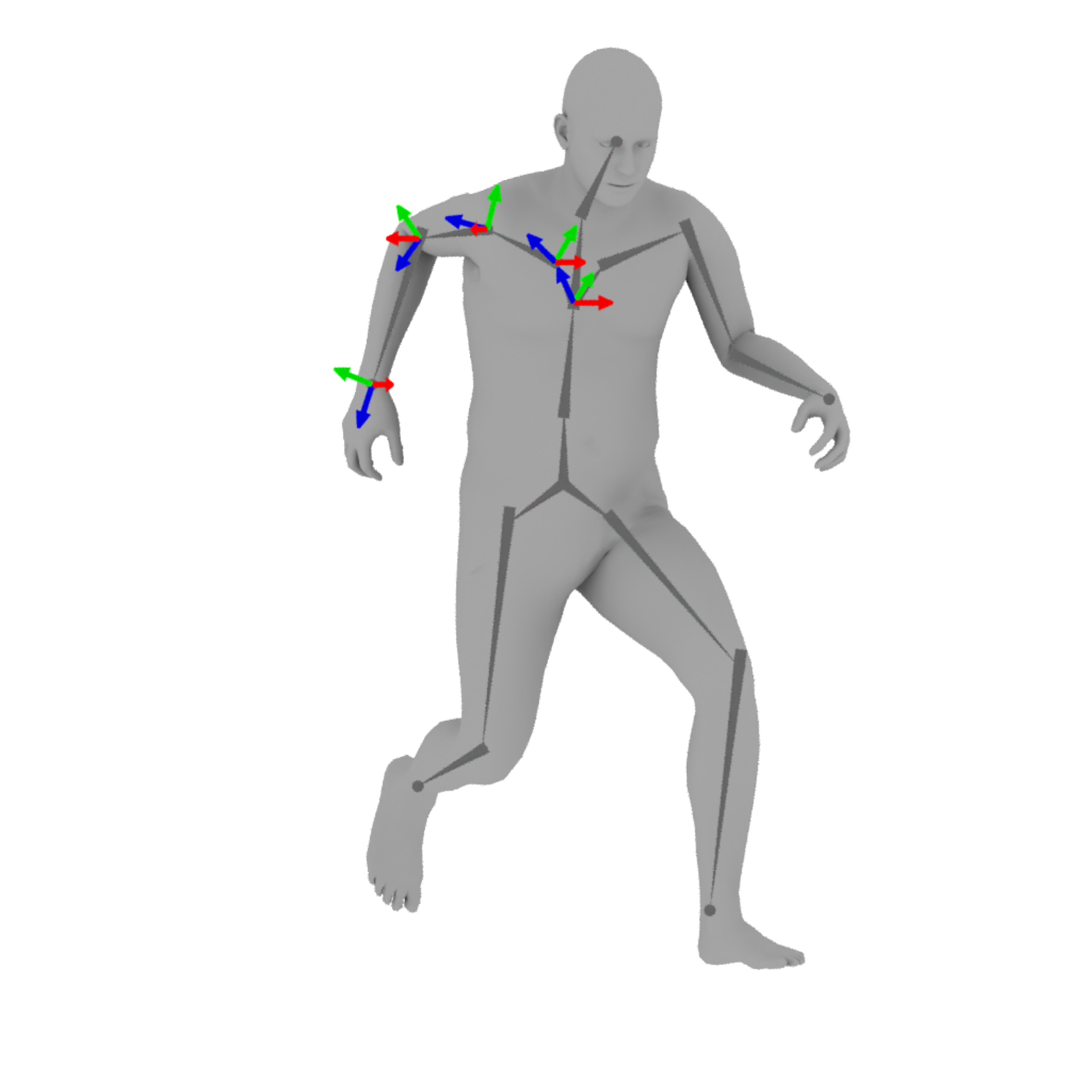}
    \caption*{(c)}
  \end{subfigure}
  \begin{subfigure}[t]{0.24\textwidth}
    \includegraphics[trim=20pt 20pt 20pt 20pt, clip, width=\linewidth]{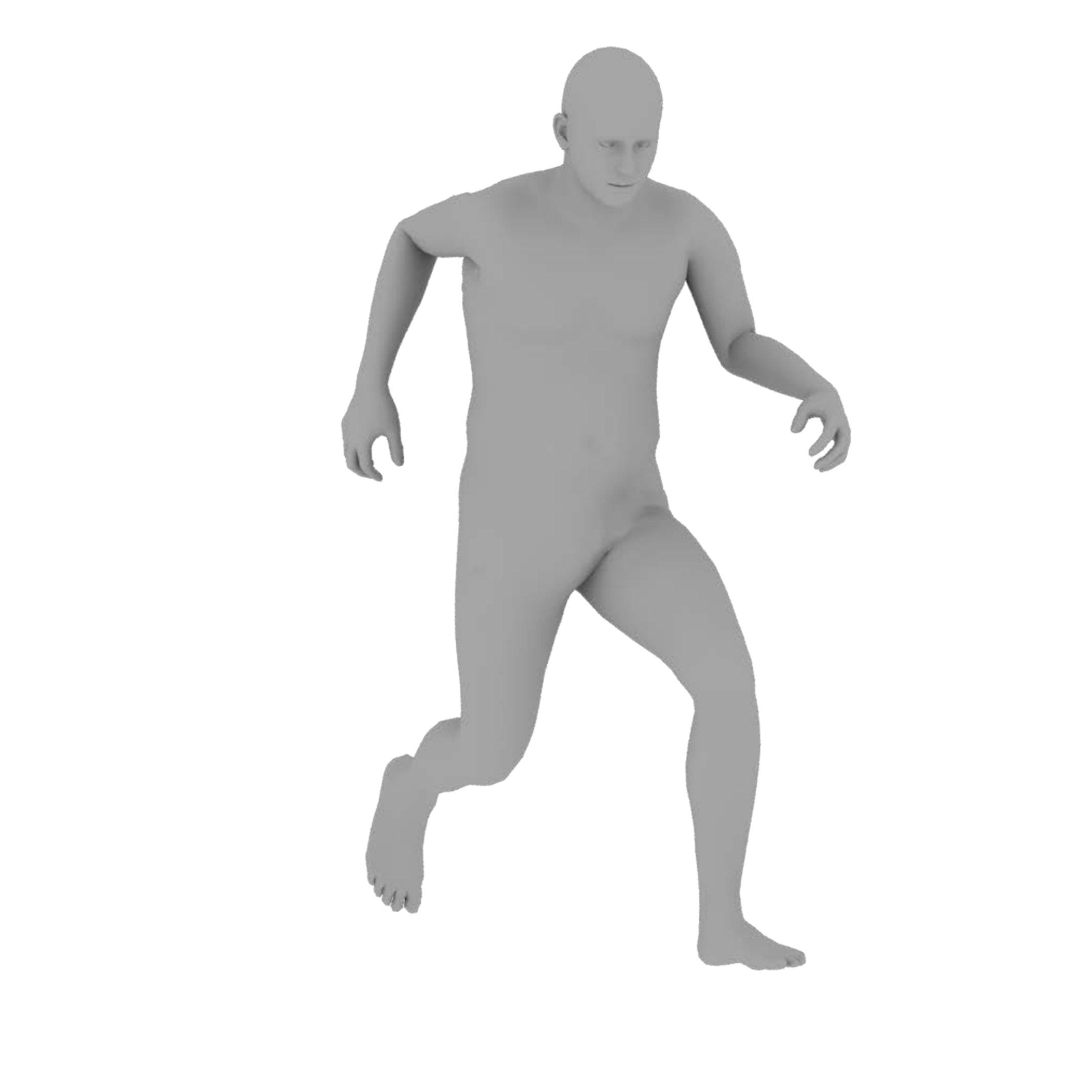}
    \caption*{(d)}
  \end{subfigure}
  \caption{Examples of different data representations: (a) 2D keypoint representation, (b) 3D keypoint representation, (c) rotation representation, and (d) mesh representation}
  \label{fig:data_representation}
\end{figure}

\paragraph{\textbf{Keypoint Representation}}

Keypoint representations describe motion as a sequence of 2D or 3D coordinates corresponding to human body joints. Each frame is represented as:
\[
\mathbf{P}_t = [\mathbf{p}_{t,1}, \mathbf{p}_{t,2}, \dots, \mathbf{p}_{t,J}] \in \mathbb{R}^{J \times d},
\]
where \( J \) is the number of joints and \( d \in \{2, 3\} \) indicates spatial dimensionality. These joint positions can be absolute or defined relative to the root joint or the first frame.

This representation is intuitive and widely used in pose estimation tasks. It is straightforward to visualize and simple to predict using regression models. However, it lacks explicit encoding of skeletal constraints, and care must be taken to ensure plausible kinematic structure during generation or inference.

\paragraph{\textbf{Rotation Representation}}

Rotation representations encode the orientation of each joint relative to its parent in a kinematic hierarchy. A frame is represented as:
\[
\mathbf{R}_t = [\mathbf{R}_{t,1}, \mathbf{R}_{t,2}, \dots, \mathbf{R}_{t,J}], \quad \mathbf{R}_{t,j} \in \text{SO}(3),
\]
where \( \text{SO}(3) \) denotes the space of 3D rotation matrices. Alternative representations include axis-angle, Euler angles, or quaternions.

Rotation representations maintain bone lengths and allow the reconstruction of joint positions via forward kinematics. They are especially suited to animation, simulation, and physically based modeling. Their non-Euclidean nature, however, introduces challenges in network design and loss computation.

\paragraph{\textbf{Mesh Representation}}

Mesh representations model the full 3D surface geometry of the human body. A prominent example is the SMPL (Skinned Multi-Person Linear) model~\citep{loper2015smpl}, which defines a deformable mesh with \( N = 6890 \) vertices through a learned function:
\[
M(\boldsymbol{\beta}, \boldsymbol{\theta}) \in \mathbb{R}^{N \times 3},
\]
where \( \boldsymbol{\beta} \in \mathbb{R}^{10} \) are shape parameters and \( \boldsymbol{\theta} \in \mathbb{R}^{3K} \) are pose parameters for \( K = 24 \) joints, typically in axis-angle format. The model uses linear blend skinning (LBS) to deform the mesh representation on joint rotations and body shape.

SMPL enables accurate and compact representation of both pose and shape, supporting downstream tasks such as rendering, contact reasoning, and motion analysis. Datasets like AMASS~\citep{mahmood2019amass} standardize motion sequences using SMPL, facilitating cross-dataset learning and benchmarking.

\medskip

These three representation types, keypoints, rotations, and meshes, are often used interchangeably or jointly depending on the target application. For example, models may predict joint rotations and use forward kinematics to derive keypoints, or generate SMPL parameters from estimated poses. Understanding the advantages and limitations of each format is essential for effective model design and evaluation.

\subsubsection{\textbf{Motion Data Collection}}

The quality and diversity of motion datasets significantly influence the effectiveness of generative models. Depending on the application, motion data may be collected through physical capture systems, estimated from images or videos, or even manually created. The methods of motion data collection vary in terms of accuracy, scalability, and realism. Below are four common approaches used in existing datasets.

\paragraph{\textbf{Marker-based Capture}}

Marker-based motion capture (MoCap) is a high-precision method where human actors wear suits embedded with reflective markers placed at specific anatomical locations. An array of synchronized infrared cameras tracks the 3D positions of these markers in space. The captured marker trajectories are then mapped to a skeletal model using inverse kinematics. This approach yields highly accurate joint trajectories and is widely used in professional animation, biomechanics, and robotics. However, it is expensive, requires a controlled environment, and limits the diversity of captured scenarios.

\paragraph{\textbf{Markerless Capture}}

Markerless motion capture removes the need for physical markers by relying on multiple RGB cameras arranged around the subject. Using multi-view geometry and computer vision algorithms, it reconstructs 3D human poses from synchronized image sequences. Techniques such as 3D keypoint triangulation, shape-from-silhouette, and volumetric body reconstruction are commonly employed. While less accurate than marker-based systems, markerless setups are more flexible and scalable, enabling motion capture in more natural and unconstrained environments.

\paragraph{\textbf{Pseudo-labeled Data}}

Pseudo-labeling methods leverage machine learning models to estimate motion data from raw videos or images, often from in-the-wild sources like YouTube or movie clips. Pre-trained pose estimation models (e.g., OpenPose, VideoPose3D, VIBE) predict 2D or 3D joint positions or SMPL parameters frame by frame. While pseudo-labeled datasets may contain noise and inaccuracies, they offer large-scale, diverse motion data with minimal manual effort. They are especially valuable for training models to generalize to unconstrained scenarios.

\paragraph{\textbf{Human Design}}

Manually authored motion data is created by human animators using digital content creation tools such as Blender, Maya, or MotionBuilder. These tools allow precise control over body pose, timing, and style, making them ideal for producing stylized or task-specific motions that are hard to capture physically. Although this approach is labor-intensive and less scalable, it enables the creation of high-quality and semantically rich motion sequences that may not occur naturally, such as exaggerated cartoon-like actions or fantastical character movements.

\subsection{\textbf{Motion Generation Methods}}
\label{motion_generation_methods_background}
This section provides a concise overview of key methodological categories for motion generation found in the literature. An overview of the approaches discussed in this section is illustrated in Fig.~\ref{fig:approaches}.

\begin{figure*}[t]
    \centering
    \includegraphics[width=\textwidth, trim=1.3cm 0cm 1.3cm 0cm, clip]{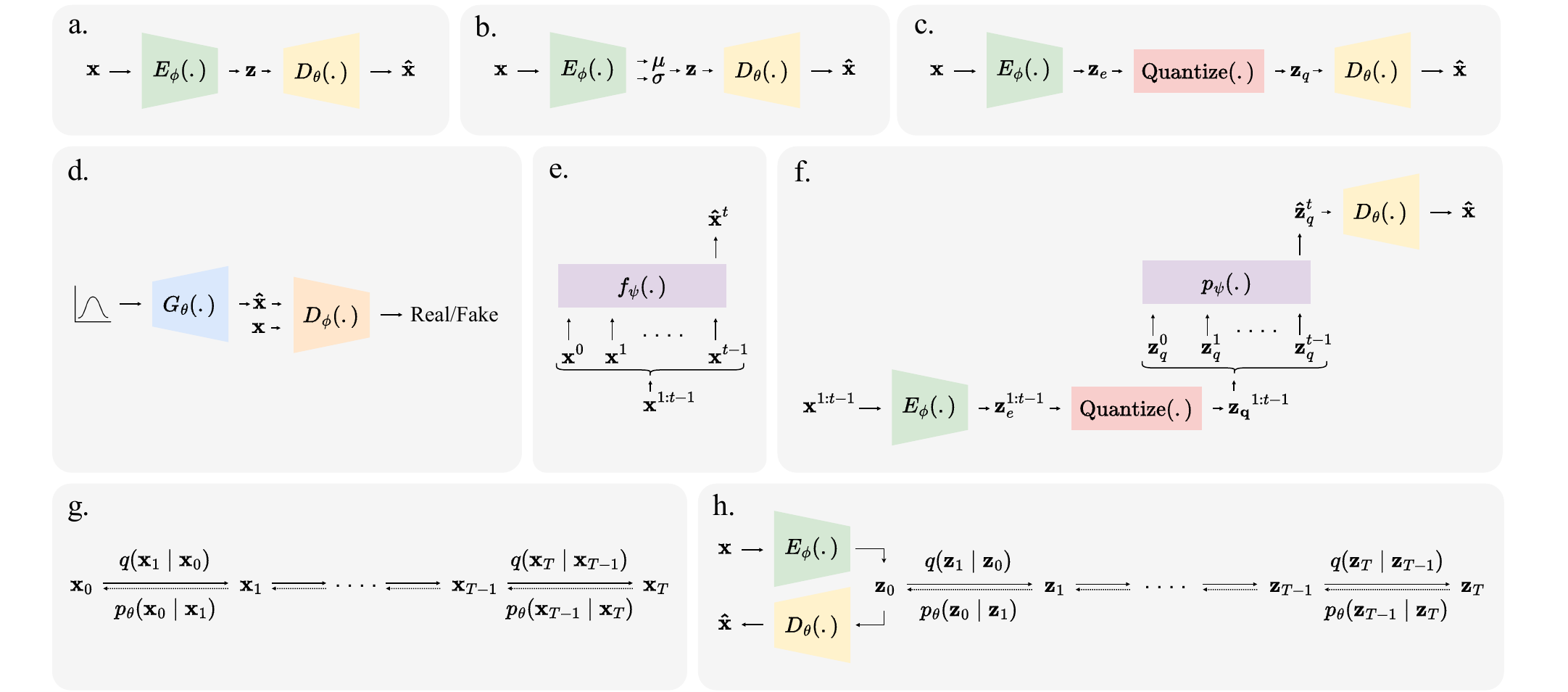}
    % \caption{Overview of the approaches we focused on in this survey. The notation used in this figure is consistent with the section~\ref{motion_generation_methods_background}. \textbf{a.} shows Autoencoder, \textbf{b.} shows Variational Autoencoder, \textbf{c.} shows Vector Quantized Variational Autoencoder, \textbf{d.} shows Generative Adversarial Networks, \textbf{e.} shows Continuous Autoregressive, \textbf{f.} shows Discrete Autoregressive, \textbf{g.} shows Diffusion and also Flow Matching, and \textbf{h.} shows Latent Diffusion}
    \caption{Overview of the motion generation approaches covered in this survey. The notations used in the figure follow the conventions introduced in Section~\ref{motion_generation_methods_background}. \textbf{(a)} Autoencoder, \textbf{(b)} Variational Autoencoder (VAE), \textbf{(c)} Vector Quantized Variational Autoencoder (VQ-VAE), \textbf{(d)} Generative Adversarial Networks (GANs), \textbf{(e)} Continuous Autoregressive Models, \textbf{(f)} Discrete Autoregressive Models, \textbf{(g)} Diffusion Models, and \textbf{(h)} Latent Diffusion Models.}
    \label{fig:approaches}
\end{figure*}
\subsubsection{\textbf{Autoencoder}}

Autoencoders~\citep{hinton2006reducing} are neural networks trained to reconstruct their input through a bottleneck structure, learning a compressed latent representation. An autoencoder consists of two components: an encoder $E_\phi$ and a decoder $D_\theta$, where the encoder maps an input sequence $\mathbf{x} \in \mathbb{R}^d$ to a latent code $\mathbf{z} \in \mathbb{R}^k$:
\[
\mathbf{z} = E_\phi(\mathbf{x}),
\]
and the decoder reconstructs the original input from the latent code:
\[
\hat{\mathbf{x}} = D_\theta(\mathbf{z}) = D_\theta(E_\phi(\mathbf{x})).
\]
To train an autoencoder, we minimize the reconstruction loss, which is typically the mean squared error (MSE):
\[
\mathcal{L}_{\text{rec}} = \| \mathbf{x} - \hat{\mathbf{x}} \|_2^2 = \| \mathbf{x} - D_\theta(E_\phi(\mathbf{x})) \|_2^2.
\]

\paragraph{\textbf{Variational Autoencoder (VAE)}}
VAEs~\citep{kingma2013auto} extend autoencoders by introducing a probabilistic latent space. The encoder outputs parameters of a Gaussian distribution: $\mu, \sigma = E_\phi(\mathbf{x})$, from which the latent vector is sampled:
\[
\mathbf{z} \sim q_\phi(\mathbf{z}|\mathbf{x}) = \mathcal{N}(\mathbf{z}; \mu, \text{diag}(\sigma^2)),
\]
and for training, it combines the reconstruction loss with a Kullback-Leibler (KL) divergence to enforce closeness to a prior $p(\mathbf{z}) = \mathcal{N}(0, I)$:
\[
\mathcal{L}_{\text{VAE}} = \mathbb{E}_{q_\phi(\mathbf{z}|\mathbf{x})}[\| \mathbf{x} - D_\theta(\mathbf{z}) \|^2] + \text{KL}(q_\phi(\mathbf{z}|\mathbf{x}) \| p(\mathbf{z})).
\]

\paragraph{\textbf{Vector Quantized Variational Autoencoder (VQ-VAE)}}  
VQ-VAE~\citep{oord2017neural} introduces discrete latent representations by learning a finite codebook $\mathcal{C} = \{\mathbf{e}_1, \mathbf{e}_2, \dots, \mathbf{e}_K\}$ of embedding vectors $\mathbf{e}_k \in \mathbb{R}^k$. The encoder outputs continuous latent vectors which are quantized to the nearest codebook entry. Formally, given encoder output $\mathbf{z}_e = E_\phi(\mathbf{x})$, the quantized latent vector $\mathbf{z}_q$ is:
\[
\mathbf{z}_q = \mathrm{Quantize}(\mathbf{z}_e) = \mathbf{e}_{k^*} \quad \text{where} \quad k^* = \arg\min_k \| \mathbf{z}_e - \mathbf{e}_k \|_2,
\]
and the decoder reconstructs the input from the quantized vector:
\[
\hat{\mathbf{x}} = D_\theta(\mathbf{z}_q).
\]

Training optimizes a loss composed of three terms \citep{oord2017neural}:
\[
\mathcal{L}_{\text{VQ}} = \| \mathbf{x} - \hat{\mathbf{x}} \|_2^2 + \| \mathrm{sg}[\mathbf{z}_e] - \mathbf{e}_{k^*} \|_2^2 + \beta \| \mathbf{z}_e - \mathrm{sg}[\mathbf{e}_{k^*}] \|_2^2,
\]
where $\mathrm{sg}[\cdot]$ denotes the stop-gradient operation, and $\beta$ is a hyperparameter controlling commitment to the codebook.

% This approach allows learning a discrete latent space, combining the benefits of autoencoding with efficient, quantized representations useful for discrete token-based generative models.

% Autoencoders are widely used for motion representation learning, denoising, and generative modeling by sampling from the latent space.
\subsubsection{\textbf{Autoregressive Models}}

Autoregressive (AR) models generate sequences by predicting each element based on previously observed elements. The core idea is to capture temporal or sequential dependencies by modeling each step conditioned on all prior steps. This paradigm is widely used across domains involving sequential data, such as speech~\citep{oord2016wavenet}, text~\citep{radford2019language}, video~\citep{villegas2017learning}, and motion~\citep{martinez2017human}, due to its effectiveness in modeling complex dynamics and temporal correlations.

We categorize AR models into two broad classes based on the representation of the sequence data: \textit{Continuous Autoregressive Models} and \textit{Discrete Autoregressive Models}.

\paragraph{\textbf{Continuous Autoregressive Models}}

In continuous AR models, the data is represented in a real-valued space. The model predicts the next element \( \hat{\mathbf{x}}^t \in \mathbb{R}^d \) from the preceding elements \( \mathbf{x}^{1:t-1} = \{\mathbf{x}^1, \dots, \mathbf{x}^{t-1}\} \) through a deterministic function \( f_\theta \):

\[
\hat{\mathbf{x}}^t = f_\psi(\mathbf{x}^{1:t-1}).
\]

Training these models typically involves minimizing a regression loss such as MSE:

\[
\mathcal{L}^{\text{reg}} = \sum_{t=1}^{T} \| \hat{\mathbf{x}}^t - \mathbf{x}^t \|^2.
\]

% Common architectures include recurrent neural networks (RNNs), long short-term memory networks (LSTMs), and Transformers. Continuous AR models excel at preserving fine-grained details but can suffer from error accumulation during sequential generation.

\paragraph{\textbf{Discrete Autoregressive Models}}

Discrete AR models first transform continuous or structured data into sequences of discrete tokens, which is often achieved using vector quantization methods like VQ-VAE~\citep{oord2017neural}. These methods encode continuous data vectors \( \mathbf{x}^t \in \mathbb{R}^d \) into discrete codes \( \mathbf{z}^t_q \in \mathcal{V} \), using \(E_\phi(.)\) and \(\mathrm{Quantize}(.)\) in the case of VQ-VAE, where \( \mathcal{V} \) is a finite vocabulary. This means that the input sequence $\mathbf{x}$ gets encoded to token sequence $\mathbf{z}_q$ which is \( \mathbf{z}_q = \{\mathbf{z}^1_q, \dots, \mathbf{z}^T_q\} \).

The model then factorizes the token distribution autoregressively as

\[
p(\mathbf{z}_q) = \prod_{t=1}^{T} p(\mathbf{z}^t_q \mid \mathbf{z}^{1:t-1}_q),
\]

and is trained to maximize the likelihood of the correct token sequence by minimizing the cross-entropy loss:

\[
\mathcal{L}^{\text{CE}} = -\sum_{t=1}^{T} \log p_\psi(\mathbf{z}^t_q \mid \mathbf{z}^{1:t-1}_q).
\]

At inference, tokens are generated sequentially, and a decoder (e.g., the VQ-VAE decoder) maps the tokens back to the original data space:

\[
\hat{\mathbf{x}}^t = D_{\theta}(\hat{\mathbf{z}}^t_q).
\]

% Discrete AR models benefit from advances in language modeling and large-scale sequence prediction, offering scalability and flexibility but may introduce quantization artifacts and lose subtle detail.

\subsubsection{\textbf{Generative Adversarial Network}}

Generative Adversarial Networks (GANs)~\citep{goodfellow2014generative} are a class of generative models that learn to synthesize realistic data by playing a two-player minimax game between two neural networks: a generator \( G_\theta \) and a discriminator \( D_\phi \). The generator maps a random noise vector \( \mathbf{z} \sim p(\mathbf{z}) \), typically sampled from a simple prior such as a multivariate standard normal distribution \( \mathcal{N}(0, I) \), to the data space:
\[
\hat{\mathbf{x}} = G_\theta(\mathbf{z}),
\]
while the discriminator attempts to distinguish between real samples \( \mathbf{x} \sim p_{\text{data}} \) and generated samples \( \hat{\mathbf{x}} \sim p_G \). The GAN objective is formulated as a minimax optimization problem:
\[
\min_{\theta} \max_{\phi} \,\, \mathbb{E}_{\mathbf{x} \sim p_{\text{data}}}[\log D_\phi(\mathbf{x})] + \mathbb{E}_{\mathbf{z} \sim p(\mathbf{z})}[\log(1 - D_\phi(G_\theta(\mathbf{z})))].
\]

The generator is trained to fool the discriminator by producing data that is indistinguishable from real data, while the discriminator is trained to correctly classify real versus fake samples. Ideally, this adversarial training results in the generator learning a distribution \( p_G \) that matches the true data distribution \( p_{\text{data}} \).

% In practice, various variants of GANs have been proposed to improve training stability and output quality. These include:
% \begin{itemize}
%   \item \textbf{Conditional GAN (cGAN)} \citep{mirza2014conditional}: extends GANs to conditional generation by providing side information \( \mathbf{c} \) (e.g., labels, trajectories, or past motion) to both \( G_\theta \) and \( D_\phi \):
%   \[
%   \hat{\mathbf{x}} = G_\theta(\mathbf{z}, \mathbf{c}), \quad D_\phi(\mathbf{x}, \mathbf{c}).
%   \]
%   \item \textbf{Wasserstein GAN (WGAN)} \citep{arjovsky2017wasserstein}: replaces the original loss with the Wasserstein distance for improved gradient behavior:
%   \[
%   \mathcal{L}_{\text{WGAN}} = \mathbb{E}_{\mathbf{x} \sim p_{\text{data}}}[D_\phi(\mathbf{x})] - \mathbb{E}_{\mathbf{z} \sim p(\mathbf{z})}[D_\phi(G_\theta(\mathbf{z}))].
%   \]
% \end{itemize}

% GANs have been widely applied in motion generation to produce realistic sequences that capture complex distributions. They are often combined with recurrent or convolutional architectures, and are especially effective when paired with discriminators that assess temporal coherence or physical plausibility.

\subsubsection{\textbf{Diffusion Models}}

Diffusion models~\citep{sohl2015deep, ho2020denoising} are a class of generative models that synthesize data by learning to reverse a gradual noising process. These models have shown strong performance in generating high-quality samples across domains such as images, audio, and motion. The core idea is to train a neural network to denoise samples through a sequence of steps, effectively learning the inverse of a diffusion process that progressively corrupts data with noise.

\paragraph{\textbf{Vanilla Diffusion Models}}

Let \( \mathbf{x}_0 \sim p_{\text{data}}(\mathbf{x}) \) denote a data sample. The forward (diffusion) process adds Gaussian noise over \( T \) steps according to a fixed schedule:
\[
q(\mathbf{x}_{1:T} \mid \mathbf{x}_0) = \prod_{t=1}^{T} q(\mathbf{x}_t \mid \mathbf{x}_{t-1}),
\]
where
\[
q(\mathbf{x}_t \mid \mathbf{x}_{t-1}) = \mathcal{N}(\mathbf{x}_t; \sqrt{1 - \beta_t} \mathbf{x}_{t-1}, \beta_t \mathbf{I}),
\]
with \( \beta_t \in (0,1) \) denoting the variance schedule.

The reverse process is defined as:
\[
p_\psi(\mathbf{x}_{0:T}) = p(\mathbf{x}_T) \prod_{t=1}^{T} p_\psi(\mathbf{x}_{t-1} \mid \mathbf{x}_t),
\]
where \( p(\mathbf{x}_T) = \mathcal{N}(0, \mathbf{I}) \) and each \( p_\psi(\mathbf{x}_{t-1} \mid \mathbf{x}_t) \) is a Gaussian whose mean and variance are predicted by a neural network trained to reverse the diffusion.

The model can be trained using different parameterizations of the reverse process, each predicting a different quantity related to the noisy sample \( \mathbf{x}_t \). The most common formulations include:

\begin{itemize}
  \item \textbf{Noise (epsilon) prediction:} The model predicts the noise \( \boldsymbol{\epsilon} \) added during the forward process~\citep{ho2020denoising}:
  \[
  \mathcal{L}_{\boldsymbol{\epsilon}} = \mathbb{E}_{t, \mathbf{x}_0, \boldsymbol{\epsilon}} \left[ \left\| \boldsymbol{\epsilon} - \epsilon_\psi(\mathbf{x}_t, t) \right\|^2 \right],
  \]
  where \( \mathbf{x}_t = \sqrt{\bar{\alpha}_t} \mathbf{x}_0 + \sqrt{1 - \bar{\alpha}_t} \boldsymbol{\epsilon} \), \( \boldsymbol{\epsilon} \sim \mathcal{N}(0, \mathbf{I}) \), and \( \bar{\alpha}_t = \prod_{s=1}^t (1 - \beta_s) \).

  \item \textbf{Sample (x\textsubscript{0}) prediction:} The model directly predicts the clean data \( \mathbf{x}_0 \)~\citep{dhariwal2021diffusion}:
  \[
  \mathcal{L}_{\mathbf{x}_0} = \mathbb{E}_{t, \mathbf{x}_0, \boldsymbol{\epsilon}} \left[ \left\| \mathbf{x}_0 - x_\psi(\mathbf{x}_t, t) \right\|^2 \right].
  \]

  \item \textbf{Velocity (v) prediction:} The model predicts the “velocity” vector \( \mathbf{v} \), a scaled combination of \( \mathbf{x}_0 \) and \( \boldsymbol{\epsilon} \), defined as~\citep{salimans2022progressive}:
  \[
  \mathbf{v} = \sqrt{\bar{\alpha}_t} \boldsymbol{\epsilon} - \sqrt{1 - \bar{\alpha}_t} \mathbf{x}_0,
  \]
  and the loss becomes:
  \[
  \mathcal{L}_{\mathbf{v}} = \mathbb{E}_{t, \mathbf{x}_0, \boldsymbol{\epsilon}} \left[ \left\| \mathbf{v} - v_\psi(\mathbf{x}_t, t) \right\|^2 \right].
  \]
\end{itemize}

Each parameterization offers trade-offs in terms of training stability, sample quality, and compatibility with conditioning mechanisms.

At inference, sampling proceeds from Gaussian noise \( \mathbf{x}_T \sim \mathcal{N}(0, \mathbf{I}) \) and is denoised step-by-step using the learned reverse process.

\paragraph{\textbf{Latent Diffusion Models}}

Latent Diffusion Models (LDMs)~\citep{rombach2022high} apply diffusion not in the original high-dimensional data space, but in a compressed latent space learned by an autoencoder. This significantly reduces the computational cost while maintaining high generative fidelity. Let \( E_\phi(\cdot) \) and \( D_\theta(\cdot) \) denote the encoder and decoder networks, respectively. A data sample \( \mathbf{x}_0 \) is first mapped to its latent representation:
\[
\mathbf{z}_0 = E_\phi(\mathbf{x}_0),
\]
and the forward diffusion process is applied in the latent space:
\[
\mathbf{z}_t = \sqrt{\bar{\alpha}_t} \mathbf{z}_0 + \sqrt{1 - \bar{\alpha}_t} \boldsymbol{\epsilon}, \quad \boldsymbol{\epsilon} \sim \mathcal{N}(0, \mathbf{I}),
\]
where \( \bar{\alpha}_t = \prod_{s=1}^t (1 - \beta_s) \).

The reverse process is learned in the latent space using the same parameterizations as in standard diffusion models (i.e., predicting noise \( \boldsymbol{\epsilon} \), the original latent code \( \mathbf{z}_0 \), or the velocity vector \( \mathbf{v} \)). Once the denoised latent vector \( \hat{\mathbf{z}}_0 \) is recovered through the learned reverse process, the final data reconstruction is obtained via decoding:
\[
\hat{\mathbf{x}}_0 = D_\theta(\hat{\mathbf{z}}_0).
\]

% LDMs preserve modeling capacity while allowing diffusion models to scale to high-dimensional domains like images, audio, and motion, where operating directly in pixel or feature space would be computationally prohibitive.

\subsubsection{\textbf{Flow Matching}}

Flow Matching (FM)~\citep{lipman2023flow} is a class of generative models that learns a continuous transformation from a simple prior distribution (e.g., Gaussian noise) to the data distribution by modeling vector fields that define deterministic flows in space. Unlike diffusion models that rely on stochastic sampling via reverse-time SDEs, flow matching directly learns the velocity field of an ordinary differential equation (ODE) that transports samples from the prior to the data distribution in finite time.

Let \( \mathbf{x}_0 \sim p_{\text{data}} \) be a sample from the data distribution, and let \( \mathbf{x}_1 \sim p_{\text{prior}} \), typically \( \mathcal{N}(0, I) \), be a sample from a base distribution. Flow matching seeks to learn a continuous-time path \( \mathbf{x}(t) \in \mathbb{R}^d \), \( t \in [0, 1] \), such that:
\[
\frac{d\mathbf{x}(t)}{dt} = \mathbf{v}_\theta(\mathbf{x}(t), t), \quad \mathbf{x}(0) = \mathbf{x}_0, \quad \mathbf{x}(1) = \mathbf{x}_1.
\]

In practice, instead of learning the exact ODE solution, FM trains a neural network to match a predefined vector field \( \mathbf{v}^* \) that describes linear interpolants between \( \mathbf{x}_0 \) and \( \mathbf{x}_1 \):
\[
\mathbf{x}(t) = (1 - t)\mathbf{x}_0 + t \mathbf{x}_1, \quad \Rightarrow \quad \mathbf{v}^*(\mathbf{x}(t), t) = \frac{d\mathbf{x}(t)}{dt} = \mathbf{x}_1 - \mathbf{x}_0.
\]

The model \( \mathbf{v}_\theta \) is then trained to match this target field by minimizing the expected squared error:
\[
\mathcal{L}_{\text{FM}} = \mathbb{E}_{\mathbf{x}_0, \mathbf{x}_1, t} \left[ \left\| \mathbf{v}_\theta(\mathbf{x}(t), t) - (\mathbf{x}_1 - \mathbf{x}_0) \right\|^2 \right],
\]
where \( t \sim \mathcal{U}[0, 1] \), and \( \mathbf{x}(t) = (1 - t)\mathbf{x}_0 + t \mathbf{x}_1 \) is a point along the interpolated trajectory between data and prior samples.

At inference time, a sample \( \mathbf{x}_1 \sim p_{\text{prior}} \) is evolved backward through the learned flow field using numerical ODE solvers such as Runge-Kutta, starting from \( t = 1 \) to \( t = 0 \), producing a synthetic data sample:
\[
\hat{\mathbf{x}}_0 = \text{ODESolve}(\mathbf{x}_1, \mathbf{v}_\theta, t=1 \to 0).
\]

% Flow Matching models are deterministic, sample-efficient, and typically require fewer function evaluations than diffusion models during sampling, making them attractive for high-speed generation tasks. They also admit flexible conditioning and are closely related to optimal transport formulations in probability.

\subsubsection{\textbf{Implicit Neural Representation}}

Implicit Neural Representations (INRs)~\citep{sitzmann2020implicit} model data as continuous functions rather than discrete arrays. Instead of storing or predicting values at discrete positions (e.g., pixels or motion frames), an INR uses a neural network \( f_\theta \) to represent the underlying continuous signal as a function of a coordinate \( \mathbf{u} \in \mathbb{R}^d \). The network directly maps input coordinates to data values:
\[
\hat{\mathbf{x}}(\mathbf{u}) = f_\theta(\mathbf{u}),
\]
where \( \mathbf{u} \) could represent time, space, or spatio-temporal coordinates (e.g., a motion frame index or joint index), and \( \hat{\mathbf{x}}(\mathbf{u}) \) is the predicted value at that coordinate.

To train the network, a set of coordinate-value pairs \( \{(\mathbf{u}_i, \mathbf{x}_i)\}_{i=1}^N \) is used, and the model minimizes the reconstruction loss, typically:
\[
\mathcal{L}_{\text{INR}} = \sum_{i=1}^N \left\| \mathbf{x}_i - f_\theta(\mathbf{u}_i) \right\|^2.
\]

Since the representation is continuous, INRs can query the signal at arbitrary resolutions and support smooth interpolation. To improve the representation of high-frequency details, the input coordinates are often passed through a positional encoding function \( \gamma(\mathbf{u}) \), such as:
\[
\gamma(\mathbf{u}) = \left[ \sin(2^0 \pi \mathbf{u}), \cos(2^0 \pi \mathbf{u}), \dots, \sin(2^{L-1} \pi \mathbf{u}), \cos(2^{L-1} \pi \mathbf{u}) \right],
\]
which helps the network approximate more complex signals~\citep{tancik2020fourier}.

INRs are particularly useful in motion modeling for representing entire motion clips as neural functions of time or space-time coordinates. This enables compact representations, continuous interpolation, and generalization to variable-length or continuous-time outputs. They are also compatible with conditioning (e.g., on style or trajectory) by augmenting the function \( f_\theta(\mathbf{u}, \mathbf{c}) \) with conditioning inputs \( \mathbf{c} \).
\\
\subsubsection{\textbf{Physics-Based Methods}}
Physics-based methods refer to computational techniques that solve physical systems by directly applying the laws of physics. With data-driven methods for physics-based motion, the main challenge lies in producing and leveraging architectures capable of learning a wide variety of realistic motions. Among the various approaches for integrating physics with data-driven methods, four strategies capture most of them effectively: physics-based priors, reinforcement learning-based (RL), differentiable solvers or simulation-driven methods, and physics-informed neural networks (PINN). Although the methods evaluated in this survey primarily focus on the first two strategies, we will briefly cover differentiable solvers and PINN.

\paragraph{\textbf{Physics-Based Priors}}
A physics-based prior encodes physical laws and constraints, such as dynamics, energy conservation, stability, contact mechanics, momentum, and so on. In machine learning, a prior generally refers to pre-existing knowledge $p$ about the likelihood of generating an output $x$ before observing any data $X$. The prior distribution $p(\theta)$ encodes this knowledge about $\theta$ before seeing $X$. After observing data $X$, we update the likelihood of this pre-existing knowledge using Bayes' theorem:
\begin{equation}
p(\theta \mid X) = \frac{p(X \mid \theta) \, p(\theta)}{p(X)}
\end{equation}

Physical priors can be introduced at various stages before or during training, but they are most often used to define more accurate loss functions using physical properties. These losses enforce physical consistency by penalizing behaviors that violate known physical laws, thereby guiding the model toward more plausible and realistic solutions.

\paragraph{\textbf{RL-Based Methods}}
Reinforcement Learning (RL)~\citep{watkins1989learning} is a framework for learning sequential decision-making policies through interaction with an environment. RL is often formalized as a Markov Decision Process defining a set of environment and agent states $\mathcal{S}$, a set of actions for the agent $\mathcal{A}$, transition probability function $\mathcal{T}(s' \mid s,a)$, and a reward function $r(s, a)$. At each time step $t$ (i.e., between two states), the agent observes a state $s_t \in \mathcal{S}$ and the current action $a_t$ taken from a policy $\pi_\theta$ (parameterized by $\theta$), and receives a reward $r_t$ accordingly. The system then transitions to a new state $s_{t+1} \sim \mathcal{T}(s_{t+1} \mid s_t, a_t)$. The goal of this system is to find a policy that maximizes the expected cumulative reward:
\begin{equation}
J(\theta) = \mathbb{E}_{\tau \sim \pi_\theta} \left[ \sum_{t=0}^{\infty} \gamma^t r(s_t, a_t) \right],
\end{equation}
where $\gamma \in [0, 1)$ is the discount factor. This is typically achieved through a policy gradient method~\citep{sutton1999policy}, which estimates $\nabla_{\theta} J(\theta)$ and updates the parameters $\theta$ accordingly.

\paragraph{\textbf{Differentiable Solvers and PINN}}
A Physics-Informed Neural Network (PINN)~\citep{raissi2019physics} is a neural network designed to solve problems governed by physical laws for complex systems expressed using Partial Differential Equations (PDE) and Ordinary Differential Equations (ODE). PINNs often require the use of differentiable solvers~\citep{um2020solver} to facilitate gradient-based learning and optimization within systems involving physical simulations and optimization tasks.

More formally, in this setting, the objective is to constrain the problem of estimating a data-driven solution to a generalized PDE formulation:
\begin{equation}
\label{eq:pde_general_form}
u(t,x) + \mathcal{N}(u) = 0,
\end{equation}
where $u(t,x)$ denotes the latent component, $\mathcal{N}(\cdot)$ a nonlinear differential operator, and the simulation domain $\Omega$. Since this study focuses on time-dependent motion generation and to streamline the associated notation, we limit our discussion in this section to continuous-time models (i.e., excluding discrete-time models). We define $f(t,x)$ as the left-hand side of Eq.~\ref{eq:pde_general_form}, i.e., the PINN:
\begin{equation}
f(t,x)=u(t,x)+\mathcal{N}(u),
\end{equation}
allowing us to approximate $u(t,x)$ throughout a deep neural network. Both $f(t,x)$ and $u(t,x)$ can be learned by minimizing the MSE as follows:
\begin{equation}
{MSE}_u = \frac{1}{N_u}\sum^{N_u}_{i=1} |u(t^i_u,x^i_u)-u^i |^2,
\end{equation}
and
\begin{equation}
{MSE}_f = \frac{1}{N_f}\sum^{N_f}_{i=1} |f(t^i_u,x^i_u)|^2.
\end{equation}
where the ${MSE}_u$ expresses the initial data while ${MSE}_f $ enforces the PDE assumption imposed by Eq~\ref{eq:pde_general_form}.
For instance, this can be applied to incompressible fluid flows using the Navier-Stokes equations and formulated as an MSE loss equal to:
\begin{eqnarray}
&& \frac{1}{N} \sum_{i=1}^N ( | u(t^i, x^i, y^i) - u^i |^2 + | \nu(t^i, x^i, y^i) - \nu^i |^2 ) \\ \nonumber
&& + \frac{1}{N} \sum_{i=1}^N ( | f(t^i, x^i, y^i) |^2 + | g(t^i, x^i, y^i) |^2 ),
\end{eqnarray}
where $u$ and $\nu$ are respectively the $x-$ and $y-$ components of the velocity field, and $f$ and $g$ are the PINN assumptions to enforce aligning the learned model so that the solution converges to a divergence-free function (i.e., making the flow incompressible: $u_x+\nu_x=0$). The same approach can be instrumental in motion to approximate energy-based momentum and soft/hard dynamics.

We refer readers to ~\citep{raissi2019physics} and~\citep{karniadakis2021physics} for more details on this topic.
\section{\textbf{Methods}}
\label{sec:methods}

In recent years, a wide variety of generation approaches have been proposed for motion generation, each leveraging different inductive biases, training objectives, and output representations. This section categorizes existing works based on the core generative framework they employ, including simple feed-forward networks~\ref{sec:simple-nn}, autoencoders~\ref{sec:ae}, variational~\ref{sec:vae} and quantized extensions~\ref{sec:vqvae}, autoregressive models (both continuous~\ref{sec:continuous-AR} and discrete~\ref{sec:discrete-AR}), generative adversarial networks~\ref{sec:GAN}, diffusion models~\ref{sec:diffusion} and their latent variants~\ref{sec:latent-diffusion}, flow matching techniques~\ref{sec:flow-matching}, implicit neural representations~\ref{sec:inr}, and physics/optimization-based methods~\ref{sec:physics-based}. For each approach, we further organize representative works according to their specific application domains ranging from text-to-motion synthesis and trajectory-constrained generation to physics-aware modeling and style transfer, highlighting how each method is adapted to the unique challenges of these tasks.

To provide a temporal perspective on research progress, we also include a timeline visualization (Fig.~\ref{fig:timeline}) that maps all surveyed works along with their associated generative approaches using symbolic markers.

In addition, we summarize representative papers for each approach in a corresponding table that reports: (1) the method name, (2) the primary architecture, referring to the core modeling component (e.g., attention in transformer-based models), (3) the condition type, indicating the input conditioning used during generation, where "Poses" may refer to past motion, target frames, or in-betweening cues, (4) the datasets employed, and (5) the application category that reflects the paper’s primary task focus.

\begin{figure*}[htbp]
    \centering
    \includegraphics[height=0.95\textheight]{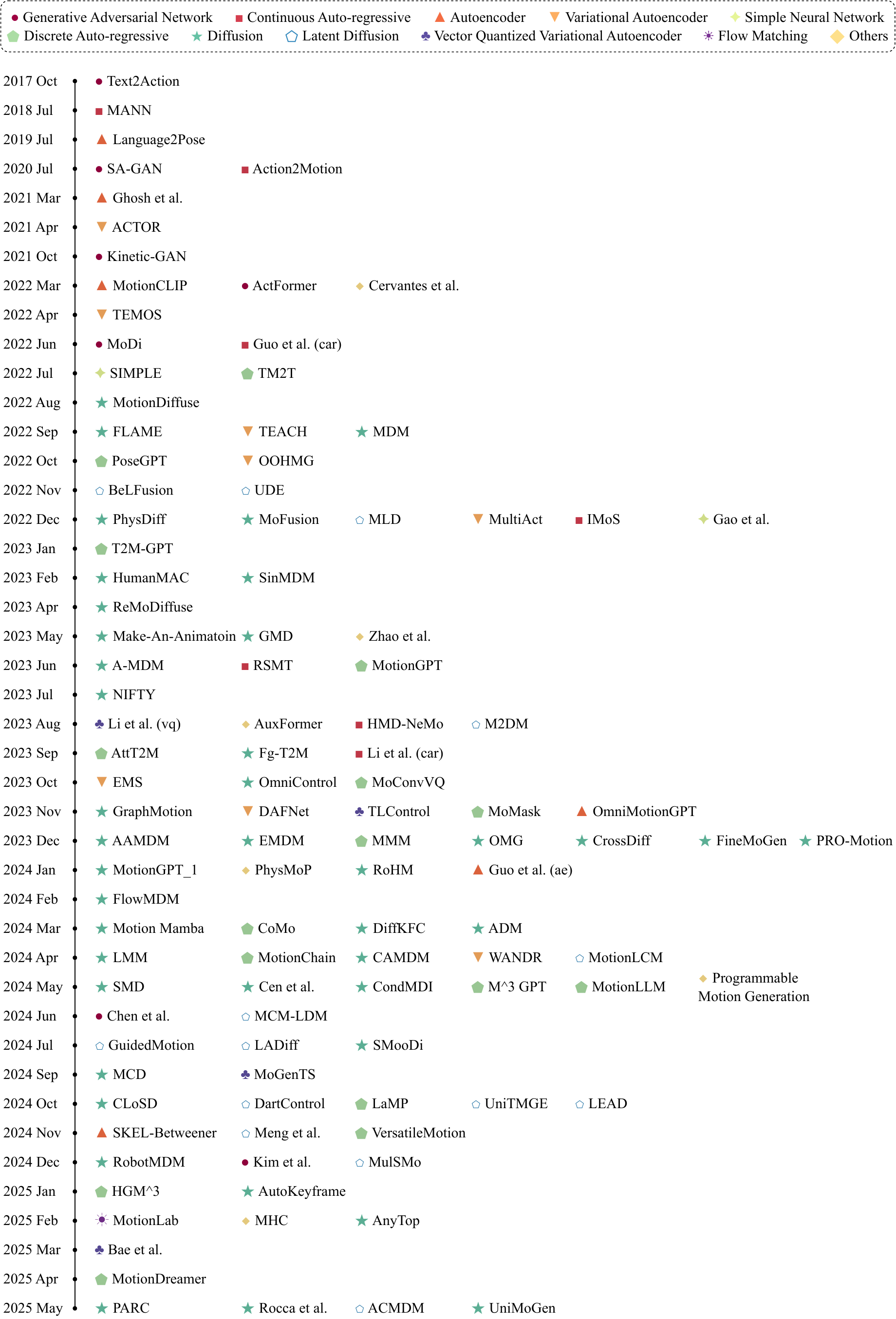}
    \vspace{10pt}
    \caption{Timeline of the reviewed papers. Papers are organized chronologically within each row.}
    \label{fig:timeline}
\end{figure*}

\subsection{\textbf{Feed-forward Motion Generation Models}}\label{sec:simple-nn}
At the opposite end of the spectrum from today’s large diffusion or transformer models, a line of work revisits \emph{minimal} feed-forward architectures, often no more than a few linear layers, to interrogate how far one can go with carefully chosen signal representations and loss functions.  
The appeal of such simple neural networks lies in their extremely low computational footprint, ease of training, and transparent inductive biases; despite their frugality, they provide competitive baselines and valuable insights into the role of frequency content and temporal correlations in motion.  
A concise overview of the corresponding studies is given in Table~\ref{tab:simple_nn}.

\paragraph{Motion Prediction, Completion \& Autoregressive Control}
The core challenge in motion prediction is to forecast future poses that remain both kinematically plausible and dynamically consistent with the immediate past.  
\textit{SIMPLE}~\citep{guo2023back} demonstrates that this can be achieved with nothing more than a linear multi-layer perceptron devoid of ReLU activations, provided the input sequence is first transformed into the frequency domain via a Discrete Cosine Transform (DCT).  
By predicting residual joint displacements and supervising an auxiliary velocity loss, the model accurately captures short-term dynamics while avoiding the over-smoothing commonly observed in more complex recurrent networks~\citep{fragkiadaki2015recurrent}.
Building on the importance of frequency analysis, \textit{Gao et al.}~\citep{gao2023decompose} propose a two-stage strategy that explicitly decomposes motion into multi-view spectral components through a Frequency Decomposition Unit (FDU) and subsequently aggregates them with a Feature Aggregation Unit (FAU).  
This decomposition-aggregation pipeline offers robustness against noise and improves long-horizon prediction fidelity, confirming that careful frequency manipulation, even within a lightweight MLP framework, can rival or surpass far heavier architectures.

Taken together, these studies illustrate that, when armed with the right signal processing tools, simple feed-forward networks remain a powerful and interpretable baseline for motion prediction and control.
\begin{table*}[htbp]
	\renewcommand{\arraystretch}{1.25}
	\resizebox{\textwidth}{!}{
		% \begin{tabular}{l|c|c|c|c}
            \begin{tabular}{lcc >{\centering\arraybackslash}p{3.0cm} c}
			\toprule
                \multicolumn{1}{c}{{\textbf{Method}}} & 
                \multicolumn{1}{c}{{\textbf{Primary Architecture}}} & 
                \multicolumn{1}{c}{{\textbf{Condition}}} & 
                \multicolumn{1}{c}{{\textbf{Dataset}}} & 
                \multicolumn{1}{c}{{\textbf{Category}}} \\	                
    		\midrule

    			   SIMPLE~\citep{guo2023back}
    			  & Linear
    			& Poses
    			& \citep{ionescu2013human3}, \citep{mahmood2019amass}, \citep{von2018recovering}   
                    & Motion Prediction / Completion \& Autoregressive Control \\
                    
    			   Gao et al.~\citep{gao2023decompose}
                    & Convolution
    			& Poses 
    			& \citep{cmu}, \citep{ionescu2013human3}, \citep{von2018recovering}   
                    & Motion Prediction / Completion \& Autoregressive Control \\
                \bottomrule
		\end{tabular}}
	\vspace{0.5\baselineskip}
	\caption{Papers using basic \textbf{feed-forward neural networks} for motion generation, without explicitly modeling the generative distribution. Entries are listed in chronological order.}
	\label{tab:simple_nn}
\end{table*}

\subsection{\textbf{Autoencoder-based Motion Generation}}\label{sec:ae}
Autoencoders constitute one of the earliest and most versatile neural architectures for motion generation.  
By compressing skeletal pose sequences into a low dimensional latent space and reconstructing them back to the original domain, autoencoders provide (i) a compact motion representation that is amenable to downstream conditioning signals such as language, trajectories, or style labels, and (ii) a differentiable decoder that can be steered by gradient-based objectives.  
Consequently, a wide spectrum of works leverage autoencoders as stand-alone generators.  
A concise overview of the papers discussed below is given in Table~\ref{tab:ae}.

\paragraph{Text-to-Motion Generation}
Mapping free-form language to plausible body movements requires a shared embedding space that captures both semantic and kinematic regularities.  
\textit{Language2Pose}~\citep{ahuja2019language2pose} pioneers this direction by jointly training a GRU-based text encoder and a motion encoder–decoder such that textual and motion embeddings co-locate. Additionally, a curriculum-learning schedule gradually increases sequence length to stabilize training.  
Building on the insight that different body parts are driven by distinct linguistic cues, the hierarchical two-stream model of~\citep{ghosh2021synthesis} factorizes the latent manifold into upper- and lower-body sub-spaces, enabling the synthesis of complex, composite actions from a single textual description.  
\textit{MotionCLIP}~\citep{tevet2022motionclip} further tightens language–motion alignment by distilling the latent space of a transformer autoencoder into that of the large-scale CLIP vision–language model, which grants the generator strong zero-shot generalization and the ability to decode CLIP embeddings back to motion.  
Together, these works establish a progression from task-specific co-embeddings to leveraging foundation vision–language models for richer semantic grounding.

\paragraph{Large Multi-modal / Multi-task Foundation Models}
While most autoencoder studies focus on human motion alone, \textit{OmniMotionGPT}~\citep{yang2024omnimotiongpt} demonstrates that shared latent space can be extended across species and tasks.  
The method employs separate motion autoencoders for humans and animals, but aligns their latents along with CLIP text features inside a GPT-style cross-domain decoder.  
This design transfers motor knowledge learned from abundant human data to the low-resource animal domain, yielding diverse animal motions from text prompts while still outperforming specialized human-motion baselines.  
The work exemplifies how the autoencoder latents can serve as the “universal language” for large multi-modal foundation models.

\paragraph{Stylized Motion and Style Transfer}
\textit{Guo et al. (ae)}~\citep{guo2024generative} revisit the latents of pre-trained motion autoencoders from a style-transfer perspective.  
By disentangling the representation into deterministic content codes and probabilistic style codes, their framework allows sampling, interpolating, or explicitly conditioning on style exemplars under various supervision regimes.  
The probabilistic modeling of style not only produces a spectrum of stylized outputs from a single content trajectory but also avoids mode collapse, often observed in deterministic style encoders.

\paragraph{Trajectory / Keyframe / Constraint-based Control and In-betweening}
Traditional animation workflows rely heavily on boundary poses and sparse joint constraints.  
\textit{SKEL-Betweener}~\citep{agrawal2024skel} translates this practice into a neural setting through a skeletal-transformer autoencoder trained to reconstruct full sequences while respecting editable neural motion curves.  
Its decoder can be interactively driven by two endpoint poses and optional per-joint masks, yielding real-time in-betweening that preserves constraint accuracy and artist intent.  
This work illustrates how autoencoder decoders can be re-purposed as neural motion rigs, bridging data-driven generation with user-controlled editing.

Collectively, the above studies highlight the flexibility of autoencoder frameworks: whether aligning motion with language, unifying cross-domain data, injecting stylistic variability, or enabling fine-grained control, the latent representation learned by an autoencoder remains a powerful currency for motion generation research.
\begin{table*}[ht]
	\renewcommand{\arraystretch}{1.25}
	\resizebox{\textwidth}{!}{
		% \begin{tabular}{l|c|c|c|c}
            \begin{tabular}{lcc >{\centering\arraybackslash}p{3.0cm} c}
			\toprule
                \multicolumn{1}{c}{{\textbf{Method}}} & 
                \multicolumn{1}{c}{{\textbf{Primary Architecture}}} & 
                \multicolumn{1}{c}{{\textbf{Condition}}} & 
                \multicolumn{1}{c}{{\textbf{Dataset}}} & 
                \multicolumn{1}{c}{{\textbf{Category}}} \\	                
    		\midrule
			
    			   Language2Pose~\citep{ahuja2019language2pose} 
                    & RNN
    			& Text 
    			& \citep{kit-ml} 
                    & Text-to-Motion Generation \\
    
    			   Ghosh et al.~\citep{ghosh2021synthesis} 
                    & RNN
    			& Text 
    			& \citep{kit-ml}  
                    & Text-to-Motion Generation  \\
    
    			   MotionCLIP~\citep{tevet2022motionclip} 
                    & Attention
    			& Text 
    			& \citep{punnakkal2021babel} 
                    & Text-to-Motion Generation  \\
            
                      OmniMotionGPT~\citep{yang2024omnimotiongpt}
                    & Attention
                    & Poses - Text
                    & \citep{yang2024omnimotiongpt}
                    & Large Multi-modal / Multi-task Foundation Models \\

                      Guo et al. (ae)~\citep{guo2024generative}
                    & Convolution
                    & Poses
                    & \citep{aberman2020unpaired}, \citep{cmu}, \citep{xia2015realtime}
                    & Stylized Motion \& Style Transfer \\
                    
                      SKEL-Betweener~\citep{agrawal2024skel}
                    & Attention
                    & Poses
                    & \citep{aristidou2017emotion}, \citep{harvey2020robust}, \citep{mahmood2019amass}
                    & Trajectory / Keyframe / Constraint-based Control \& In-betweening \\
                \bottomrule
		\end{tabular}}
	\vspace{0.5\baselineskip}
	\caption{Papers employing standard \textbf{autoencoders} to learn compact motion representations for reconstruction or generation tasks. Entries are listed in chronological order.}
	\label{tab:ae}
\end{table*}

\subsection{\textbf{Variational Autoencoder–based Motion Generation}}\label{sec:vae}
Variational autoencoders (VAEs) cast motion synthesis as probabilistic inference in a latent space: an encoder approximates the posterior distribution over latent variables given an observed pose sequence, and a decoder reconstructs or generates motion by sampling from the learned prior.  
This framework offers three central advantages for motion modeling: (i) it captures the intrinsic multi-modality of future motion through stochastic sampling; (ii) it supports one-shot, non-autoregressive decoding, which eliminates cumulative error and enables direct length control; and (iii) it provides a principled way to fuse heterogeneous conditioning signals such as text, actions, trajectories, or scene geometry via conditional priors.  
A summary of the VAE-based works discussed below is provided in Table~\ref{tab:vae}.

\paragraph{Text-to-Motion Generation}
Early work in this vein, \textit{TEMOS}~\citep{petrovich2022temos}, learns a joint latent space of language and motion using transformer layers inside a conditional VAE, thereby generating diverse sequences from a single textual prompt.  
\textit{TEACH}~\citep{athanasiou2022teach} extends this idea to paragraph-length descriptions by incorporating past motion as additional encoder input; within each atomic action, decoding is non-autoregressive, whereas a lightweight AR module stitches successive actions, leading to coherent long-form behavior.  
\textit{OOHMG}~\citep{lin2023being} removes the need for paired text–motion data altogether: a \emph{text-to-pose} generator, supervised by a distilled CLIP-based alignment module, creates a static pose that serves as a prompt for a masked-motion VAE, delivering a fully \emph{offline}, zero-shot pipeline.  
Focusing on compositionality, \textit{EMS}~\citep{Qian2023BreakingTL} decomposes complex sentences into atomic motions using a multimodal VAE and then refines temporal coherence with a motion-conditioned VAE; prompt engineering and synthetic augmentation further enhance controllability over long, multi-clause instructions.  
Collectively, these works chart a trajectory from paired-data VAEs to unsupervised and compositional generators capable of handling extensive narratives.

\paragraph{Action-conditioned Motion Generation}
\textit{ACTOR}~\citep{petrovich2021action} introduces a sequence-level conditional VAE equipped with transformers, producing variable-length motions in a single forward pass when given an action label and desired duration.  
A key benefit is the removal of any dependency on initial poses, allowing direct sampling of complete sequences.  
\textit{MultiAct}~\citep{lee2023multiact} targets the more challenging setting of long-range, multi-action synthesis.  
By recurrently rolling a VAE over past motion while maintaining a face-front canonical coordinate frame, it delivers minute-long animations that smoothly transition across action labels, demonstrating that latent-space sampling can be repeatedly re-initialized without degradation.

\paragraph{Scene \& Object-interaction Motion Generation}
\textit{DAFNet}~\citep{jin2023dafnet} employs a Gaussian-mixture VAE (GMVAE) to model the distribution of whole-body poses conditioned on target root and upper-body configurations as well as surrounding furniture geometry.  
The probabilistic latent captures multiple feasible interaction styles (e.g., different sitting strategies) and adapts to varying shapes, illustrating how VAEs can mediate contact-rich human–scene interactions without explicit physics simulation.

\paragraph{Trajectory / Keyframe / Constraint-based Control \& In-betweening}
Addressing goal-directed movement, \textit{WANDR}~\citep{diomataris2024wandr} augments the latent representation with \emph{intention} features, which are functions of current pose, goal location, and remaining time, computed on-the-fly at inference.  
These features guide the decoder towards the target while preserving stylistic diversity, showing that simple, differentiable heuristics embedded in the VAE prior suffice for accurate, temporally constrained navigation.

In summary, VAE-based frameworks span a spectrum from zero-shot text grounding to action sequencing, scene interaction, and goal-conditioned control.  
Their stochastic latents, non-autoregressive decoding, and flexible conditioning mechanisms make them a 
% \textcolor{red}{indispensable}
building block in the contemporary toolbox for motion generation.
\begin{table*}[ht]
	\renewcommand{\arraystretch}{1.25}
	\resizebox{\textwidth}{!}{
		% \begin{tabular}{l|c|c|c|c}
            \begin{tabular}{lcc >{\centering\arraybackslash}p{3.0cm} c}
			\toprule
                \multicolumn{1}{c}{{\textbf{Method}}} & 
                \multicolumn{1}{c}{{\textbf{Primary Architecture}}} & 
                \multicolumn{1}{c}{{\textbf{Condition}}} & 
                \multicolumn{1}{c}{{\textbf{Dataset}}} & 
                \multicolumn{1}{c}{{\textbf{Category}}} \\	                
    		\midrule
			
              ACTOR~\citep{petrovich2021action} 
            & Attention
            & Action - Time 
            & \citep{guo2020action2motion}, \citep{ji2018large}, \citep{shahroudy2016ntu}
            & Action-conditioned Motion Generation \\

              TEMOS~\citep{petrovich2022temos} 
            & Attention
            & Text 
            & \citep{kit-ml}  
            & Text-to-Motion Generation \\

              TEACH~\citep{athanasiou2022teach} 
            & Attention
            & Text - Time
            & \citep{punnakkal2021babel} 
            & Text-to-Motion Generation \\

              OOHMG~\citep{lin2023being} 
            & Attention
            & Text 
            & \citep{mahmood2019amass}, \citep{punnakkal2021babel} 
            & Text-to-Motion Generation \\

              MultiAct~\citep{lee2023multiact} 
            & Attention
            & Action - Poses
            & \citep{punnakkal2021babel}   
            & Action-conditioned Motion Generation \\

              EMS~\citep{Qian2023BreakingTL}
            & Attention
            & Text
            & \citep{kit-ml}, \citep{punnakkal2021babel}
            & Text-to-Motion Generation \\

              DAFNet~\citep{jin2023dafnet}
            & RNN
            & Poses - Trajectory
            & \citep{hassan2021stochastic}
            & Scene \& Object-interaction Motion Generation \\

              WANDR~\citep{diomataris2024wandr}
            & Linear
            & Poses - Time - Trajectory
            & \citep{araujo2023circle}, \citep{mahmood2019amass}
            & Trajectory / Keyframe / Constraint-based Control \& In-betweening \\
                
            \bottomrule
		\end{tabular}}
	\vspace{0.5\baselineskip}
	\caption{Papers leveraging \textbf{variational autoencoders (VAEs)} to model stochastic motion distributions, enabling diverse and controllable synthesis. Entries are grouped by research category.}
	\label{tab:vae}
\end{table*}

\subsection{\textbf{Vector-Quantized VAE (VQ-VAE) Motion Generation}}\label{sec:vqvae}
Vector-quantized variational autoencoders (VQ-VAEs) compress a motion sequence into an array of discrete codebook indices, thereby turning high-dimensional, real-valued trajectories into tokens that can be modeled with powerful sequence architectures such as transformers or masked autoregressive models.  
Compared with continuous autoencoders, the discrete latent space (i) regularizes the representation and limits the accumulation of small reconstruction errors, (ii) permits the direct application of natural-language techniques (e.g., masking, token weighting, or discrete diffusion) and (iii) enables part-wise or hierarchy-aware codebooks that reflect skeletal topology.  
Table~\ref{tab:vqvae} summarizes existing VQ-VAE-based works, which we review below.

\paragraph{Unconditional or Minimal-data Motion Generation}
In settings where no external control is provided, the goal is to learn a robust motion prior that can generate long, diverse sequences.  
\textit{Li et al. (vq)}~\citep{li2024controlling} demonstrate that combining a VQ-VAE prior with reinforcement learning (RL) can alleviate drift and out-of-distribution failures that typically cause problems during autoregressive generation.  
Their policy samples latent tokens to drive a physics controller over extended horizons, producing idle and ambient behaviors without user oversight.  
The study underscores the advantage of discrete symbol spaces for RL exploration and highlights VQ-VAE priors as a compelling alternative to continuous latent dynamics.

\paragraph{Trajectory / Keyframe / Constraint-based Control \& In-betweening}
Precise spatial control requires the generator to satisfy partial trajectories while retaining naturalness.  
\textit{TLControl}~\citep{wan2024tlcontrol} learns six part-specific codebooks that respect human skeletal topology.  
Given sparse trajectory and textual cues, a Masked Trajectory Transformer first predicts a coarse sequence of latent tokens; a subsequent optimization phase then refines the tokens to better satisfy user constraints.  
The two-stage design leverages the edit ability of discrete latents: coarse prediction is feed-forward and fast, while gradient-based refinement exploits the non-differentiable codebook through straight-through estimators, achieving accurate yet expressive in-betweening.

\paragraph{Text-to-Motion Generation}
\textit{MoGenTS}~\citep{yuan2024mogents} reformulates motion as a two-dimensional spatial-temporal token map, enabling the use of image-style 2D convolutions and masking strategies.  
By applying attention along both temporal and skeletal axes, the model captures local joint correlations and long-range choreography simultaneously.  
This spatially structured quantization illustrates how VQ-VAEs can bridge the gap between video/image and motion domains, opening avenues for cross-modal pre-training and efficient token compression.

\paragraph{Physics-aware / Physics-based Motion Generation}
\textit{Bae et al.}~\citep{bae2025versatile} introduce a hybrid latent space that blends continuous embeddings with discrete VQ tokens, serving as a flexible prior for downstream physics-based or RL generation.  
The continuous component affords fine-grained interpolation, while the discrete part supplies strong high-level regularization; together they improve convergence speed and physical plausibility when synthesizing motions that must satisfy dynamic constraints or target locations.  
This work demonstrates that VQ-VAE latents can seamlessly integrate with control policies, paving the way for physically grounded yet diverse motion generation.

Across these studies, VQ-VAE frameworks prove effective at distilling complex motion into compact, editable, and semantically meaningful token sequences.  
Whether employed for unconditional synthesis, trajectory-guided editing, text-driven animation, or physics-based control, the discrete latent representation offers an appealing balance between expressiveness and tractability within the broader landscape of motion generation.
\begin{table*}[ht]
	\renewcommand{\arraystretch}{1.25}
	\resizebox{\textwidth}{!}{
		% \begin{tabular}{l|c|c|c|c}
            \begin{tabular}{lcc >{\centering\arraybackslash}p{3.0cm} c}
			\toprule
                \multicolumn{1}{c}{{\textbf{Method}}} & 
                \multicolumn{1}{c}{{\textbf{Primary Architecture}}} & 
                \multicolumn{1}{c}{{\textbf{Condition}}} & 
                \multicolumn{1}{c}{{\textbf{Dataset}}} & 
                \multicolumn{1}{c}{{\textbf{Category}}} \\	                
    		\midrule
			
              Li et al. (vq)~\citep{li2024controlling}
            & Linear
            & $\varnothing$
            & \citep{aist++}, \citep{ferstl2018investigating}, \citep{li2024controlling}
            & Unconditional or Minimal-data Motion Generation \\

              TLControl~\citep{wan2024tlcontrol} 
            & Attention - Convolution
            & Text - Trajectory
            & \citep{guo2022generating}, \citep{kit-ml}  
            & Trajectory / Keyframe / Constraint-based Control \& In-betweening \\

              MoGenTS~\citep{yuan2024mogents}
            & Attention - Convolution
            & Text
            & \citep{guo2022generating}, \citep{kit-ml}
            & Text-to-Motion Generation \\

              Bae et al.~\citep{bae2025versatile}
            & Linear
            & Poses - Time - Trajectory 
            & \citep{harvey2020robust}
            & Physics-aware / Physics-based Motion Generation \\

            \bottomrule
		\end{tabular}}
	\vspace{0.5\baselineskip}
	\caption{Papers employing \textbf{vector-quantized variational autoencoders (VQ-VAEs)} to discretize motion data, enabling token-based generation and efficient compression. Entries are grouped by research category.}
	\label{tab:vqvae}
\end{table*}

\subsection{\textbf{Continuous Autoregressive Motion Generation}}\label{sec:continuous-AR}
Continuous autoregressive (AR) models generate motion frame-by-frame, conditioning every newly synthesized pose on a short history of previously generated poses.  
By explicitly modeling local dependencies, such models naturally capture fine-grained temporal dynamics, offer extremely low inference latency, and allow the generator to be seamlessly steered online by external control signals.  
The approach, therefore, forms a practical backbone for interactive applications such as real-time animation, virtual-reality avatar control, and keyframe in-betweening.  
A concise overview of the AR papers discussed in this subsection is provided in Table~\ref{tab:continuous_ar}.

\paragraph{Text-to-Motion Generation}
Mapping free-form language to continuous body motion is challenging because the motion length and kinematic details are only implicitly specified in text.  
\textit{Guo et al. (car)}~\citep{guo2022generating} tackle this under an autoregressive lens by decomposing the problem into two sequential stages: \emph{text2length} first samples a distribution over plausible sequence lengths, after which \emph{text2motion} employs a temporal variational autoencoder that autoregressively emits poses of the chosen length.  
The explicit length modeling resolves the inherent one-to-many ambiguity of language and permits diverse, sentence-consistent motions within the AR decoding framework.

\paragraph{Action-conditioned Motion Generation}
When the conditioning signal is a coarse action label or sparse sensory cue, the generator must extrapolate detailed full-body dynamics.  
\textit{Action2Motion}~\citep{guo2020action2motion} combines a Lie-algebra pose representation with a conditional temporal VAE whose prior is recursively updated from already synthesized poses, enabling physically coherent sequences that respect the target action category.  
\textit{HMD-NeMo}~\citep{aliakbarian2023hmd} extends the setting to virtual-reality input, where only 6-DoF head and hand tracks are available and may be intermittently missing.  
A spatio-temporal encoder fortified with mask tokens compensates for dropped observations, and two separate AR decoders predict body pose and global trajectory, jointly achieving real-time, full-body avatar animation.

\paragraph{Trajectory / Keyframe / Constraint-based Control \& In-betweening}
Producing stylistically consistent motion that exactly hits user-specified keyframes remains a central need in digital content creation.  
\textit{RSMT}~\citep{tang2023rsmt} meets this need by autoregressively generating in-between poses that are guided by the start-end keyframes, the allotted duration, and a desired action style.  
Its recurrent architecture is carefully regularized to avoid collapsing to neutral motion, thereby maintaining high-fidelity stylistic characteristics even with limited data.  
Focusing on physical plausibility, \textit{Li et al. (car)}~\citep{li2024foot} incorporate foot-contact states into a foot-constrained spatio-temporal transformer and reconstruct the root trajectory in a differentiable post-process, effectively eliminating foot-sliding artifacts while still operating in an AR decoding regime.

\paragraph{Motion Prediction, Completion \& Autoregressive Control}
Beyond generation from scratch, AR models excel at rolling out future frames conditioned on a short prefix.  
\textit{MANN}~\citep{zhang2018mode} adopts a Mixture-of-Experts formulation in which a lightweight gating network blends multiple trajectory-conditioned experts at each time step, capturing the multi-modal nature of quadruped locomotion and enabling fast, data-efficient forecasting that can run interactively in simulation loops.

\paragraph{Scene \& Object-interaction Motion Generation}
Finally, the challenge of producing intent-driven motion in cluttered scenes is addressed by \textit{IMoS}~\citep{ghosh2023imos}.  
Starting from textual instructions that mention objects and desired actions, the framework encodes intent, leverages dual conditional VAEs for body and object dynamics, and autoregressively refines object poses to maintain realistic contact.  
The model advances the state of the art on intent-conditioned human–object interaction tasks, demonstrating the versatility of AR decoders when complemented with scene-level reasoning modules.

Collectively, these studies illustrate the breadth of continuous AR approaches: from language grounding and sparse sensor completion to keyframe in-betweening, motion forecasting, and scene interaction, the sequential nature of AR decoding provides a unifying mechanism for generating temporally coherent, controllable, and computationally efficient motion.
\begin{table*}[ht]
	\renewcommand{\arraystretch}{1.25}
	\resizebox{\textwidth}{!}{
		% \begin{tabular}{l|c|c|c|c}
            \begin{tabular}{lcc >{\centering\arraybackslash}p{3.0cm} c}
			\toprule
                \multicolumn{1}{c}{{\textbf{Method}}} & 
                \multicolumn{1}{c}{{\textbf{Primary Architecture}}} & 
                \multicolumn{1}{c}{{\textbf{Condition}}} & 
                \multicolumn{1}{c}{{\textbf{Dataset}}} & 
                \multicolumn{1}{c}{{\textbf{Category}}} \\	                
    		\midrule

                      MANN~\citep{zhang2018mode}
    			  & Linear
                    & Action - Trajectory
                    & \citep{zhang2018mode}
                    & Prediction / Completion \& Autoregressive Control\\
                    
                      Action2Motion~\citep{guo2020action2motion}
                    & RNN
                    & Action
                    & \citep{cmu},  \citep{guo2020action2motion}, \citep{shahroudy2016ntu}
                    & Action-conditioned Motion Generation\\
                    
                      Guo et al. (car)~\citep{guo2022generating}
                    & RNN
                    & Text
                    & \citep{guo2022generating}, \citep{kit-ml}
                    & Text-to-Motion Generation\\
                    
                      IMoS~\citep{ghosh2023imos}
                    & Attention
                    & Action - Poses
                    & \citep{grab}
                    & Scene \& Object-interaction Motion Generation\\
                    
                      RSMT~\citep{tang2023rsmt}
                    & Attention - Convolution - RNN
                    & Action - Poses - Time
                    & \citep{mason2022real}
                    & Trajectory / Keyframe / Constraint-based Control \& In-betweening\\
                    
                      HMD-NeMo~\citep{aliakbarian2023hmd}
                    & Attention - RNN
                    & Action
                    & \citep{mahmood2019amass}
                    & Action-conditioned Motion Generation\\
                    
                      Li et al. (car)~\citep{li2024foot}
                    & Attention
                    & Poses - Trajectory
                    & \citep{aristidou2017emotion}, \citep{harvey2020robust}, \citep{zhou2018auto}
                    & Trajectory / Keyframe / Constraint-based Control \& In-betweening \\

                \bottomrule
		\end{tabular}}
	\vspace{0.5\baselineskip}
	\caption{Papers implementing \textbf{continuous autoregressive} models that predict motion frame-by-frame, conditioned on previous outputs. Entries are listed in chronological order.}
	\label{tab:continuous_ar}
\end{table*}

\subsection{\textbf{Discrete Autoregressive Motion Generation}}\label{sec:discrete-AR}
Discrete autoregressive (AR) methods first quantize continuous pose sequences into a finite set of symbols, typically via a VQ-VAE or related vector-quantization scheme, and then model the resulting token sequence with an AR language model such as a GRU or transformer.  
This design enjoys several advantages: (i) operating in a discrete space removes the need to learn fine-grained geometric detail inside the generative model, (ii) it allows direct reuse of powerful NLP architectures, masking objectives, and hierarchical token schedules, and (iii) the symbolic representation is easily editable, permitting fine-grained control and plug-and-play conditioning.  
Table~\ref{tab:discrete_ar} summarizes the literature reviewed below.

\paragraph{Text-to-Motion Generation}
Pioneering discrete AR approaches, \textit{TM2T}~\citep{guo2022tm2t} trains a VQ-VAE on motion and a GRU‐based neural machine translator to map text embeddings to motion tokens, while also supporting the reverse motion-to-text task.  
\textit{T2M-GPT}~\citep{zhang2023generating} replaces the GRU with a GPT-style transformer and demonstrates that large-scale language-model pretraining principles transfer to motion token sequences.  
\textit{AttT2M}~\citep{zhong2023attt2m} further disentangles word-level and sentence-level semantics through dual attention paths and a body-part VQ-VAE, improving local action fidelity.  
To alleviate quantization loss, \textit{MoMask}~\citep{guo2024momask} adopts residual VQ codebooks and masked transformers for token prediction, yielding sharper reconstructions.  
\textit{LaMP}~\citep{li2024lamp} jointly pretrains language-informed motion and motion-informed language embeddings via contrastive and bidirectional objectives before fine-tuning a masked transformer, thereby replacing CLIP with motion-specific alignments.  
Finally, \textit{HGM$^{3}$}~\citep{jeonghgm3} introduces hierarchical VQ-VAEs and \emph{hard-token mining} so that generation proceeds from coarse to fine semantic levels (motion $\rightarrow$ actions $\rightarrow$ specifics), pushing tokenized pipelines to unprecedented granularity.

\paragraph{Action-conditioned Motion Generation}
\textit{PoseGPT}~\citep{lucas2022posegpt} extends the VQ-VAE+GPT recipe to condition on action labels, desired duration, and motion history, and crucially operates on full 3D meshes rather than skeletal joints, illustrating that discrete AR modeling scales to higher-dimensional surface data.

\paragraph{Physics-aware / Physics-based Motion Generation}
\textit{MoConvVQ}~\citep{yao2024moconvq} augments the token decoder with a differentiable \emph{world model} that approximates physics simulation.  
The resulting T2M-MoConGPT generator, therefore, outputs token sequences that are immediately converted into physically plausible motions. 
% \textcolor{red}{, marking the first physics-based discrete AR framework for general text-to-motion synthesis}.

\paragraph{Motion Editing \& Fine-grained Controllability}
\textit{MMM}~\citep{pinyoanuntapong2024mmm} shows that editable motion can be achieved by masking subsets of tokens such as temporal spans, body parts, or text-aligned segments and regenerating only the masked portions, enabling fast local edits.  
\textit{CoMo}~\citep{huang2024controllable} goes further by designing body-part-level pose codes and leveraging large language models to interpret iterative editing instructions, thereby supporting joint-wise and frame-wise refinement through natural language.

\paragraph{Large Multi-modal / Multi-task Foundation Models}
\textit{MotionGPT}~\citep{jiang2023motiongpt} popularizes the view of "motion as a foreign language", introducing prompt-based learning on top of a motion tokenizer.  After tokenizing the motion and text, they use motion-language pre-training and prompt fine-tuning.
\textit{MotionChain}~\citep{jiang2024motionchain} adds conversational, multi-turn capabilities by fusing text, images, and prior motion tokens for long-horizon generation.  
\textit{M$^{3}$GPT}~\citep{luo2024m} expands the input and output modalities to include text, music, and motion. Unlike earlier methods, it jointly trains the language model and the motion de-tokenizer, optimizing the model in both the discrete semantic space and the continuous motion space.
\textit{MotionLLM}~\citep{wu2024motion} enlarges an LLM's vocabulary with motion tokens, enabling captioning, dialogue, and editing without specialized datasets.  
\textit{VersatileMotion}~\citep{ling2024motionllama} combines flow matching with VQ-VAE to improve the decoding of quantized codes and employs a LoRA-fine-tuned LLaMA to enable multimodal understanding across text, audio, and motion, highlighting the potential of discrete autoregressive tokens as a unified representation for foundation models.

\paragraph{Unconditional or Minimal-data Motion Generation}
\textit{MotionDreamer}~\citep{wang2025motiondreamer} tackles the extreme regime of generating from a \emph{single} demonstration.  
A codebook regularizer prevents collapse, while sliding-window masked transformers model both local patterns and inter-window transitions, proving that discrete AR models can synthesize diverse, long sequences from minimal data.

Across these studies, discrete autoregressive frameworks evolve from modest GRU translators to sophisticated, multimodal transformers with hierarchical codebooks and physics integration.  
Their symbolic representation not only enhances sample quality and edit ability but also opens the door to seamless integration with contemporary language and foundation-model pipelines, establishing discrete AR modeling as a key pillar of modern motion generation research.
\begin{table*}[ht]
	\renewcommand{\arraystretch}{1.25}
	\resizebox{\textwidth}{!}{
		% \begin{tabular}{l|c|c|c|c}
            \begin{tabular}{lcc >{\centering\arraybackslash}p{3.0cm} c}
			\toprule
                \multicolumn{1}{c}{{\textbf{Method}}} & 
                \multicolumn{1}{c}{{\textbf{Primary Architecture}}} & 
                \multicolumn{1}{c}{{\textbf{Condition}}} & 
                \multicolumn{1}{c}{{\textbf{Dataset}}} & 
                \multicolumn{1}{c}{{\textbf{Category}}} \\	                
    		\midrule
			
              TM2T~\citep{guo2022tm2t}
            & Attention - Convolution - RNN
            & Text
            & \citep{guo2022generating}, \citep{kit-ml}
            & Text-to-Motion Generation\\
            
              PoseGPT~\citep{lucas2022posegpt}
            & Attention
            & Action - Poses - Time
            & \citep{grab}, \citep{guo2020action2motion}, \citep{punnakkal2021babel} 
            & Action-conditioned Motion Generation\\
            
              T2M-GPT~\citep{zhang2023generating}
            & Attention - Convolution
            & Text
            & \citep{guo2022generating}, \citep{kit-ml}
            & Text-to-Motion Generation\\
            
              MotionGPT~\citep{jiang2023motiongpt}
            & Attention - Convolution
            & Text
            & \citep{guo2022generating}, \citep{kit-ml}
            & Large Multi-modal / Multi-task Foundation Models\\
            
              AttT2M~\citep{zhong2023attt2m}
            & Attention - Convolution
            & Text
            & \citep{guo2022generating}, \citep{kit-ml}
            & Text-to-Motion Generation\\
            
              MoConvVQ~\citep{yao2024moconvq}
            & Attention - Convolution
            & Action - Text
            & \citep{harvey2020robust}, \citep{mahmood2019amass} 
            & Physics-aware / Physics-based Motion Generation\\
            
              MoMask~\citep{guo2024momask}
            & Attention - Convolution
            & Text
            & \citep{guo2022generating}, \citep{kit-ml}
            & Text-to-Motion Generation\\
            
              MMM~\citep{pinyoanuntapong2024mmm}
            & Attention
            & Poses - Text
            & \citep{guo2022generating}, \citep{kit-ml}
            & Motion Editing \& Fine-grained Controllability\\
            
              CoMo~\citep{huang2024controllable}
            & Attention - Convolution
            & Poses - Text
            & \citep{guo2022generating}, \citep{kit-ml}
            & Motion Editing \& Fine-grained Controllability\\
            
              MotionChain~\citep{jiang2024motionchain}
            & Attention - Convolution
            & Image - Poses - Text - Video
            & \citep{bedlam}, \citep{guo2022generating}, \citep{mahmood2019amass},  \citep{punnakkal2021babel}
            & Large Multi-modal / Multi-task Foundation Models\\

              M$^3$ GPT~\citep{luo2024m}
            & Attention - Convolution
            & Audio - Poses - Text
            & \citep{aist++}, \citep{li2023finedance}, \citep{lin2023motion} 
            & Large Multi-modal / Multi-task Foundation Models \\
            
              MotionLLM~\citep{wu2024motion}
            & Attention - Convolution
            & Poses - Text
            & \citep{guo2022generating}, \citep{kit-ml}
            & Large Multi-modal / Multi-task Foundation Models\\
            
              LaMP~\citep{li2024lamp}
            & Attention - Convolution
            & Text
            & \citep{guo2022generating}, \citep{kit-ml}
            & Text-to-Motion Generation\\
            
              VersatileMotion~\citep{ling2024motionllama}
            & Attention - Convolution
            & Audio - Poses - Text
            & \citep{ling2024motionllama}
            & Large Multi-modal / Multi-task Foundation Models\\
            
              HGM$^3$~\citep{jeonghgm3}
            & Attention
            & Text
            & \citep{guo2022generating}, \citep{kit-ml}
            & Text-to-Motion Generation\\
            
              MotionDreamer~\citep{wang2025motiondreamer}
            & Attention - Convolution
            & $\varnothing$
            & \citep{mixamo}, \citep{truebones}
            & Unconditional or Minimal-data Motion Generation\\
            \bottomrule
		\end{tabular}}
	\vspace{0.5\baselineskip}
	\caption{Papers adopting \textbf{discrete autoregressive} models, typically following vector quantization of continuous motion data. Entries are grouped by research category.}
	\label{tab:discrete_ar}
\end{table*}

\subsection{\textbf{Generative Adversarial Network–based Motion Generation}}\label{sec:GAN}
Generative Adversarial Networks (GANs) formulate motion synthesis as a two-player game in which a generator strives to produce realistic pose sequences while a discriminator learns to detect artifacts.  
This adversarial learning paradigm offers three key advantages for motion modeling: (i) it yields high-fidelity samples without requiring explicit likelihoods, (ii) it readily incorporates task-specific losses or multiple discriminators to enforce spatial-temporal plausibility, and (iii) it can be combined with recurrent, graph, or transformer backbones to address diverse conditioning modalities.  
Table~\ref{tab:gan} summarizes the GAN-based works reviewed below.

\paragraph{Text-to-Motion Generation}
Pioneering the intersection of natural language and GANs, \textit{Text2Action}~\citep{ahn2018text2action} employs a Seq2Seq architecture whose generator is an RNN that maps sentence embeddings to pose trajectories, while an adversarial discriminator enforces realism.
% \textcolor{red}{, establishing the first proof-of-concept that free-form text can drive motion synthesis. } 
Recent progress emphasizes richer multimodal context: \textit{Kim et al.}~\citep{kim2024body} integrate large-language-model parsing of audiovisual cues with gesture-phasing theory, and train a multimodal GAN that produces fluid full-body gestures aligned with conversational speech.  
Together, these works demonstrate the evolution from language-only descriptions toward holistic, dialogue-aware non-verbal behavior generation.

\paragraph{Action-conditioned Motion Generation}
When an action label is provided, the challenge shifts to generating diverse yet category-consistent motions.  
\textit{SA-GAN}~\citep{yu2020structure} introduces a self-attention mechanism that learns a sparse, adaptive graph of temporal dependencies and employs both frame-level and sequence-level discriminators to balance local accuracy and global realism.  
\textit{Kinetic-GAN}~\citep{degardin2022generative} further unifies graph convolutional networks with GAN training, augmenting diversity via latent-space disentanglement and scaling to more than one hundred action classes.  
\textit{ActFormer}~\citep{xu2023actformer} brings transformers into the adversarial framework and incorporates a Gaussian-process prior to model long-range temporal correlations.
Notably, this method handles multi-person interactions through shared latent representation and an interaction-specific Transformer encoder, marking a first step toward GAN-based group motion coordination.

\paragraph{Motion Prediction, Completion \& Autoregressive Control}
For short-term forecasting, \textit{Chen et al.}~\citep{chen2024rethinking} propose the Symplectic Integral Neural Network (SINN).  
By encoding poses on a symplectic manifold and stacking Hamiltonian-inspired residual blocks, the generator maintains numerical stability over extended roll-outs, while an adversarial objective ensures realism.  
The approach illustrates how physically grounded inductive biases can be married to GAN training to improve both accuracy and long-horizon robustness.

\paragraph{Unconditional or Minimal-data Motion Generation}
Finally, \textit{MoDi}~\citep{raab2023modi} demonstrates that adversarial learning can uncover structured latent spaces even without external conditioning.  
Leveraging 3D skeleton-aware convolutions, the model clusters semantically related motions in its latent manifold and produces long, temporally coherent sequences despite heterogeneous training data, highlighting the capacity of GANs to learn rich priors from unlabeled motion corpora.

Taken together, GAN-based approaches have progressed from early RNN generators to sophisticated graph and transformer architectures, addressing text grounding, action diversity, physically plausible prediction, and data-efficient unconditional synthesis.  
% While diffusion and flow-matching methods have recently gained traction, adversarial training remains a competitive option, particularly when sample fidelity, latent manipulation, or real-time generation with compact networks is paramount.
\begin{table*}[ht]
	\renewcommand{\arraystretch}{1.25}
	\resizebox{\textwidth}{!}{
		% \begin{tabular}{l|c|c|c|c}
            \begin{tabular}{lcc >{\centering\arraybackslash}p{3.0cm} c}
			\toprule
                \multicolumn{1}{c}{{\textbf{Method}}} & 
                \multicolumn{1}{c}{{\textbf{Primary Architecture}}} & 
                \multicolumn{1}{c}{{\textbf{Condition}}} & 
                \multicolumn{1}{c}{{\textbf{Dataset}}} & 
                \multicolumn{1}{c}{{\textbf{Category}}} \\	                
    		\midrule
			
    			   Text2Action~\citep{ahn2018text2action}
    			  & RNN
    			& Text 
    			& \citep{xu2016msr}
                & Text-to-Motion Generation \\

    			   SA-GCN~\citep{yu2020structure} 
    			  & Attention - Convolution
    			& Action 
    			& \citep{ionescu2013human3}, \citep{shahroudy2016ntu} 
                & Action-conditioned Motion Generation \\
    
    			   Kinetic-GAN~\citep{degardin2022generative} 
                    & Convolution
    			& Action
    			& \citep{ionescu2013human3}, \citep{liu2019ntu}, \citep{shahroudy2016ntu} 
                & Action-conditioned Motion Generation \\
                    
    			   ActFormer~\citep{xu2023actformer} 
                    & Attention
    			& Action
    			& \citep{liu2019ntu}, \citep{punnakkal2021babel}, \citep{xu2023actformer}
                & Action-conditioned Motion Generation \\
    
    			   MoDi~\citep{raab2023modi} 
                    & Convolution
    			& $\varnothing$
    			& \citep{guo2020action2motion}, \citep{mixamo}
                & Unconditional or Minimal-data Motion Generation \\

                  Chen et al.~\citep{chen2024rethinking}
                & RNN 
                & Poses
                & \citep{cmu}, \citep{ionescu2013human3}, \citep{von2018recovering}
                & Motion Prediction / Completion \& Autoregressive Control \\
                    
                  Kim et al.~\citep{kim2024body}
                & Attention
                & Audio - Text
                & \citep{kim2024body}
                & Text-to-Motion Generation \\
    
                \bottomrule
		\end{tabular}}
	\vspace{0.5\baselineskip}
	\caption{Papers using \textbf{generative adversarial networks (GANs)} to synthesize realistic motion by training against a discriminator. Entries are listed in chronological order.}
	\label{tab:gan}
\end{table*}

\subsection{\textbf{Diffusion-based Motion Generation}}\label{sec:diffusion}
Diffusion models have rapidly become a dominant paradigm for motion synthesis.  
By learning to iteratively denoise samples drawn from a simple prior distribution, these models provide (i) strong mode coverage that captures the high variability of articulated motion, (ii) a likelihood proxy that enables principled conditioning through classifier-free or guidance techniques, and (iii) modular neural architectures that can be combined with transformers, GANs, or physics simulators.  
Across a wide spectrum of tasks ranging from text-to-motion to physics-aware control, researchers have adopted diffusion to push the frontier of realism, diversity, and controllability.  
An overview of the works reviewed in this subsection is summarized in Table~\ref{tab:diffusion}.

\paragraph{Text-to-Motion Generation}
Text-to-motion generation aims to synthesize realistic and expressive motion sequences conditioned on textual descriptions. The adoption of diffusion models has significantly advanced this field, offering powerful generative capabilities. 

Among the pioneers, \textit{MotionDiffuse}~\citep{zhang2024motiondiffuse} introduced the first diffusion model-based framework for text-driven motion generation, notable for its use of linear attention for efficiency and its ability to control motion generation for different body parts and handle sequential actions. Following this, \textit{MDM}~\citep{tevethuman} proposed predicting the clean sample ($\mathbf{x}_0$) directly at each diffusion step, which facilitated the integration of geometric losses like foot contact. \textit{MoFusion}~\citep{dabral2023mofusion} further explored diffusion for conditional human motion synthesis, including music-to-dance, utilizing a lightweight 1D U-Net and incorporating kinematic losses with a time-varying weight schedule to improve realism. Seeking to enhance diversity, \textit{ReMoDiffuse}~\citep{zhang2023remodiffuse} introduces a retrieval-augmented generation framework. The method begins by retrieving reference motions from a database based on both semantic and kinematic similarity. It then leverages information from the retrieved motions to generate motion sequences that are semantically consistent with the input. A different approach to leverage data was taken by \textit{Make-An-Animatoin}~\citep{azadi2023make}, which trained a two-stage model by first using a large dataset of text-pseudo-pose pairs extracted from image-text datasets, then fine-tuning on motion capture data. To better capture intricate linguistic details, \textit{Fg-T2M}~\citep{wang2023fg} proposes a fine-grained motion generation framework that incorporates linguistic structure and progressive reasoning within a diffusion-based model. In contrast to prior methods that process text holistically, Fg-T2M employs a Linguistic Structure-Assisted Module (LSAM) to decompose textual input and a Context-Aware Progressive Reasoning (CAPR) module to iteratively refine the generated motion, enabling the model to capture both local and global semantic cues. \textit{GraphMotion}~\citep{jin2023act} transforms motion descriptions into hierarchical semantic graphs with three levels including: motions, actions, and specifics, and aligns each level with a corresponding transformer-based denoising model within a three-stage diffusion process.

Exploring modularity, \textit{OMG}~\citep{liang2024omg} presented a framework using a Mixture of Controllers guided by open-vocabulary text for diverse and plausible motions. \textit{CrossDiff}~\citep{ren2024realistic} learned a shared latent space between 3D and 2D projected motions to generate more expressive outputs and learn from 2D data. A multi-step generation process was proposed by \textit{PRO-Motion}~\citep{liu2024plan}, which uses an LLM (GPT) to first generate pose descriptions for a \emph{Posture-Diffusion} module, which transforms the descriptions into poses, optimizing for likely pose transitions.  Finally, a third diffusion module links the key poses together into a motion sequence. 
To address the limitations of coarse and ambiguous textual annotations in motion models, \textit{MotionGPT\_1}~\citep{ribeiro2024motiongpt_1} 
extends MDM~\citep{tevethuman} with a doubly text-conditioned training strategy using both high-level descriptions from HumanML3D~\citep{guo2022generating} and low-level annotations from BABEL~\citep{punnakkal2021babel}. To enhance motion diversity and improve alignment with training data, they incorporate zero-shot GPT-3 prompting during inference to generate finer-grained motion details. To address the challenge of generating seamless compositions from a series of varying textual descriptions, unlike previous methods that often produce isolated motions limited to short durations, \textit{FlowMDM}~\citep{barquero2024seamless} introduces Blended Positional Encodings (BPE), which combine absolute and relative positional information during the denoising process. This strategy promotes global motion coherence in the early stages and ensures smooth, realistic transitions in the later stages of denoising. Adapting alternative architectures, \textit{Motion Mamba}~\citep{zhang2024motion} explored using State Space Models (Mamba) for long sequence modeling with Hierarchical Temporal Mamba and Bidirectional Spatial Mamba. An inference-time technique to decompose text into simpler phrases using ChatGPT and combine generated motions was presented by \textit{MCD}~\citep{mandelli2024generation}, noted for its compatibility with any diffusion model. %These diverse approaches highlight the rapid evolution and various directions being explored within diffusion-based text-to-motion generation.

\paragraph{Action-conditioned Motion Generation}
For scenarios where the control signal is a discrete action or sparse user input, diffusion models must reconcile high-level intent with low-level kinematics.  
\textit{AAMDM}~\citep{li2024aamdm} adopts a two-stage pipeline that produces a coarse latent motion via a diffusion-GAN denoiser and refines it with a few extra steps, achieving interactive rates.  
\textit{CAMDM}~\citep{chen2024taming} integrates past motion and trajectory cues with the given action within a transformer diffusion backbone, yielding diverse futures that respond to real-time character-control commands.

\paragraph{Trajectory / Keyframe / Constraint-based Control and In-betweening}
\textit{GMD}~\citep{karunratanakul2023guided} shows that dense guidance and emphasis projection can embed trajectory, obstacle, and key-frame constraints into the denoising loop.  
\textit{DiffKFC}~\citep{wei2024enhanced} tackles text-driven in-betweening with dilated mask attention that accommodates variable keyframe spacing, while \textit{CondMDI}~\citep{cohan2024flexible} presents a unified framework that flexibly incorporates sparse keyframes, partial poses, and optional text cues at inference time. This is achieved through a masked conditional diffusion model, where portions of the input motion are masked, and the model is trained to denoise and reconstruct the missing parts. These works demonstrate that diffusion naturally supports hard spatial constraints and user-authored timing.

\paragraph{Motion Prediction, Completion \& Autoregressive Control}
\textit{HumanMAC}~\citep{chen2023humanmac} casts motion prediction as masked completion in the Discrete Cosine Transform domain, enabling also motion completion and body-part controllable prediction.
\textit{A-MDM}~\citep{shi2024interactive} replaces the typical space-time diffusion with an autoregressive, frame-wise MLP denoiser for real-time rollout.  
\textit{RoHM}~\citep{zhang2024rohm} introduces dual diffusion networks for global trajectory and local pose, achieving robust reconstruction from noisy or missing inputs.

\paragraph{Stylised Motion and Style Transfer}
\textit{SMooDi}~\citep{zhong2024smoodi} adapts ControlNet to inject style exemplars into a pre-trained text-to-motion diffuser, producing diverse motions that inherit the appearance of a reference clip while following novel textual commands.

\paragraph{Physics-aware / Physics-based Motion Generation}
Bridging the gap between kinematic synthesis and physical feasibility, \textit{PhysDiff}~\citep{yuan2023physdiff} interleaves a physics simulator within each denoising step to correct artifacts online.  
\textit{CLoSD}~\citep{tevet2024closd} runs an autoregressive diffusion planner in a closed loop with an RL controller for interactive, goal-directed behavior.  
\textit{RobotMDM}~\citep{serifi2024robot} fine-tunes text-driven motion diffusion models to produce physics-aware motions by training a reward surrogate that ensures generated motions are feasible and compatible with reinforcement learning-based robot controllers, whereas \textit{PARC}~\citep{xu2025parc} combines a kinematic motion generator with a physics-based reinforcement learning tracker for terrain-aware motion synthesis; the generator proposes trajectories, while the RL tracker refines them for physical plausibility, enabling iterative self-improvement. \textit{Rocca et al.}~\citep{rocca2025policy} propose a diffusion-based method that generates compact neural control policies in weight space, enabling efficient adaptation to new characters and environments without retraining.

\paragraph{Scene \& Object-interaction Motion Generation}
\textit{NIFTY}~\citep{kulkarni2024nifty} introduces neural interaction fields learned from synthetic interaction data to condition diffusion on object proximity.  
\textit{ADM}~\citep{wang2024move} leverages affordance maps extracted from 3D scenes, which describe the potential actions that the environment offers for interaction, while \textit{Cen et al.}~\citep{cen2024generating} combine ChatGPT-based scene graphs with trajectory-first diffusion, yielding context-aware motions that respect target objects and spatial layouts.

\paragraph{Skeleton-agnostic or Mesh-based Motion Generation}
Moving beyond fixed rig topologies, \textit{SMD}~\citep{xue2025shape} employs spectral mesh encoding to diffuse directly in mesh space, maintaining shape identity via a decoupled conditioning pathway.  
\textit{AnyTop}~\citep{gat2025anytop} introduces enrichment and skeletal-temporal transformer blocks that support non-homeomorphic skeletons by utilizing text embeddings of joint names. Expanding on the concept of skeleton-agnostic generation, \textit{UniMoGen}~\citep{khani2025unimogen} adapts this approach to action-conditioned settings through an efficient, topology-agnostic diffusion framework that incorporates both spatial and temporal attention mechanisms.

\paragraph{Fast / Few-step Sampling \& Real-time Diffusion}
To reduce the typical hundreds of denoising iterations, \textit{EMDM}~\citep{zhou2024emdm} combines a diffusion GAN with geometric losses, achieving generation times below 50 milliseconds for both action- and text-controlled sequences without sacrificing quality.

\paragraph{Large Multi-modal / Multi-task Foundation Models}
\textit{LMM}~\citep{zhang2024large} unifies sixteen motion datasets and trains a transformer diffuser with ArtAttention to handle ten tasks spanning text, audio, and video conditioning, illustrating that diffusion can serve as the generative core of large, versatile motion foundation models.

\paragraph{Motion Editing \& Fine-grained Controllability}
\textit{FLAME}~\citep{kim2023flame} proposes joint text-to-motion generation and editing by appending language, length, and timestep tokens to the motion sequence.  
\textit{OmniControl}~\citep{xie2023omnicontrol} introduces spatio-temporal control tokens that let users pin individual joints at arbitrary times, while \textit{FineMoGen}~\citep{zhang2023finemogen} employs mixture-of-experts attention to satisfy part-level textual edits.  
\textit{AutoKeyframe}~\citep{autokeyframe_sig25} looks ahead to skeleton-based gradient guidance that confines changes to selected keyframes, promising precise local adjustments.

\paragraph{Unconditional or Minimal-data Motion Generation}
Finally, \textit{SinMDM}~\citep{raabsingle} shows that even a single motion clip suffices: by limiting the receptive field through query-narrow attention, the model avoids overfitting and can synthesize arbitrarily long, diverse sequences, highlighting the data efficiency of diffusion when architectural inductive biases are carefully designed.

Together, these studies demonstrate the flexibility of diffusion models: a single probabilistic framework can accommodate diverse conditioning signals, rigorous physical constraints, topology-agnostic representations, and even real-time requirements, making diffusion a cornerstone of modern motion generation research.
\begin{table*}[ht]
	\renewcommand{\arraystretch}{1.25}
	\resizebox{\textwidth}{!}{
		% \begin{tabular}{l|c|c|c|c}
            \begin{tabular}{lcc >{\centering\arraybackslash}p{3.0cm} c}
			\toprule
                \multicolumn{1}{c}{{\textbf{Method}}} & 
                \multicolumn{1}{c}{{\textbf{Primary Architecture}}} & 
                \multicolumn{1}{c}{{\textbf{Condition}}} & 
                \multicolumn{1}{c}{{\textbf{Dataset}}} & 
                \multicolumn{1}{c}{{\textbf{Category}}} \\	                
    		\midrule
                      MotionDiffuse~\citep{zhang2024motiondiffuse}
                    & Attention
                    & Action - Text
                    & \citep{guo2020action2motion}, \citep{guo2022generating}, \citep{ji2018large}, \citep{kit-ml}
                    & Text-to-Motion Generation \\
                    
                      FLAME~\citep{kim2023flame}
                    & Attention
                    & Text
                    & \citep{guo2022generating}, \citep{kit-ml}, \citep{punnakkal2021babel}
                    & Motion Editing \& Fine-grained Controllability\\
                    
                      MDM~\citep{tevethuman}
                    & Attention
                    & Action - Text
                    & \citep{guo2020action2motion}, \citep{guo2022generating}, \citep{ji2018large}, \citep{kit-ml}
                    & Text-to-Motion Generation\\
                    
                      PhysDiff~\citep{yuan2023physdiff}
                    & Architecture-agnostic
                    & Action - Text
                    & \citep{guo2020action2motion}, \citep{guo2022generating},  \citep{ji2018large}
                    & Physics-aware / Physics-based Motion Generation\\
                    
                      MoFusion~\citep{dabral2023mofusion}
                    & Attention - Convolution
                    & Audio - Text
                    & \citep{guo2022generating}, \citep{punnakkal2021babel}
                    & Text-to-Motion Generation\\
                    
                      HumanMAC~\citep{chen2023humanmac}
                    & Attention
                    & Poses
                    & \citep{cmu}, \citep{ionescu2013human3}, \citep{mahmood2019amass}, \citep{sigal2010humaneva}  
                    & Motion Prediction / Completion \& Autoregressive Control\\
                    
                      SinMDM~\citep{raabsingle}
                    & Attention - Convolution
                    & $\varnothing$
                    & \citep{guo2022generating}, \citep{mixamo}, \citep{raabsingle}, \citep{truebones}, 
                    & Unconditional or Minimal-data Motion Generation\\

                      ReMoDiffuse~\citep{zhang2023remodiffuse}
                    & Attention
                    & Poses - Text
                    & \citep{guo2022generating}, \citep{kit-ml}
                    & Text-to-Motion Generation \\
                    
                      Make-An-Animatoin~\citep{azadi2023make}
                    & Attention - Convolution
                    & Text
                    & \citep{azadi2023make}, \citep{mahmood2019amass} 
                    & Text-to-Motion Generation \\
                    
                      GMD~\citep{karunratanakul2023guided}
                    & Convolution
                    & Obstacle - Poses - Text - Trajectory
                    & \citep{guo2022generating}
                    & Trajectory / Keyframe / Constraint-based Control \& In-betweening\\
                    
                      A-MDM~\citep{shi2024interactive}
                    & Linear
                    & Poses
                    & \citep{harvey2020robust}, \citep{mahmood2019amass}, \citep{mason2022real} 
                    & Motion Prediction / Completion \& Autoregressive Control\\
                    
                      NIFTY~\citep{kulkarni2024nifty}
                    & Attention
                    & Obstacles - Poses - Scene
                    & \citep{kulkarni2024nifty}
                    & Scene \& Object-interaction Motion Generation\\
                    
                      Fg-T2M~\citep{wang2023fg}
                    & Attention - Convolution
                    & Text
                    & \citep{guo2022generating}, \citep{kit-ml} 
                    & Text-to-Motion Generation \\
                    
                      OmniControl~\citep{xie2023omnicontrol}
                    & Attention
                    & Poses - Text - Trajectory
                    & \citep{guo2022generating}, \citep{kit-ml}
                    & Motion Editing \& Fine-grained Controllability\\
                    
                      GraphMotion~\citep{jin2023act}
                    & Attention
                    & Text
                    & \citep{guo2022generating}, \citep{kit-ml}
                    & Text-to-Motion Generation\\
                    
                      AAMDM~\citep{li2024aamdm}
                    & Linear
                    & Action
                    & \citep{harvey2020robust}
                    & Action-conditioned Motion Generation\\
                    
                      EMDM~\citep{zhou2024emdm}
                    & Attention
                    & Action - Text
                    & \citep{guo2020action2motion}, \citep{guo2022generating}, \citep{kit-ml} 
                    & Fast / Few-step Sampling \& Real-time Diffusion\\
                    
                      OMG~\citep{liang2024omg}
                    & Attention - Convolution
                    & Text
                    & \citep{aist++}, \citep{araujo2023circle}, \citep{beat}, \citep{cai2022humman}, \citep{egobody}, \citep{grab}, \citep{guo2022generating}, \citep{harvey2020robust}, \citep{ionescu2013human3}, \citep{liang2023hybridcap}, \citep{liang2024intergen}, \citep{mahmood2019amass}, \citep{mason2022real}, \citep{trumble2017total}
                    & Text-to-Motion Generation \\
                    
                      CrossDiff~\citep{ren2024realistic}
                    & Attention
                    & Text
                    & \citep{guo2022generating}, \citep{kit-ml}, \citep{ucf101}
                    & Text-to-Motion Generation\\
                    
                      FineMoGen~\citep{zhang2023finemogen}
                    & Attention
                    & Text
                    & \citep{guo2022generating}, \citep{kit-ml}, \citep{zhang2023finemogen}
                    & Text-to-Motion Generation\\
                    
                      PRO-Motion~\citep{liu2024plan}
                    & Attention
                    & Text
                    & \citep{delmas2022posescript}, \citep{guo2022generating}, \citep{lin2023motion}, \citep{mahmood2019amass}   
                    & Motion Editing \& Fine-grained Controllability\\

                      MotionGPT\_1~\citep{ribeiro2024motiongpt_1}
                    & Attention
                    & Text
                    & \citep{guo2022generating}, \citep{punnakkal2021babel}
                    & Text-to-Motion Generation\\
                    
                      RoHM~\citep{zhang2024rohm}
                    & Attention - Convolution
                    & Poses
                    & \citep{egobody}, \citep{hassan2019resolving}, \citep{mahmood2019amass}  
                    & Motion Prediction / Completion \& Autoregressive Control\\
                    
                      FlowMDM~\citep{barquero2024seamless}
                    & Attention
                    & Text
                    & \citep{guo2022generating}, \citep{punnakkal2021babel}
                    & Text-to-Motion Generation\\
                    
                      Motion Mamba~\citep{zhang2024motion}
                    & Attention - Convolution
                    & Text
                    & \citep{guo2022generating}, \citep{kit-ml}
                    & Text-to-Motion Generation\\
                    
                      DiffKFC~\citep{wei2024enhanced}
                    & Attention
                    & Poses - Text
                    & \citep{guo2022generating}, \citep{kit-ml}
                    & Trajectory / Keyframe / Constraint-based Control \& In-betweening\\
                    
                      ADM~\citep{wang2024move}
                    & Attention
                    & Scene - Text
                    & \citep{guo2022generating}, \citep{wang2022humanise}
                    & Scene \& Object-interaction Motion Generation\\

                      LMM~\citep{zhang2024large}
                    & Attention
                    & Action - Audio - Poses - Text
                    & \citep{zhang2024large}
                    & Large Multi-modal / Multi-task Foundation Models\\
                    
                      CAMDM~\citep{chen2024taming}
                    & Attention
                    & Action - Poses - Trajectory
                    & \citep{mason2022real}
                    & Action-conditioned Motion Generation\\
                    
                      SMD~\citep{xue2025shape}
                    & Attention - Convolution
                    & Action - Poses - Text
                    & \citep{guo2022generating}, \citep{punnakkal2021babel}
                    & Skeleton-agnostic or Mesh-based Motion Generation\\

                      Cen et al.~\citep{cen2024generating}
                    & Attention
                    & Poses - Text
                    & \citep{hassan2019resolving}, \citep{mahmood2019amass}, \citep{wang2022humanise}
                    & Scene \& Object-interaction Motion Generation\\
                    
                      CondMDI~\citep{cohan2024flexible}
                    & Convolution
                    & Poses - Text - Trajectory
                    & \citep{guo2022generating}
                    & Trajectory / Keyframe / Constraint-based Control \& In-betweening\\
                    
                      SMooDi~\citep{zhong2024smoodi}
                    & Attention
                    & Poses - Text
                    & \citep{guo2022generating}, \citep{mason2022real}
                    & Stylized Motion \& Style Transfer\\
                    
                      MCD~\citep{mandelli2024generation}
                    & Architecture-agnostic
                    & Text
                    & \citep{guo2022generating}, \citep{kit-ml}
                    & Text-to-Motion Generation\\
                    
                      CLoSD~\citep{tevet2024closd}
                    & Attention
                    & Poses - Text - Trajectory
                    & \citep{guo2022generating}
                    & Physics-aware / Physics-based Motion Generation\\
                    
                      RobotMDM~\citep{serifi2024robot}
                    & Attention
                    & Text
                    & \citep{guo2022generating}
                    & Physics-aware / Physics-based Motion Generation\\

                      AnyTop~\citep{gat2025anytop}
                    & Attention
                    & Skeleton
                    & \citep{truebones}
                    & Skeleton-agnostic or Mesh-based Motion Generation\\

                      PARC~\citep{xu2025parc}
                    & Architecture-agnostic
                    & $\varnothing$
                    & \citep{xu2025parc}
                    & Physics-aware / Physics-based Motion Generation \\
                    
                      Rocca et al.~\citep{rocca2025policy}
                    & Attention
                    & Action
                    & \citep{harvey2020robust}
                    & Physics-aware / Physics-based Motion Generation\\
                    
                      UniMoGen~\citep{khani2025unimogen}
                    & Attention - Convolution
                    & Action - Poses - Trajectory
                    & \citep{harvey2020robust}, \citep{mason2022real}
                    & Skeleton-agnostic or Mesh-based Motion Generation\\
                    
                      AutoKeyframe~\citep{autokeyframe_sig25}
                    & Attention
                    & Action - Poses - Trajectory
                    & unknown
                    & Motion Editing \& Fine-grained Controllability \\

                \bottomrule
		\end{tabular}
        }
	\vspace{0.5\baselineskip}
	\caption{Papers applying \textbf{diffusion} probabilistic models directly in motion space to iteratively refine samples from noise. Entries are listed in chronological order.}
	\label{tab:diffusion}
\end{table*}

\subsection{\textbf{Latent–Diffusion–based Motion Generation}}\label{sec:latent-diffusion}
Latent diffusion models (LDMs) consist of autoencoding and denoising: a variational (or vector‐quantized) autoencoder first compresses motion sequences into a compact latent space, after which a diffusion process is learned and executed entirely in that space.  
This separation brings two key benefits for motion generation: (i) low per-step dimensionality drastically reduces training and sampling cost and (ii) the autoencoder can inject structural priors (e.g., disentangled content or discrete codebooks) that facilitate controllability. %\textcolor{red}{and (iii) standard diffusion tool-kits such as classifier-free guidance, consistency distillation, or ControlNet can be readily ported to the motion domain.}  
Table~\ref{tab:latent_diffusion} summarizes the LDM literature reviewed in this subsection.

\paragraph{Text-to-Motion Generation}
\textit{MLD}~\citep{chen2023executing} initiates the paradigm by training a VAE whose latent trajectories are denoised with a standard DDPM, showing that operating at 1/16\textsuperscript{th} of the original dimensionality preserves perceptual quality while accelerating inference.  
Subsequent works enrich the latent representation.  
\textit{M2DM}~\citep{kong2023priority} tokenizes motion using a Transformer-based VQ-VAE and applies a discrete diffusion model instead of the commonly used continuous ones. This model prioritizes denoising based on token importance, which is estimated either statically (via entropy) or dynamically (via a learned agent), enabling the reconstruction of more informative tokens earlier in the generation process.
More recently, \textit{GuidedMotion}~\citep{karunratanakul2023guided} proposed a local action-guided diffusion model for controllable text-to-motion generation that decomposes motion descriptions into multiple local actions, allowing diverse sampling to match user preferences. It employs graph attention networks to weigh local motions during synthesis and uses a three-stage diffusion process to ensure stable generation aligned with both global and local motion cues.
\textit{LADiff}~\citep{sampieri2024length} makes sequence length an explicit conditioning variable; by mapping short sequences to a dedicated low-rank subspace it adapts the diffusion dynamics to durations ranging from a few beats to several seconds, yielding richer, length-aware motion styles.
Looking ahead, work like \textit{Meng et al.}~\citep{meng2024rethinking} explores combining diffusion models with masked token prediction without using a tokenizer. The approach first encodes motion frames using an autoencoder, and then employs a transformer for masked autoregressive prediction. Then, the transformer's output is used as a condition for a diffusion model to generate the final latent vector, which will be passed to the decoder of the autoencoder.
Finally, in contrast to other methods, \textit{ACMDM}~\citep{meng2025absolute} investigates using absolute joint positions, instead of relative joint positions, as input.

\paragraph{Motion Prediction, Completion \& Autoregressive Control}
\textit{BeLFusion}~\citep{barquero2023belfusion} targets short-term prediction with a focus on smooth transitions from past to future motion: a behavior module disentangles "what" (behavior) from "how" (pose) in the latent space, and a latent diffuser samples futures that remain consistent with the recent past.  
\textit{DartControl}~\citep{zhao2024dartcontrol} extends this idea to interactive control by conditioning the latent primitives jointly on motion history and free-form text; thanks to an aggressive ten-step schedule, it delivers near-real-time responses without quality degradation.

\paragraph{Stylized Motion \& Style Transfer}
\textit{MCM-LDM}~\citep{song2024arbitrary} is a latent diffusion model that disentangles and integrates content, style, and trajectory information from motion clips using prioritized multi-condition guidance. Operating in a compact latent space, it preserves motion trajectory to enable smooth and realistic style transfers while maintaining the original motion’s structure and intent.
\textit{MulSMo}~\citep{li2024mulsmo} generalizes style control to multimodal cues, such as text, motion, or image, via contrastive alignment and introduces motion-aligned temporal latent diffusion (MaTLD) for smoother temporal consistency.  
Both studies illustrate how the autoencoder stage of LDMs enables explicit factorization of stylistic attributes.

\paragraph{Fast / Few-step Sampling \& Real-time Diffusion}
\textit{MotionLCM}~\citep{dai2024motionlcm} adopts the \emph{consistency-model} framework: a single network is distilled from a pre-trained motion latent diffusion model to map noisy latents directly to clean ones at arbitrary noise levels.  
Combined with a ControlNet branch trained in both latent and motion space, the system achieves one- or few-step sampling, reducing generation time to real-time while retaining spatial-temporal controllability.

\paragraph{Large Multi-modal / Multi-task Foundation Models}
\textit{UDE}~\citep{zhou2023ude} builds a full stack of four modules: a VQ-VAE encoder, a Modality-Agnostic Transformer Encoder that brings text and audio into the same token space, a Unified Token Transformer for sequence modeling, and a Diffusion Motion Decoder that refines the quantized tokens.  
The architecture demonstrates that latent diffusion can scale to heterogeneous inputs and tasks without task-specific decoders.

\paragraph{Motion Editing \& Fine-grained Controllability}
\textit{UniTMGE}~\citep{wang2025unitmge} aligns motion and CLIP text latents via a learnable projection, enabling arithmetic operations in the shared space, for instance, adding a textual concept to an existing motion.  
\textit{LEAD}~\citep{andreou2025lead} presents a text-to-motion model based on latent diffusion that improves semantic alignment between motion and language representations by introducing a projector module that maps VAE latents into a CLIP-compatible space. The module, implemented as an autoencoder, enhances both motion generation and a new task called Motion Textual Inversion (MTI), which aims to learn language embeddings from a few motion examples for exemplar-consistent generation. The approach enables semantically grounded motion synthesis and improves generalization from limited data.

Collectively, these studies show that latent diffusion offers an attractive compromise between expressiveness and efficiency.  
By leveraging compact representations, LDMs inherit the diversity and controllability of full-space diffusion while opening the door to real-time sampling, multi-modal conditioning, and fine-grained latent manipulations.
\begin{table*}[ht]
	\renewcommand{\arraystretch}{1.25}
	\resizebox{\textwidth}{!}{
		% \begin{tabular}{l|c|c|c|c}
            \begin{tabular}{lcc >{\centering\arraybackslash}p{3.0cm} c}
			\toprule
                \multicolumn{1}{c}{{\textbf{Method}}} & 
                \multicolumn{1}{c}{{\textbf{Primary Architecture}}} & 
                \multicolumn{1}{c}{{\textbf{Condition}}} & 
                \multicolumn{1}{c}{{\textbf{Dataset}}} & 
                \multicolumn{1}{c}{{\textbf{Category}}} \\	                
    		\midrule
			
                 BeLFusion~\citep{barquero2023belfusion}
               & Attention - Convolution - RNN
               & Poses
               & \citep{ionescu2013human3}, \citep{mahmood2019amass}
               & Motion Prediction / Completion \& Autoregressive Control \\

                 UDE~\citep{zhou2023ude}
               & Attention
               & Audio - Text
               & \citep{aist++}, \citep{guo2022generating} 
               & Large Multi-modal / Multi-task Foundation Models \\

                 MLD~\citep{chen2023executing}
               & Attention
               & Action - Text
               & \citep{guo2020action2motion}, \citep{guo2022generating}, \citep{ji2018large}, \citep{kit-ml}  
               & Text-to-Motion Generation \\

                 M2DM~\citep{kong2023priority}
               & Attention
               & Text
               & \citep{guo2022generating}, \citep{kit-ml}
               & Text-to-Motion Generation \\

                 MotionLCM~\citep{dai2024motionlcm}
               & Attention
               & Poses - Text - Trajectory
               & \citep{guo2022generating}
               & Fast / Few-step Sampling \& Real-time Diffusion \\

                 MCM-LDM~\citep{song2024arbitrary}
               & Attention
               & Poses - Trajectory
               & \citep{guo2022generating}
               & Stylized Motion \& Style Transfer \\

                 GuidedMotion~\citep{jin2024local}
               & Attention
               & Text
               & \citep{guo2022generating}, \citep{kit-ml}
               & Text-to-Motion Generation\\

                 LADiff~\citep{sampieri2024length}
               & Attention
               & Text - Time
               & \citep{guo2022generating}, \citep{kit-ml}
               & Text-to-Motion Generation \\

                 DartControl~\citep{zhao2024dartcontrol}
               & Attention
               & Poses - Text
               & \citep{guo2022generating}, \citep{punnakkal2021babel}
               & Motion Prediction / Completion \& Autoregressive Control \\

                 UniTMGE~\citep{wang2025unitmge}
               & Attention
               & Poses - Text
               & \citep{guo2022generating}, \citep{lin2023motion}
               & Motion Editing \& Fine-grained Controllability \\

                 LEAD~\citep{andreou2025lead}
               & Attention
               & Text
               & \citep{guo2022generating}, \citep{kit-ml}, \citep{mason2022real}
               & Motion Editing \& Fine-grained Controllability \\

                 Meng et al.~\citep{meng2024rethinking}
               & Attention - Convolution
               & Text
               & \citep{guo2022generating}, \citep{kit-ml}
               & Text-to-Motion Generation \\

                 MulSMo~\citep{li2024mulsmo}
               & Attention
               & Image - Poses - Text
               & \citep{guo2022generating}, \citep{kit-ml}, \citep{mason2022real}, \citep{xia2015realtime} 
               & Stylized Motion \& Style Transfer \\

                 ACMDM~\citep{meng2025absolute}
               & Attention - Convolution
               & Text
               & \citep{guo2022generating}, \citep{kit-ml}
               & Text-to-Motion Generation \\
            \bottomrule
		\end{tabular}}
	\vspace{0.5\baselineskip}
	\caption{Papers performing \textbf{latent diffusion} in a compressed latent space (e.g., using VAE or encoder-based representations), enhancing efficiency and sample quality. Entries are grouped by research category.}
	\label{tab:latent_diffusion}
\end{table*}

\subsection{\textbf{Flow-Matching–based Motion Generation}}\label{sec:flow-matching}
Flow matching (FM) is an emerging class of score‐based generative methods that learns an explicit continuous mapping or ``flow'' between a simple reference distribution and complex data manifolds.  
Compared with diffusion models, FM eliminates stochastic sampling schedules and instead integrates an ordinary differential equation conditioned on a learned velocity field, thereby achieving high sample quality with significantly fewer integration steps.  
Because the velocity field can be conditioned on arbitrary external information, FM provides a principled and computationally efficient backbone for multi-task motion generation.  
The only work that employs this approach so far is listed in Table~\ref{tab:flow_matching}.

\paragraph{Large Multi-modal / Multi-task Foundation Models}
The versatility of flow matching is epitomized by \textit{MotionLab}~\citep{guo2025motionlab}, which formulates a generic \emph{Motion-Condition-Motion} (MCM) transformation: given a \emph{source} motion and a heterogeneous conditioning signal (e.g., free-form text, sparse keyframes, or trajectories), the model integrates a learned velocity field to reach a \emph{target} motion that satisfies the constraints.  
This unified formulation allows a single network to tackle seemingly disparate tasks, including text-to-motion generation, trajectory-guided editing, and style transfer without resorting to task-specific heads or specialized loss terms.  
Practically, \textit{MotionLab} demonstrates that FM can match or surpass diffusion baselines while reducing inference costs, highlighting the promise of flow matching as a scalable foundation model for controllable, high-fidelity motion synthesis.

\begin{table*}[ht]
	\renewcommand{\arraystretch}{1.25}
	\resizebox{\textwidth}{!}{
		% \begin{tabular}{l|c|c|c|c}
            \begin{tabular}{lcc >{\centering\arraybackslash}p{3.0cm} c}
			\toprule
                \multicolumn{1}{c}{{\textbf{Method}}} & 
                \multicolumn{1}{c}{{\textbf{Primary Architecture}}} & 
                \multicolumn{1}{c}{{\textbf{Condition}}} & 
                \multicolumn{1}{c}{{\textbf{Dataset}}} & 
                \multicolumn{1}{c}{{\textbf{Category}}} \\	                
    		\midrule
			
    			   MotionLab~\citep{guo2025motionlab} 
    			  & Attention
    			& Poses - Text - Trajectory
    			& \citep{athanasiou2024motionfix}, \citep{guo2022generating} 
                    & Large Multi-modal / Multi-task Foundation Models \\ %\cdashline{2-5}[2pt/2pt]
                \bottomrule
		\end{tabular}}
	\vspace{0.5\baselineskip}
	\caption{The only paper utilizing \textbf{flow-matching} to learn deterministic mappings between noise and data through continuous dynamics.}
	\label{tab:flow_matching}
\end{table*}

\subsection{\textbf{Implicit Neural–Representation Motion Generation}}\label{sec:inr}
Implicit neural representations (INRs) parametrize a time‐varying signal as a continuous function $f_\theta:\mathbb{R}\!\to\!\mathbb{R}^{J\!\times\!3}$, typically realized by a small multi-layer perceptron that maps a real‐valued timestamp to $J$ joint positions or rotations.  
This coordinate‐based encoding dispenses with fixed frame rates and sequence lengths, brings sub–frame temporal resolution, and enables direct evaluation, interpolation, or differentiation at arbitrary times.  
Consequently, INR-based generators provide a compact and analytically convenient alternative to discrete autoregressive or diffusion models, particularly suitable for applications that demand variable‐length synthesis or continuous‐time control.  
An overview of existing INR works is given in Table~\ref{tab:inr}.

\paragraph{Action-conditioned Motion Generation}
The first explicit exploration of INRs for human motion is due to \textit{Cervantes et al.}~\citep{cervantes2022implicit}.  
Their framework decomposes the continuous motion function into three latent components: an \emph{action} code that captures class semantics, a \emph{sequence} code that models instance-level stylistic variability, and a \emph{temporal} code that modulates fine-grained timing.  
A variational objective is used to learn a distribution over these latents, while a shared MLP decoder reconstructs joint trajectories of arbitrary duration by querying the concatenated latent and timestamp.  
Because synthesis consists of sampling latent vectors and evaluating the INR at any desired set of times, the method naturally generates motions of unbounded length, offers smooth temporal derivatives, and avoids cumulative error typical of frame-wise autoregression.  
This study demonstrates that INRs can match the realism of conventional sequence models while providing unique advantages in temporal flexibility and memory efficiency, thereby spotlighting implicit representations as a promising direction for action-conditioned motion generation.
\begin{table*}[ht]
	\renewcommand{\arraystretch}{1.25}
	\resizebox{\textwidth}{!}{
		% \begin{tabular}{l|c|c|c|c|c}
		\begin{tabular}{lcc >{\centering\arraybackslash}p{3.0cm} c}
			\toprule
                \multicolumn{1}{c}{{\textbf{Method}}} & 
                \multicolumn{1}{c}{{\textbf{Primary Architecture}}} & 
                \multicolumn{1}{c}{{\textbf{Condition}}} & 
                \multicolumn{1}{c}{{\textbf{Dataset}}} & 
                \multicolumn{1}{c}{{\textbf{Category}}} \\	                
    		\midrule
			
    			   Cervantes et al.~\citep{cervantes2022implicit}
    			& Attention
    			& Action 
    			& \citep{cervantes2022implicit}, \citep{guo2020action2motion}, \citep{ji2018large}
                    & Action-conditioned Motion Generation \\ %\cdashline{2-5}[2pt/2pt]
                    
                \bottomrule
		\end{tabular}}
	\vspace{0.5\baselineskip}
	\caption{The sole paper in our survey that utilizes \textbf{implicit neural representations}.
% \textcolor{red}{Some of these are not discussed in the text?}
}
	\label{tab:inr}
\end{table*}
\subsection{\textbf{Physics/Optimization-Based Motion Generation}}\label{sec:physics-based}
Physics-based methods for motion generation aim to produce realistic and physically plausible motion by explicitly incorporating physical principles such as dynamics, kinematics, contact forces, and momentum conservation. Alternatively, optimization-based methods are computational approaches that formulate a problem as the minimization or maximization of an objective function, often subject to constraints. Please refer to Table~\ref{tab:others} for a summary of the methods.
Please note that, unlike purely data-driven approaches, these methods often rely on numerical solvers, soft and hard constraints, or differentiable physics models to ensure that generated motions respect the laws of motion and exhibit natural behaviors such as balance, smooth transitions, and ground interaction. In physics-based learning, physical properties such as energy conservation, force balance, or contact consistency are integrated into the training process by adding them as constraints or penalties in the loss function. These properties are quantified by evaluating how closely the model's outputs align with established physical laws, rules, or heuristics based on prior knowledge.
% The resulting physics-based loss is combined with standard objectives, like reconstruction or prediction error, to form a total loss.

\paragraph{Motion Prediction, Completion \& Autoregressive Control}
\textit{AuxFormer}~\citep{xu2023auxiliary} enhances skeleton-based human motion prediction by incorporating auxiliary tasks that provide additional learning signals related to motion dynamics and physical constraints. These tasks help the model better capture the underlying biomechanics and physical plausibility of human movement, enabling it to generate more realistic and coherent motion sequences. In the context of motion completion and prediction, this approach allows the system to fill in missing parts or forecast future poses in a way that respects natural motion patterns and related physical properties. Furthermore, by using an autoregressive control framework, the model generates motion incrementally, conditioning each new prediction on previous states and auxiliary cues. This iterative process helps maintain temporal consistency, stability, and realism throughout the motion sequence, resulting in smooth and physically plausible outputs.

\paragraph{Physics-aware / Physics-based Motion Generation}
\textit{PhysMoP}~\citep{zhang2024incorporating} anchors motion progression in physically grounded dynamics such as momentum and acceleration. To further improve realism and prevent drifting from the physically plausible solution over time, their method aggregates these physics-driven predictions with a data-driven model, adaptively weighting both to correct deviations. 

\paragraph{Scene \& Object-interaction Motion Generation}
\textit{Zhao et al.}~\citep{zhao2023synthesizing} propose an optimization-driven approach that leverages knowledge of 3D indoor environments and object properties alongside physical constraints to create realistic and physically plausible human motions. Their approach enforces the preservation of physical components such as balance and contact forces while optimizing trajectories to minimize energy or maximize stability during object manipulation. By incorporating scene context and object affordances, the method synthesizes diverse and physically plausible motions that adapt to different environments and tasks.

\paragraph{Motion Editing \& Fine-grained Controllability}
\textit{Liu et al.}~\citep{liu2024programmable} enables physics-aware motion editing by allowing users to define detailed task objectives through modular constraints, such as maintaining keyframes, following trajectories, or preserving physical balance. Instead of directly generating motion, the system performs optimization in the latent space of a pre-trained diffusion model to satisfy these constraints. This approach allows precise control over motion outputs while retaining realism, making it well-suited for generating physically plausible and finely tuned motion across a wide range of control scenarios. Augmenting these principles through virtual reality, \textit{Masked Humanoid Controller (MHC)}~\citep{shrestha2024generating} further advances fine-grained, physics-based motion editing by enabling users to input partial or multi-modal motion directives, such as VR controller signals, or video cues. To convert these inputs into physically realistic motion, MHC is trained via reinforcement learning with a multi-objective reward that jointly optimizes directive adherence, motion plausibility, and energy efficiency. This optimization-driven formulation enables the system to produce responsive and physically coherent motions that remain closely aligned with user intent.

\begin{table*}[ht]
	\renewcommand{\arraystretch}{1.25}
	\resizebox{\textwidth}{!}{
		% \begin{tabular}{l|c|c|c|c|c}
		\begin{tabular}{lccc >{\centering\arraybackslash}p{3.0cm} c}
			\toprule
			\multicolumn{1}{c}{{\textbf{Approach}}} & 
                \multicolumn{1}{c}{{\textbf{Method}}} & 
                \multicolumn{1}{c}{{\textbf{Primary Architecture}}} & 
                \multicolumn{1}{c}{{\textbf{Condition}}} & 
                \multicolumn{1}{c}{{\textbf{Dataset}}} & 
                \multicolumn{1}{c}{{\textbf{Category}}} \\	                
    		\midrule
                    Reinforcement Learning
                    & Zhao et al.~\citep{zhao2023synthesizing}
                    & Linear
                    & Scene
                    & \citep{hassan2019resolving}, \citep{hassan2021stochastic},  \citep{mahmood2019amass}, \citep{straub2019replica}
                    & Scene \& Object-interaction Motion Generation \\
                    
                    Auxiliary Learning
                    & AuxFormer~\citep{xu2023auxiliary}
                    & Attention
                    & Poses
                    & \citep{cmu}, \citep{ionescu2013human3}, \citep{von2018recovering}
                    & Motion Prediction / Completion \& Autoregressive Control \\

                    Physics-based
    			& PhysMoP~\citep{zhang2024incorporating}
                    & Linear
    			& Poses
    			& \citep{ionescu2013human3}, \citep{mahmood2019amass}, \citep{von2018recovering} 
                    & Physics-aware / Physics-based Motion Generation \\

                    Optimization
                    & Programmable Motion Generation~\citep{liu2024programmable}
                    & Attention
                    & Action - Obstacle - Others - Poses - Scene - Text - Trajectory
                    & \citep{guo2022generating}, \citep{punnakkal2021babel}
                    & Motion Editing \& Fine-grained Controllability\\

                    Reinforcement Learning
                    & MHC~\citep{shrestha2024generating}
                    & Linear
                    & Joystick - Poses - Video - VR Controller
                    & \citep{reallusion}
                    & Motion Editing \& Fine-grained Controllability\\
                    
                \bottomrule
		\end{tabular}}
	\vspace{0.5\baselineskip}
	\caption{Papers employing alternative or hybrid approaches beyond major generative families, including rule-based, simulation-based, or optimization-driven techniques. Entries are listed in chronological order.
% \textcolor{red}{Some of these are not discussed in the text?}
}
	\label{tab:others}
\end{table*}

\section{\textbf{Datasets}}
\label{sec:datasets}

In this section, we provide an overview of the datasets used across the literature surveyed in this work. In Table~\ref{tab:datasets}, we present a summary of the statistics of all datasets referenced in the reviewed papers, including size information, annotated attributes, and collection method.
This table is intended as a curated reference to facilitate dataset selection for future research, especially for those developing or benchmarking generative models of motion.

We organize the datasets based on the type of conditioning signal they support: \emph{action labels}, \emph{text descriptions}, \emph{audio or speech}, and \emph{interaction context}. This organization reflects the most common generative tasks in the literature, such as action-to-motion synthesis, text-to-motion generation, and gesture generation from speech.

\subsection{\textbf{Action-Conditioned Datasets}}
These datasets associate motion clips with discrete action labels and are commonly used in action-to-motion generation and recognition tasks. Most of them are acquired via marker-based motion capture, providing high-precision 3D joint data and consistent skeletal formats.

Notable examples include LAFAN~\citep{harvey2020robust} and 100Style~\citep{mason2022real}, which contain short motion segments with labeled styles or activities, enabling training and evaluation of transition-aware models. HumanAct12~\citep{guo2020action2motion} offers a compact set of well-labeled actions suitable for controlled motion generation, while the NTU RGB+D series~\citep{shahroudy2016ntu,liu2019ntu,cervantes2022implicit} provides large-scale action annotations from RGB-D sensors and is often used in classification and generative pipelines. BABEL~\citep{punnakkal2021babel} adds action labels at two levels of abstraction to raw motion clips from AMASS~\citep{mahmood2019amass}, allowing models to learn both high-level categories and fine-grained motions.

\subsection{\textbf{Text-Conditioned Datasets}}
Text-to-motion generation relies on datasets that align free-form language with motion sequences. These datasets are typically built using either marker-based motion capture (e.g., BABEL, KIT-ML) or pose reconstructions from video (e.g., HumanML3D), and vary in vocabulary complexity and annotation style.

HumanML3D~\citep{guo2022generating} is currently the most widely adopted in this space, offering over 14,000 motion-language pairs extracted from various sources and normalized into a common format. KIT-ML~\citep{kit-ml} provides controlled sentence templates paired with motion clips, suitable for training with constrained linguistic input. BABEL~\citep{punnakkal2021babel} supplements action annotations with crowd-sourced natural language descriptions, enabling models to ground free-form input to motion in real-world tasks.

\subsection{\textbf{Audio and Speech-Conditioned Datasets}}
These datasets pair motion with audio signals (speech or music) and support tasks such as gesture generation and dance synthesis. %Collection methods vary: some rely on video recordings with estimated pose sequences, while others use marker-based motion capture for precise alignment.

BEAT~\citep{beat}, TED-Gesture++~\citep{yoon2020speech}, and Gesture Dataset~\citep{kim2024body} are among the most comprehensive for speech-driven motion, combining audio, transcripts, emotion labels, and high-quality pose annotations. Trinity Gesture~\citep{ferstl2018investigating} and Speech2Gesture-3D~\citep{kucherenko2021moving} offer earlier baselines with simpler settings and smaller vocabularies. For music-conditioned generation, AIST++~\citep{aist++}, FineDance~\citep{li2023finedance}, and Dance~\citep{aristidou2017emotion} provide paired motion and music data with rhythm, genre, and tempo annotations. These datasets are typically captured with marker-based systems or reconstructed from performance videos.

\subsection{\textbf{Interaction-Conditioned Datasets}}
Interaction-conditioned datasets involve contextual cues beyond the motion itself, such as objects, scenes, or other agents, and are collected using a mix of marker-based capture and video-based 3D reconstruction. % They support generation of physically and socially grounded behaviors.

EgoBody~\citep{egobody} and PROX~\citep{hassan2019resolving} contain human motion recorded in realistic 3D environments, enabling the study of scene-aware behavior and physical interactions. GRAB~\citep{grab} focuses on whole-body grasping and object manipulation, using marker-based capture and detailed object models. InterHuman~\citep{liang2024intergen} and Inter-X~\citep{xu2024inter} model multi-person interactions with motion and language, while datasets like Circle~\citep{araujo2023circle} include goal-directed motion in interactive scenes. Additionally, Kulkarni et al.\citep{kulkarni2024nifty} synthesized interaction data for sitting and lifting using HuMoR~\citep{rempe2021humor}. These datasets are essential for learning socially intelligent and physically plausible motion patterns.

Finally, some datasets are generated through simulation, game engines, or retargeting tools, enabling large-scale motion generation with specific constraints or stylizations. Although virtual datasets may lack precise realism or biomechanical constraints, these datasets are particularly useful for pretraining or for domains with limited real-world capture data, as in the case of non-human characters.
For example, GTA Combat~\citep{xu2023actformer} leverages motion data extracted from video games to study multi-agent interactions.

\medskip
In summary, the current dataset landscape reflects the growing diversity of generative motion tasks. %From controlled MoCap studios to unconstrained video collections and large-scale synthetic datasets, researchers now have access to a wide variety of motion data sources. 
From categorical action labels to natural language, audio, and physical interaction, researchers now have access to a wide range of motion data sources with varying scale, fidelity, and coverage.
The Table~\ref{tab:datasets} and taxonomy aim to support more informed dataset selection and fairer comparisons across future work.

\begin{table*}[ht]
\centering
% \begin{adjustbox}{max size={\textwidth}{0.95\textheight}}
{\footnotesize
	\renewcommand{\arraystretch}{1.2}
	\resizebox{\textwidth}{!}{
            \begin{tabular}{lcccccccccc}
% \begin{tabular}
% {
% p{2.5cm}  % Dataset
% p{1.0cm}  % Condition
% p{2.0cm}  % Collection Method
% p{1.5cm}  % Representation
% p{0.8cm}  % Subject No.
% p{1.0cm}  % Sequences
% p{0.8cm}  % Frames
% p{0.8cm}  % Duration
% p{1.8cm}  % Annotation
% p{0.6cm}  % Joints No.
% p{0.6cm}  % Action No.
% }
\toprule
\textbf{Dataset} & 
\textbf{Condition} & 
\textbf{Collection Method} & 
\textbf{Representation} & 
\textbf{Subject No.} & 
\textbf{Sequences} & 
\textbf{Frames} & 
\textbf{Duration} & 
\textbf{Annotation} & 
\textbf{Joints No.} & 
\textbf{Action No.} \\
\midrule
AMASS \citep{mahmood2019amass}         & -          & Marker           & Rotation     & 346  & 11451  & -      & 42h      & -                          & 52 & -   \\
CMU \citep{cmu}                        & -          & Marker           & Rotation     & 109  & 2605   & -      & 9h       & -                          & 31 & 23  \\
Mixamo \citep{mixamo}                  & -          & Software         & Rotation     & 95   & 2453   & -      & 6h       & -                     & 22 & -   \\
VTuber-EMOCA \citep{li2024controlling}& -          & Video            & 3D Meshes    & 1    & -      & -      & 46.4h    & -                          & -  & -   \\
MotionVerse \citep{zhang2024large}& -  &  Other datasets  & Keypoint 3D & 1  & 320k & 100M  & - & Action, Music, Speech, Text, Video & 52 & -  \\
100Style \citep{mason2022real}        & Action     & Marker           & Rotation     & 1    & 1372   & 4094607 & -        & Action                     & 25 & 100 \\
Aberman et al. \citep{aberman2020unpaired}        & Action     & Marker           & Keypoint 3D  & 1    & 10500  & -      & -        & Action                     & -  & 16  \\
BABEL \citep{punnakkal2021babel}                   & Action     & Marker           & Rotation     & 346  & 13220  & -      & 43.5h    & Action, Text               & 52 & 260 \\
Dog dataset \citep{zhang2018mode}        & Action     & Marker           & Rotation     & -    & -      & 266K   & 30m      & Action                     & 27 & -   \\
Human3.6M \citep{ionescu2013human3}           & Action     & Marker           & Keypoint 3D  & 11   & -      & 3.6M   & 20h      & Action                     & 32 & 17  \\
HumanAct12 \citep{guo2020action2motion}         & Action     & Video            & Keypoint 3D  & 12   & 1191   & 90099  & 6h       & Action                     & 24 & 12  \\
HumanEva-I \citep{sigal2010humaneva}         & Action     & Marker           & Keypoint 3D  & 4    & 56     & 80000  & -        & Action                     & 15 & 6   \\
HuMMan \citep{cai2022humman}                 & Action     & Video            & Keypoint 3D  & 1000 & 400K   & 60M    & -        & Action                     & 23 & 500 \\
LAFAN \citep{harvey2020robust}                   & Action     & Marker           & Rotation     & 5    & 77     & 496672 & 4.6h     & Action                     & 22 & 15  \\
MPI-INF-3DHP \citep{mehta2017monocular}     & Action     & Video            & Keypoint 3D  & 8    & -      & 1.3M   & -        & Action                     & 15 & 8   \\
NTU RGB+D \citep{shahroudy2016ntu}            & Action     & Video            & Keypoint 3D  & 40   & 56880  & 4M     & -        & Action                     & 25 & 60  \\
NTU-120 RGB+D \citep{liu2019ntu}    & Action     & Video            & Keypoint 3D  & 106  & 114480 & 8M     & -        & Action                     & 25 & 120 \\
NTU13 \citep{cervantes2022implicit}                   & Action     & Video            & Keypoint 3D  & -    & 3900   & -      & -        & Action                     & 18 & 13  \\
Reallusion MoCap \citep{reallusion}   & Action     & Video            & 3D Meshes    & 1    & 87     & -      & -        & Action                     & -  & 87  \\
SAMP \citep{hassan2021stochastic}                     & Action     & Video            & 3D Meshes    & 1    & -      & 185K   & 100m     & Action                     & 22 & -   \\
TotalCapture \citep{trumble2017total}     & Action     & Video            & Keypoint 3D  & 5    & -      & 1.9M   & -        & Action                     & 21 & 5   \\
Truebones Zoo \citep{truebones}       & Action     & Virtual          & Rotation     & 70   & 1219   & 147178 & -        & Action                     & -  & -   \\
UESTC \citep{ji2018large}                   & Action     & Video            & Keypoint 3D  & 118  & 25600  & -      & 83h      & Action                     & 25 & 40  \\
UCF101 \citep{ucf101}                 & Action     & Video            & -            & -    & 13K    & -      & 27h      & Action                     & -  & 101 \\
Xia et al. \citep{xia2015realtime}                       & Action     & -                & Rotation     & -    & 579    & -      & 25m      & Action                     & -  & 8   \\
AIST++ \citep{aist++}                 & Audio      & Video            & Rotation     & 30   & 1408   & -      & 5.2h     & Music                      & 24 & -   \\
Dance \citep{aristidou2017emotion}                   & Audio      & Marker           & Keypoint 3D  & 6    & 108    & 93347  & -        & Music                      & 21 & -   \\
FineDance \citep{li2023finedance}           & Audio      & Marker           & Keypoint 3D  & 27   & -      & -      & 14.6h    & Music                      & 52 & -   \\
HCM \citep{liang2023hybridcap}                       & Audio      & Marker           & Keypoint 3D  & 16   & 100    & 300K   & 2.9h     & Music                      & 25 & -   \\
Trinity Gesture \citep{ferstl2018investigating}& Audio      & Video, Audio, Marker & Rotation   & 1    & 23     & -      & 244m     & Speech                     & -  & -   \\
MotionHub \citep{ling2024motionllama}          & Audio-Text & Aggregated       & Rotation     & -    & 358847 & -      & 596.48h  & Audio, Text                & 22 & -   \\
Beat \citep{beat}                     & Audio-Text & Marker           & Rotation     & -    & 1639   & 18M    & 76h      & Audio, Text, Emotion       & 76 & -   \\
Speech2Gesture-3D \citep{kucherenko2021moving}& Audio-Text & Video      & Keypoint 3D  & -    & 1047   & 1M     & -        & Speech, Text               & 52 & -   \\
TED-Expressive \citep{liu2022learning} & Audio-Text & Video            & Keypoint 3D  & -    & 27221  & 8M     & -        & Speech, Text               & 43 & -   \\
TED-Gesture++ \citep{yoon2020speech}   & Audio-Text & Video            & Keypoint 3D  & 1766 & 34491  & 10M    & 106.1h   & Speech, Text               & 10 & -   \\
Gesture Dataset \citep{kim2024body} & Audio-Gesture & collected from 4 sources & Keypoint 3D & 6 & - & - & 24m & Speech, Gesture & - & - \\
3DPW \citep{von2018recovering}                     & Interaction& Marker, Video   & Keypoint 3D  & 7    & 60     & 51K    & -        & Scene Interaction          & 23 & -   \\
Bedlam \citep{bedlam}                 & Interaction& Video            & 3D Meshes    & 217  & 2311   & 380K   & -        & -                          & -  & -   \\
Circle \citep{araujo2023circle}                 & Interaction& Marker           & Rotation     & 5    & 7K     & 4306K  & 10h      & Virtual Obstacle, Goal     & 22 & -   \\
Egobody \citep{egobody}               & Interaction& Video            & Keypoint 3D  & 36   & 125    & 220K   & -        & Scene Interaction          & 22 & -   \\
GRAB \citep{grab}                     & Interaction& Marker           & 3D Meshes    & 10   & 1334   & 1622459& -        & Object Interaction         & -  & -   \\
GTA Combat \citep{xu2023actformer}          & Interaction& Virtual          & Limb Vectors & -    & 7000   & -      & -        & Action, Human Interaction  & -  & -   \\
PROX \citep{hassan2019resolving}                     & Interaction& Video            & 3D Meshes    & 20   & -      & 100K   & -        & -                          & 55 & -   \\
Synthetic using HuMoR \citep{kulkarni2024nifty} & Interaction & Generated by HuMoR & Keypoint 3D & 14 & 310K & - & - & Scene Interaction & 22 & - \\
Humanise \citep{wang2022humanise}             & Interaction-Text & From AMASS/BABEL & 3D Meshes    & -    & 19.6K  & 1.2M   & -        & Text, Object Interaction   & -  & -   \\
Inter-X \citep{xu2024inter}               & Interaction-Text & Marker           & Keypoint 3D  & -    & 11K    & 8.1M   & -        & Text, Human Interaction    & 55 & -   \\
InterHuman \citep{liang2024intergen}         & Interaction-Text & Video            & Keypoint 3D  & -    & 7779   & 107M   & 6.56h    & Text, Human Interaction    & 23 & -   \\
AnimalML3D \citep{yang2024omnimotiongpt}                 & Text       & DeformingThings4D& Keypoint 3D  & 36   & 1240   & -      & -        & Text                       & 35 & -   \\
HumalML3D \citep{guo2022generating}           & Text       & Marker, Video   & Rotation     & -    & 14616  & -      & 28.59h   & Text                       & 22 & -   \\
HuMMan-MoGen \citep{zhang2023finemogen}     & Text       & From humman      & Keypoint 3D  & -    & 2968   & -      & -        & Text                       & 23 & 160 \\
KIT-ML \citep{kit-ml}                 & Text       & Marker           & Keypoint 3D  & 111  & 3911   & -      & 11.23h   & Text                       & -  & -   \\
Motion-X \citep{lin2023motion}             & Text       & Other datasets   & Keypoint 3D  & -    & 81084  & 15.6M  & 144.2h   & Text                       & 55 & -   \\
MotionFix \citep{athanasiou2024motionfix}           & Text       & Subset of AMASS  & Rotation     & -    & 6730   & -      & -        & Text                       & 22 & -   \\
MSR-VTT \citep{xu2016msr}               & Text       & Video            & -            & -    & -      & -      & 2.71h    & Text                       & -  & -   \\
PoseScript \citep{delmas2022posescript}         & Text       & From AMASS       & Keypoint 3D  & -    & -      & 20K    & -        & Text                       & -  & -   \\
TPP \citep{azadi2023make}                       & Text       & Images           & Rotation     & -    & -      & 35M    & -        & Text                       & -  & -   \\
\bottomrule
\end{tabular}}
}
\caption{Overview of motion datasets used in generative modeling.}\label{tab:datasets}
\end{table*}

\section{\textbf{Evaluation Metrics}}
\label{sec:evaluation_metrics}

Evaluating motion generation shares the broader challenges of generative modeling, such as capturing realism and diversity, while adding the complexity of spatio-temporal dependencies and limited benchmark size compared to fields like text or image generation. In the following, we review the most widely adopted evaluation metrics, grouped by their purpose: realism, diversity, ground-truth comparison, condition consistency, and efficiency. An exhaustive list of the metrics found in the articles reviewed in the present survey is reported in Table \ref{tab:evaluation_metrics}.

\subsection{\textbf{Realism}}

Realism metrics aim to assess the physical plausibility and naturalness of generated motions. This can be done by comparing to a reference motion dataset, or computing physics quantities.

\subsubsection{\textbf{Fidelity}}

The comparison with reference data is often performed in a feature space.
The most common example is the \emph{Fréchet Inception Distance (FID)}, which compares the distribution of features from generated and real motion data. It can also be found under the definition of \emph{Fréchet Motion Distance (FMD)} \citep{rocca2025policy,song2024arbitrary}. Several extensions have been proposed, including FID variants computed on specific body parts (e.g., upper/lower-body FID) \citep{ren2024realistic}, \emph{SiFID}, which evaluates the distribution over sub-windows in the motion \citep{raabsingle}, and \emph{FGD} (Frechet Gesture Distance) for gesture quality in the speech-to-gesture application \citep{ling2024motionllama}.
Similarly, \emph{Kernel Inception Distance (KID)} measures the similarity of the real and generated motion feature distributions using the squared Maximum Mean Discrepancy (MMD) with a polynomial kernel \citep{raab2023modi}.
\emph{Maximum Mean Discrepancy (MMD)} is also used \citep{degardin2022generative, yu2020structure}. Defined as the distance between the embeddings in the kernel Hilbert space (RKHS), it works by mapping the distributions to RKHS, where it measures the difference between their first moments. 

\subsubsection{\textbf{Physical Plausibility}}

To capture physically implausible artifacts, multiple specialized metrics are used. \emph{Foot Sliding Distance} measures the average sliding of the foot joints when contact with the ground is expected, while \emph{Skating} refers to the average foot velocity during contact frames. The variants, \emph{Footskating Frame Ratio (FFR)} and \emph{Foot Sliding Factor (FSF)}, capture the proportion of frames affected \citep{li2024aamdm, song2024arbitrary}. \emph{Penetration} quantifies the average depth of interpenetration of body parts throughout the sequence, with variants of the penetration frequency. Similarly, \emph{floating} is the average distance from the lowest body joint to the ground.

Motion smoothness is assessed using \emph{Maximum Joint Acceleration} or its variant, \emph{Average Acceleration Norm}, to penalize abrupt dynamics. Similarly, \emph{Keyframe Transition Smoothness (K-TranS)} quantifies smoothness around keyframes \citep{wei2024enhanced}. 
To further quantify smoothness, \citep{barquero2024seamless} introduced two novel metrics built upon the concept of jerk (i.e., the time derivative of acceleration), which is indicative of motion smoothness and is known to be sensitive to kinetic irregularities. The \emph{Peak Jerk (PJ)} and \emph{Area Under the Jerk (AUJ)} measure instantaneous and cumulative deviations in jerk, respectively. The \emph{percentage of High Jerk Frames (\%HJF)} is a related metric \citep{xu2025parc}. 
Additionally, \citep{barquero2023belfusion} introduced \emph{Cumulative Motion Distribution (CMD)} to quantify the temporal consistency of predicted motion by measuring deviations in frame-to-frame displacement over time. CMD helps assess whether the generated motion follows a statistically plausible temporal distribution, rather than merely being locally smooth. 

For quadrupeds, leg motion smoothness can be compromised due to poor coordination between the front and back limbs. \emph{Leg stiffness} metric evaluates this by averaging joint angle updates in the rear limbs versus all the limbs \citep{zhang2018mode}. 

When skeleton deformation is possible due to a motion representation based on joint position, it is also crucial to compute the \emph{Bone Length Error} to ensure that the character shape is preserved \citep{shi2024interactive}. 

Finally, in the reinforcement learning-based approach \citep{serifi2024robot}, \emph{Realism Score}, defined via a critic network's accumulated reward, serves as a proxy for motion feasibility.

\subsection{\textbf{Diversity}}

Generative diversity is typically measured by \emph{Diversity}, defined as the average pairwise distance between randomly selected samples, also referred to as \emph{Average Pairwise Distance (APD)} \citep{barquero2023belfusion}. Popular versions of this metric include \emph{inter-diversity} (across samples), \emph{intra-diversity} (within motion sequences), and \emph{intra-diversity difference} (relative to input motion) \citep{raabsingle, wang2025motiondreamer}. \emph{MultiModality (MModality)} assesses diversity under conditional generation, computing the distance between motions generated from the same prompt. The \emph{Coverage} metric evaluates how well the generated set spans the real distribution, using nearest-neighbor thresholds in a feature space.

\subsection{\textbf{Consistency with the Input Condition}}

When it comes to conditional generation, it is also crucial to verify the consistency of generated motions with their intended controls. For this task, several metrics have been proposed that are specific to the input modality.

\subsubsection{\textbf{Text}}
In text-conditioned setups, semantic alignment is key, and several metrics have been proposed to quantify such alignment. They often operate in a joint learned embedding space. The most popular ones are the retrieval-based metrics such as \emph{R-precision}, which measures the proportion of times the correct text is top-ranked for the motion, and \emph{MultiModal Distance (MM-Dist)}, which computes distances in embedding space. The \emph{MotionCLIP Score (mCLIP)} and \emph{Mutual Information Divergence (MID)} assess cross-modal similarity via cosine similarity or mutual information, respectively \citep{tevet2022motionclip}. \emph{Content Encoding Error (CEE)} measures the L2 distance between the content embeddings of text and motion, while \emph{Style Encoding Error (SEE)} compares their Gram matrices, assessing distributional similarity \citep{ghosh2021synthesis}.  The \emph{TMR-M2T} score complements these by evaluating text-to-motion similarity in the learned embedding space of TMR~\citep{petrovich2023tmr}. Finally, \emph{LaMPBertScore} adapts BERTScore by measuring the cosine similarity between generated motion features and text features in the joint embedding space of LaMP~\citep{li2024lamp}.

\subsubsection{\textbf{Action}}

When actions or styles are given, accuracy can be assessed via classifiers trained to detect them. These are reported as \emph{Action Accuracy} in the context of action-to-motion  \citep{guo2020action2motion}, \emph{Style Recognition Accuracy (SRA)} or \emph{Content Recognition Accuracy (CRA)} in the stylization case \citep{zhong2024smoodi, li2024mulsmo, song2024arbitrary}. 
Transition metrics to evaluate responsiveness to style switches are also considered in \citep{chen2024taming}. These are named \emph{Transition Duration} and \emph{Transition Success Rate}.

\subsubsection{\textbf{Trajectory}}

In trajectory-controlled motion generation, \emph{Trajectory Error} is the mean distance to target path points. \emph{Trajectory Orientation Error} measures angular deviation in heading, and the \emph{Trajectory Similarity Index (TSI)} evaluates overall alignment between generated and ground-truth paths \citep{song2024arbitrary}.

\subsubsection{\textbf{Music}}

\emph{Beat Alignment (BA)} is defined as the fraction of predicted motion beats that are within a fixed threshold of ground-truth music beats, indicating how well the generated motions align temporally with the music \citep{zhou2023ude, ling2024motionllama}.

\subsubsection{\textbf{Speech}}

In the speech-to-gesture scenario, analogous rhythmic consistency metrics are considered, like the \emph{Percentage of Matched Beats (PMB)}, which assesses how well the generated motions align with the rhythmic features \citep{kim2024body}, and the aforementioned Beat Alignment. Unlike dance-music alignment, a BA closer to ground truth is preferred over higher values in this case \citep{ling2024motionllama}.

\subsubsection{\textbf{Constraint Satisfaction}}

Metrics in this category assess compliance with spatial constraints. These include contact metrics, distance to objects, \emph{Final Waypoint Distance (FWD)} \citep{xu2025parc}. Under interactive control, \emph{Tracking Error under User Control (TE-UC)} is used \citep{li2024aamdm}.

\subsection{\textbf{Comparison to Ground Truth}}

Comparison with a ground-truth motion sequence is meaningful in supervised or evaluation scenarios where paired real data is available.

To evaluate alignment with a reference motion, the most basic metric is \emph{Average Position Error (APE)}, equivalent to the \emph{Mean Per-Joint Position Error (MPJPE)} or its global variant \emph{GMPJPE} \citep{zhang2024rohm}. A more sophisticated version of the latter is \emph{Procrustes aligned Mean Per Joint Position Error (PA-MPJPE)}, which calculates MPJPE after aligning with GT in translation, rotation, and scale \citep{zhu2023motionbert}. Similarly, \emph{L2P} measures global positional error, while orientation discrepancies are evaluated via \emph{Geodesic Rotation Error (GeoR)}, and \emph{Quaternion L2 distance (L2Q)} is computed in joint rotation space. The \emph{Mean Absolute Joint Errors (MAJE)} gauges the precision of generated joint positions relative to the ground truth across all joints and time steps \citep{kim2024body}. 
\emph{Average Variance Error (AVE)} compares second-order statistics. \emph{Keyframe Generative Error (K-Err)} measures the average discrepancies between the keyframe and the corresponding generated frame \citep{wei2024enhanced}, while \emph{Probability of Correct Keypoints (PCK)} computes how many keypoints fall within a distance threshold \citep{ahuja2019language2pose}. 

Higher order quantities are considered in \emph{Mean Per Joint Velocity Error (MPJVE)} and \emph{Mean Acceleration Differences (MAD)}, or \emph{ACCL} \citep{chen2023executing}, measuring the average velocity and acceleration discrepancies between the generated and reference joint movements, providing insight into the dynamic accuracy of the models \citep{kim2024body}.

Sequence-level accuracy is quantified by \emph{Average Displacement Error (ADE)} and \emph{Final Displacement Error (FDE)}, measuring L2 distances over all timesteps and the final frame, respectively \citep{chen2023humanmac}.

To assess frequency alignment, the \emph{Normalized Power Spectrum Similarity (NPSS)} compares the spectral content of the motion \citep{tang2023rsmt}. 

Finally, in \citep{mandelli2024generation} \emph{TMR-M2M} score is computed using the joint text-motion embedding model \citep{petrovich2023tmr}, quantifying the similarity between generated and ground-truth motions. 

\subsection{\textbf{Efficiency}}

Performance is reported in terms of \emph{Frames Per Second (FPS)}, indicating generation speed, in \citep{li2024aamdm}, and as \emph{Average Inference Time per Sentence (AITS)}, the average time taken to generate a motion for each sentence during inference, in \citep{guo2025motionlab, dai2024motionlcm, chen2023executing}.

% \subsection{Other Metrics}

% Some works report classification-based metrics in latent spaces, such as \emph{Top-1} and \emph{Top-5 Accuracy} using pretrained motion-language models. In stylization settings, \emph{Content Preservation} is evaluated using the geodesic distance of local joint rotations between original and stylized motions.

\subsection{\textbf{User Study}}

Conducting a user study is a reliable way to assess realism and plausibility in complex data such as motion. However, user studies remain limited in practice due to the challenges involved in their setup and execution
\citep{zhao2023synthesizing, kulkarni2024nifty, andreou2025lead, dabral2023mofusion,cen2024generating}.

\subsection{\textbf{Discussion}}

Despite the proliferation of metrics, the evaluation of motion generation remains a challenge. 
FID, though widely adopted, has been criticized for its weak correlation with human judgments and inconsistent implementation across works \citep{tseng2023edge}.
Diversity, while easier to quantify, is hindered by inconsistent definitions, whether based on joint space, embedding space, or sampling strategies. Moreover, classifiers used for evaluation are often retrained ad hoc \citep{kim2023flame, chen2024taming, song2024arbitrary, ling2024motionllama, tevethuman}, which raises concerns over the fairness and reproducibility of reported results. 
This lack of standardization underscores the need for more robust and agreed-upon evaluation protocols in the field.

\begin{table*}[htbp]
\centering
\renewcommand{\arraystretch}{1.15}
	\resizebox{\textwidth}{!}{
\begin{tabular}{p{3.0cm} p{5.2cm} p{8.5cm}}
\toprule
\textbf{Category} & \textbf{Metric} & \textbf{Used in / Introduced by} \\
\midrule

\multirow{10}{*}{Realism - Fidelity}
& Fréchet Inception Distance (FID) / Fréchet Motion Distance (FMD) 
& \citep{ahn2018text2action, andreou2025lead, azadi2023make, barquero2024seamless, cervantes2022implicit, chen2023executing, chen2024taming, cohan2024flexible, dai2024motionlcm, degardin2022generative, gat2025anytop, ghosh2023imos, guo2020action2motion, guo2022tm2t, guo2024momask, guo2025motionlab, huang2024controllable, jeonghgm3, jiang2023motiongpt, jiang2024motionchain, jin2023act, jin2024local, khani2025unimogen, kim2023flame, kong2023priority, lee2023multiact, li2024aamdm, li2024controlling, li2024lamp, li2024mulsmo, liang2024omg, ling2024motionllama, liu2024plan, liu2024programmable, lucas2022posegpt, luo2024m, mandelli2024generation, martinez2017human, meng2024rethinking, meng2025absolute, petrovich2021action, pinyoanuntapong2024mmm, Qian2023BreakingTL, raab2023modi, raabsingle, ren2024realistic, ribeiro2024motiongpt_1, sampieri2024length, serifi2024robot, tan2025think, tevet2024closd, tevethuman, wan2024tlcontrol, wang2024move, wang2025unitmge, wei2024enhanced, wu2024motion, xie2023omnicontrol, xu2023actformer, xue2025shape, yang2024omnimotiongpt, yao2024moconvq, yuan2023physdiff, yuan2024mogents, zhang2023finemogen, zhang2023remodiffuse, zhang2024large, zhang2024motion, zhang2024motiondiffuse, zhao2024dartcontrol, zhong2023attt2m, zhong2024smoodi, zhou2023ude, zhou2024emdm} 
/ \citep{rocca2025policy, song2024arbitrary, tang2023rsmt} \\
& SiFID (Sub-window FID) & \citep{raabsingle} \\
& FID upper/lower body & \citep{ren2024realistic} \\
& Fréchet Gesture Distance (FGD) & \citep{ling2024motionllama} \\
& Kernel Inception Distance (KID) & \citep{raab2023modi} \\
& Maximum Mean Discrepancy (MMD) & \citep{degardin2022generative, yu2020structure} \\

\midrule
\multirow{15}{*}{Realism - Physics}
& Foot Sliding Distance & \citep{chen2024taming, jin2023dafnet, khani2025unimogen, shi2024interactive, xue2025shape} \\
& Skating & \citep{cohan2024flexible, diomataris2024wandr, guo2025motionlab, kulkarni2024nifty, tevet2024closd, xie2023omnicontrol, xue2025shape, zhang2018mode, zhang2024rohm, zhong2024smoodi} \\
& Footskating Frame Ratio (FFR), Foot Sliding Factor (FSF) & \citep{li2024aamdm, song2024arbitrary} \\
& Penetration (depth, frequency) & \citep{li2024aamdm, song2024arbitrary} \\
& Floating & \citep{tevet2024closd,xue2025shape} \\
& Maximum Joint Acceleration, Average Acceleration Norm & \citep{chen2024taming, liu2024programmable, shi2024interactive} \\
& Peak Jerk (PJ), Area Under Jerk (AUJ), \% High Jerk Frames (\%HJF) & \citep{barquero2024seamless, xu2025parc, zhao2024dartcontrol} \\
& Keyframe Transition Smoothness (K-TranS) & \citep{wei2024enhanced} \\
& Cumulative Motion Distribution (CMD) & \citep{barquero2023belfusion} \\
& Leg Stiffness (quadrupeds) & \citep{zhang2018mode} \\
& Bone Length Error & \citep{shi2024interactive} \\
& Realism Score (RL critic) & \citep{serifi2024robot} \\

\midrule
\multirow{8}{*}{Diversity}
& Diversity / Average Pairwise Distance (APD) & \citep{ahn2018text2action, azadi2023make, barquero2024seamless, cervantes2022implicit, chen2023executing, chen2024taming, cohan2024flexible, dabral2023mofusion, dai2024motionlcm, ghosh2023imos, guo2022tm2t, guo2025motionlab, huang2024controllable, jeonghgm3, jiang2023motiongpt, jiang2024motionchain, jin2023act, jin2024local, karunratanakul2023guided, kong2023priority, li2024lamp, li2024mulsmo, liang2024omg, liu2024programmable, ling2024motionllama, luo2024m, lucas2022posegpt, mandelli2024generation, martinez2017human, pinyoanuntapong2024mmm, raab2023modi, ren2024realistic, ribeiro2024motiongpt_1, sampieri2024length, serifi2024robot, shi2024interactive, tan2025think, tang2023rsmt, tevet2024closd, tevethuman, wan2024tlcontrol, wang2023fg, wang2024move, wang2025unitmge, wei2024enhanced, wu2024motion, xie2023omnicontrol, xu2023actformer, xue2025shape, yang2024omnimotiongpt, yuan2024mogents, zhang2023finemogen, zhang2023remodiffuse, zhang2024large, zhang2024motion, zhang2024motiondiffuse, zhao2024dartcontrol, zhong2023attt2m, zhong2024smoodi, zhou2023ude, zhou2024emdm} / \citep{chen2023humanmac, lee2023multiact, shi2024interactive, wang2024move} \\
& Inter / Intra Diversity, Intra Diversity Diff. & \citep{gat2025anytop, khani2025unimogen, raabsingle, wang2025motiondreamer} \\
& MultiModality (MModality) & \citep{andreou2025lead, cervantes2022implicit, chen2023executing, dai2024motionlcm, ghosh2023imos, guo2020action2motion, guo2022tm2t, guo2024momask, jeonghgm3, jiang2023motiongpt, jin2023act, jin2024local, kong2023priority, lee2023multiact, liang2024omg, martinez2017human, meng2024rethinking, meng2025absolute, petrovich2021action, ren2024realistic, ribeiro2024motiongpt_1, serifi2024robot, tevethuman, wan2024tlcontrol, wang2023fg, wang2024move, wu2024motion, yang2024omnimotiongpt, zhang2023remodiffuse, zhang2024large, zhang2024motion, zhang2024motiondiffuse, zhou2023ude, zhou2024emdm} \\
& Coverage & \citep{gat2025anytop, luo2024m, ling2024motionllama, raabsingle, wang2025motiondreamer} \\

\midrule
\multirow{9}{*}{Condition  – Text}
& R-Precision & \citep{andreou2025lead, dabral2023mofusion, dai2024motionlcm, guo2020action2motion, guo2024momask, guo2025motionlab, jeonghgm3, kim2023flame, kong2023priority, liang2024omg, ling2024motionllama, liu2024programmable, luo2024m, martinez2017human, meng2024rethinking, meng2025absolute, ribeiro2024motiongpt_1, serifi2024robot, tevethuman, wan2024tlcontrol, wang2024move, wang2025unitmge, wei2024enhanced, xie2023omnicontrol, xue2025shape, yang2024omnimotiongpt, yao2024moconvq, yuan2023physdiff, zhang2023finemogen, zhang2024motiondiffuse, zhong2023attt2m} \\
& MultiModal Distance (MM-Dist) & \citep{andreou2025lead, chen2023executing, dabral2023mofusion, dai2024motionlcm, guo2024momask, jeonghgm3, kong2023priority, ling2024motionllama, meng2025absolute, ribeiro2024motiongpt_1, serifi2024robot, tan2025think, tevet2022motionclip, wang2024move, yang2024omnimotiongpt} \\
& MotionCLIP Score (mCLIP) & \citep{kim2023flame, tevet2022motionclip} \\
& Mutual Information Divergence (MID) & \citep{kim2023flame} \\
& Content Encoding Error (CEE), Style Encoding Error (SEE) & \citep{ghosh2021synthesis} \\
& TMR-M2T & \citep{mandelli2024generation, petrovich2023tmr} \\
& LaMPBERTScore & \citep{li2024lamp} \\

\midrule
\multirow{4}{*}{Condition  – Action}
& Action Accuracy & \citep{cen2024generating, cervantes2022implicit, chen2024taming, guo2020action2motion, lee2023multiact, petrovich2021action, Qian2023BreakingTL, tevethuman, xu2023actformer, yu2020structure, yuan2023physdiff, zhang2024motiondiffuse, zhu2023motionbert} \\
& Style Recognition Accuracy (SRA) / Content Recognition Accuracy (CRA) & \citep{guo2025motionlab, guo2024generative, li2024mulsmo, song2024arbitrary, zhong2024smoodi} \\
& Transition Duration, Transition Success Rate & \citep{chen2024taming} \\

\midrule
\multirow{2}{*}{Condition  – Trajectory}
& Trajectory Error, Orientation Error, Trajectory Similarity Index (TSI) & \citep{chen2024taming, dai2024motionlcm, khani2025unimogen, song2024arbitrary, wan2024tlcontrol, xie2023omnicontrol} \\

\midrule
\multirow{1}{*}{Condition  – Music}
& Beat Alignment (BA) & \citep{ling2024motionllama, zhou2023ude} \\

\midrule
\multirow{2}{*}{Condition  – Speech}
& Percentage of Matched Beats (PMB), Beat Alignment (BA) & \citep{kim2024body, ling2024motionllama} \\

\midrule
\multirow{3}{*}{Condition  – Constraints}
& Goal Distance & \citep{cen2024generating, diomataris2024wandr, kulkarni2024nifty, wang2024move, zhao2023synthesizing} \\
& Final Waypoint Distance (FWD) & \citep{xu2025parc} \\
& TE-UC & \citep{li2024aamdm} \\

\midrule
\multirow{2}{*}{Efficiency}
& Frames Per Second (FPS) & \citep{li2024aamdm} \\
& Average Inference Time per Sentence (AITS) & \citep{chen2023executing, dai2024motionlcm, guo2025motionlab} \\

\bottomrule
\end{tabular}}
\vspace{10pt}
\caption{Evaluation metrics for motion generation, grouped by category. Each metric is listed with the works where it is used or introduced.}
\label{tab:evaluation_metrics}
\end{table*}

\section{\textbf{Statistical Insights}}
\label{sec:statistical_insights}

To provide a comprehensive understanding of the current landscape in motion generation research, we conducted a statistical analysis of 106 representative papers in the field (which does not include 6 papers in Table~\ref{tab:others}). This analysis examines the distribution of research approaches, architectural choices, dataset preferences, and conditioning strategies employed across the motion generation literature. The insights derived from this analysis reveal important trends and patterns that characterize the evolution and current state of the field.

\subsection{\textbf{Research Approach Distribution}}

The analysis of research approaches reveals the methodological diversity within motion generation research. As shown in Fig.~\ref{fig:approach_popularity}, diffusion-based methods have emerged as the dominant paradigm, reflecting the recent surge in interest in diffusion models for generative tasks. This is followed by discrete autoregressive-based approaches, which have gained popularity through their success in language generative models. %which have been a cornerstone of motion generation research.
\begin{figure}[htbp]
\centering
\includegraphics[width=0.45\textwidth]{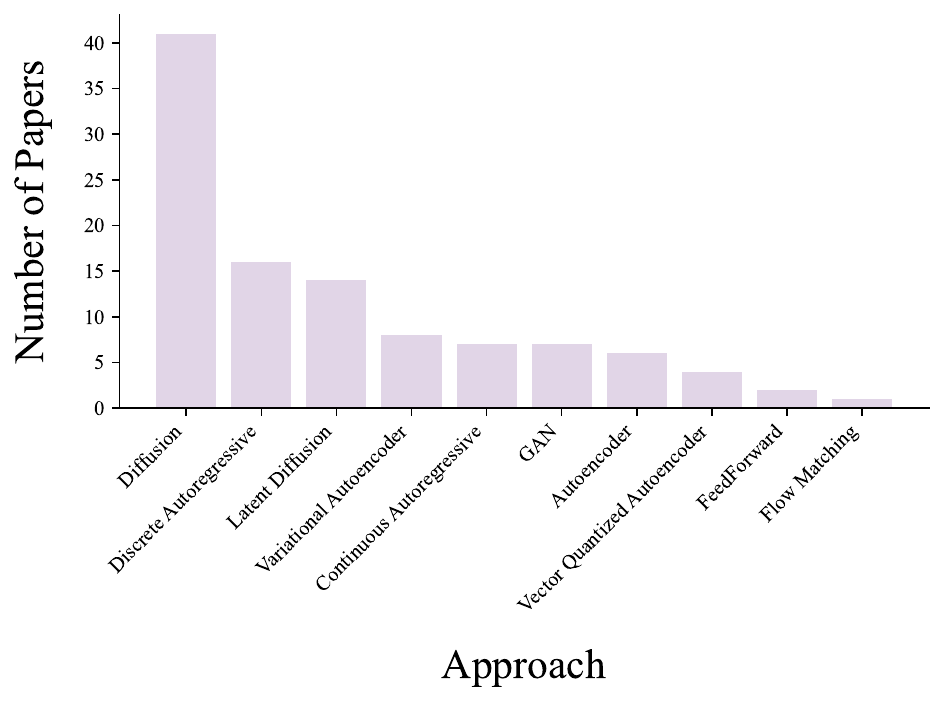}
\caption{Distribution of research approaches in motion generation literature. Diffusion-based methods dominate the field, followed by discrete Autoregressive-based and latent diffusion approaches, reflecting recent methodological trends in generative modeling.}
\label{fig:approach_popularity}
\end{figure}

\subsection{\textbf{Architectural Preferences}}

The examination of neural network architectures employed in motion generation research demonstrates clear preferences within the community (Fig.~\ref{fig:architecture_usage}). Attention-based architectures dominate the landscape, reflecting the success of transformer models in capturing long-range dependencies in sequential motion data. Convolutional architectures maintain significant usage, particularly in approaches that treat motion as spatial-temporal data, while RNN-based methods, though less prevalent, continue to be employed for their natural fit to sequential modeling tasks.
\begin{figure}[htbp]
\centering
\includegraphics[width=0.45\textwidth]{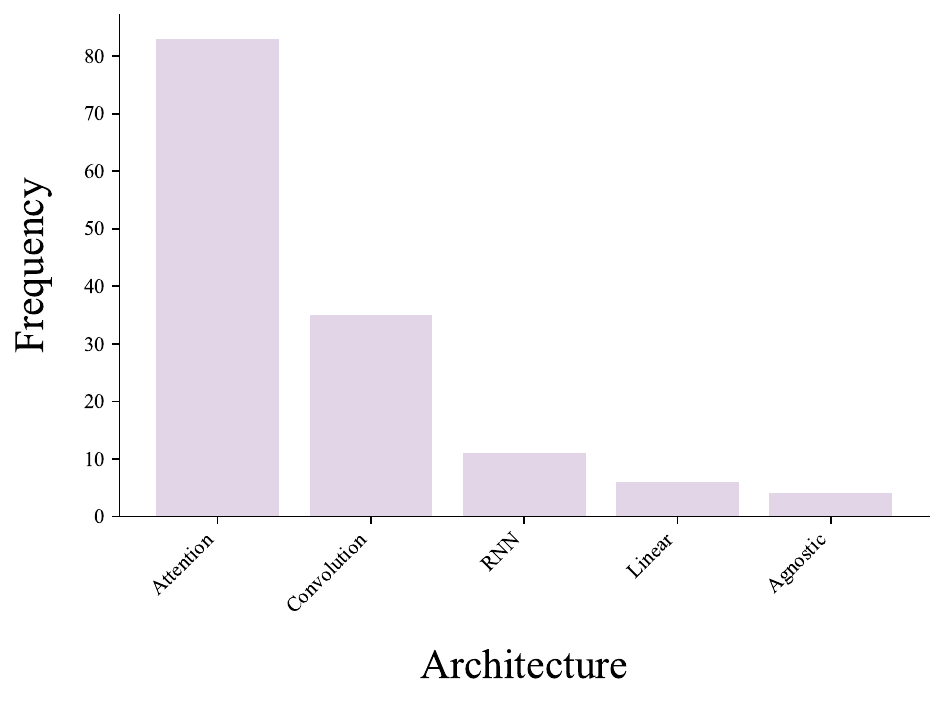}
\caption{Frequency of neural network architectures employed in motion generation research. Attention-based architectures are most prevalent, demonstrating the influence of transformer models on the field.}
\label{fig:architecture_usage}
\end{figure}

\subsection{\textbf{Dataset Landscape}}

The analysis of dataset usage patterns (Fig.~\ref{fig:dataset_usage}) reveals the established benchmarks that drive research validation in the motion generation community. HumanML3D~\citep{guo2022generating} and KIT-ML~\citep{kit-ml} emerge as the most frequently used datasets, establishing themselves as de facto standards for human motion generation research. The presence of specialized datasets such as AIST++~\cite{aist++} for dance motion and various CMU motion capture~\citep{cmu} datasets demonstrates the diversity of motion types being studied.
\begin{figure}[htbp]
\centering
\includegraphics[width=0.45\textwidth]{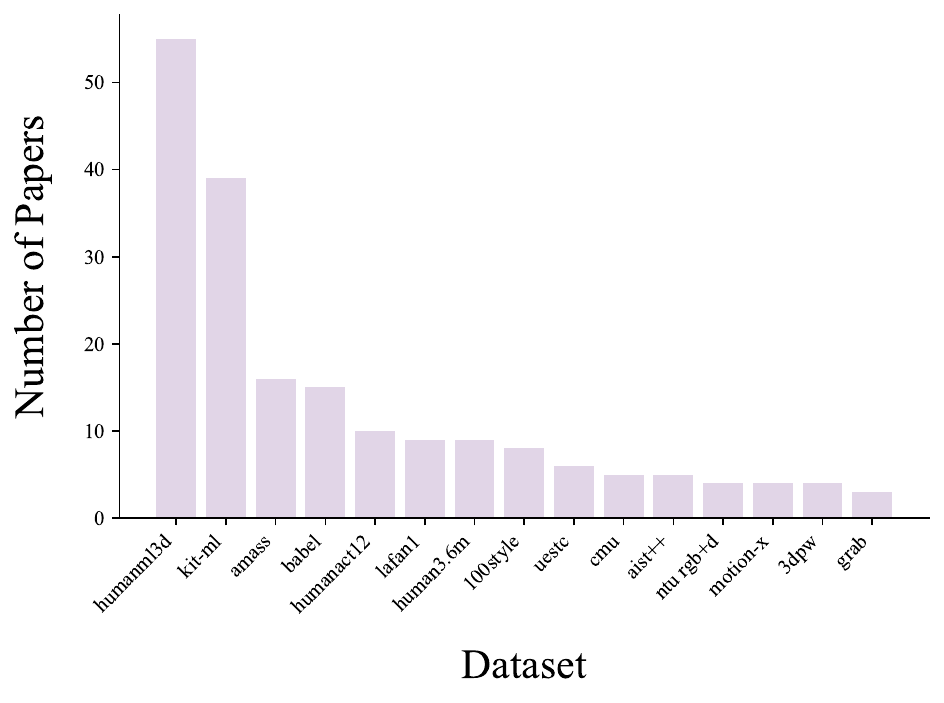}
\caption{Usage frequency of the top 15 datasets in motion generation research. HumanML3D and KIT-ML emerge as the most commonly used benchmarks, establishing themselves as community standards.}
\label{fig:dataset_usage}
\end{figure}

\subsection{\textbf{Conditioning Strategies}}

Input conditioning is a critical component of motion generation systems, as it determines how external control signals guide the generation process. Figure~\ref{fig:input_conditioning} illustrates the prevalence of different conditioning modalities, with text-based conditioning being the most common, followed by pose- and action-based conditioning. Here, pose may refer to previous motion, a target frame, or frames used for in-betweening. This distribution reflects the growing interest in natural language interfaces for motion control.
\begin{figure}[htbp]
\centering
\includegraphics[width=0.45\textwidth]{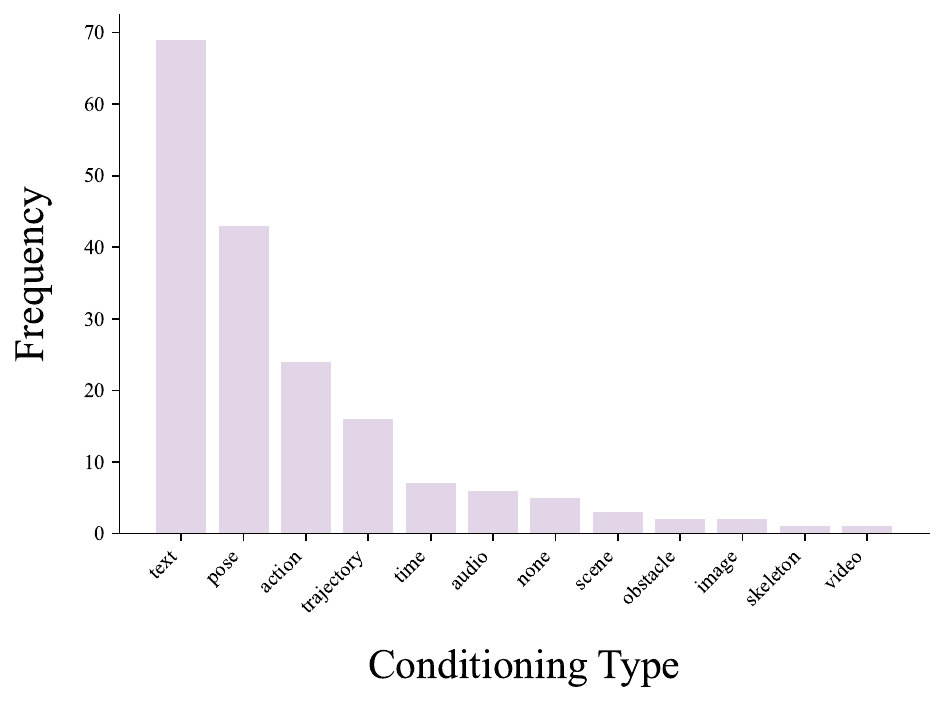}
\caption{Distribution of input conditioning modalities in motion generation systems. Text-based conditioning leads the field, followed by pose and action-based conditioning strategies.}
\label{fig:input_conditioning}
\end{figure}

\subsection{\textbf{Methodological Relationships}}

The relationship between research approaches and architectural choices reveals interesting patterns in the field's methodological landscape. Fig.~\ref{fig:approach_architecture_heatmap} demonstrates that certain approach-architecture combinations are more prevalent, suggesting established best practices and successful paradigms. For instance, diffusion methods show a strong association with attention-based architectures, while GAN approaches maintain diversity across different architectural choices.
\begin{figure}[htbp]
\centering
\includegraphics[width=0.45\textwidth]{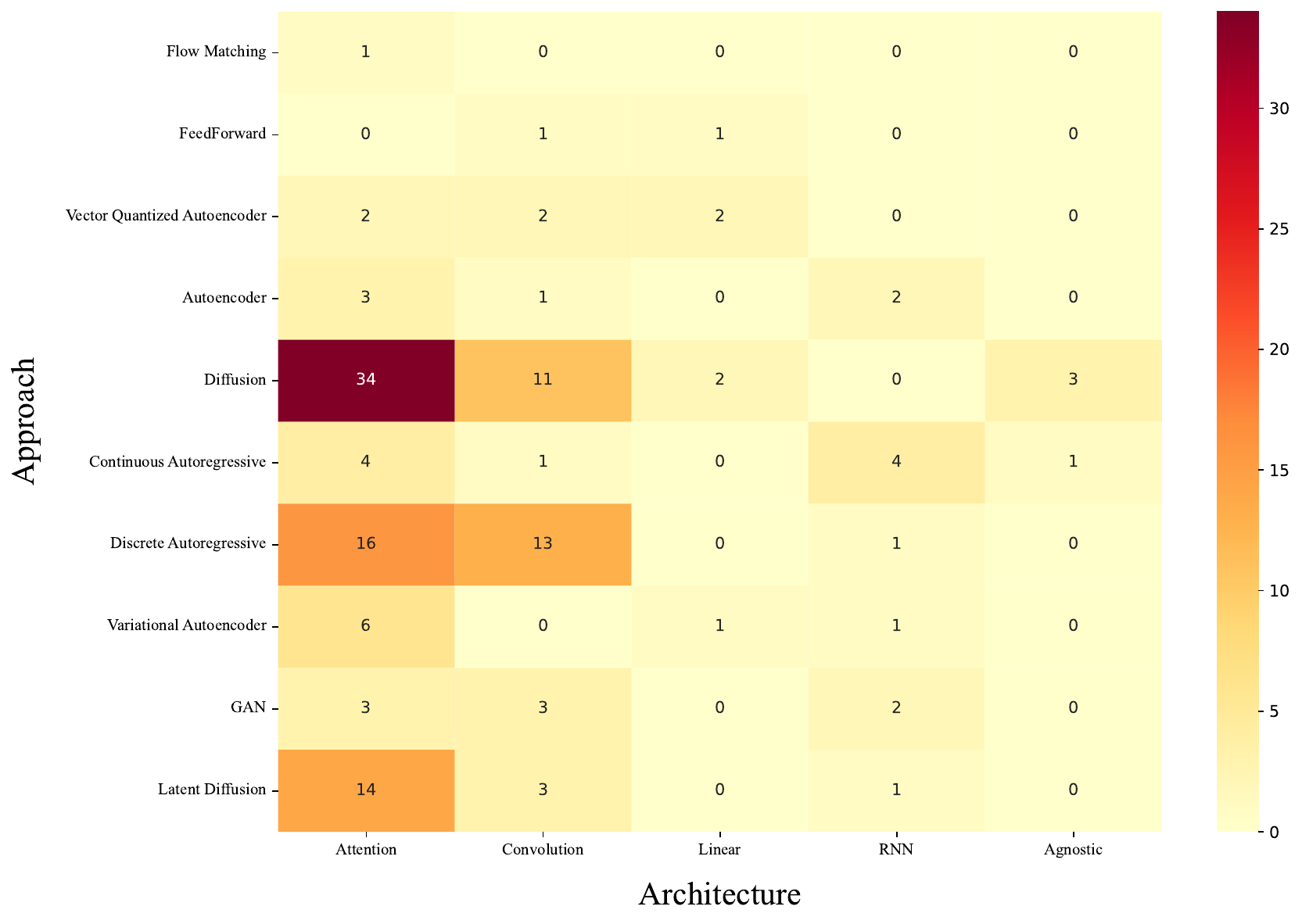}
\caption{Heatmap showing the relationship between research approaches and neural network architectures. The visualization reveals preferred approach-architecture combinations and methodological patterns in the field.}
\label{fig:approach_architecture_heatmap}
\end{figure}

Similarly, the approach-dataset relationships (Fig.~\ref{fig:approach_dataset_heatmap}) indicate preferences for certain evaluation protocols within different methodological frameworks. The concentration of activity around specific dataset-approach combinations suggests the emergence of standardized evaluation practices, which are crucial for fair comparison and reproducible research.
\begin{figure}[htbp]
\centering
\includegraphics[width=0.45\textwidth]{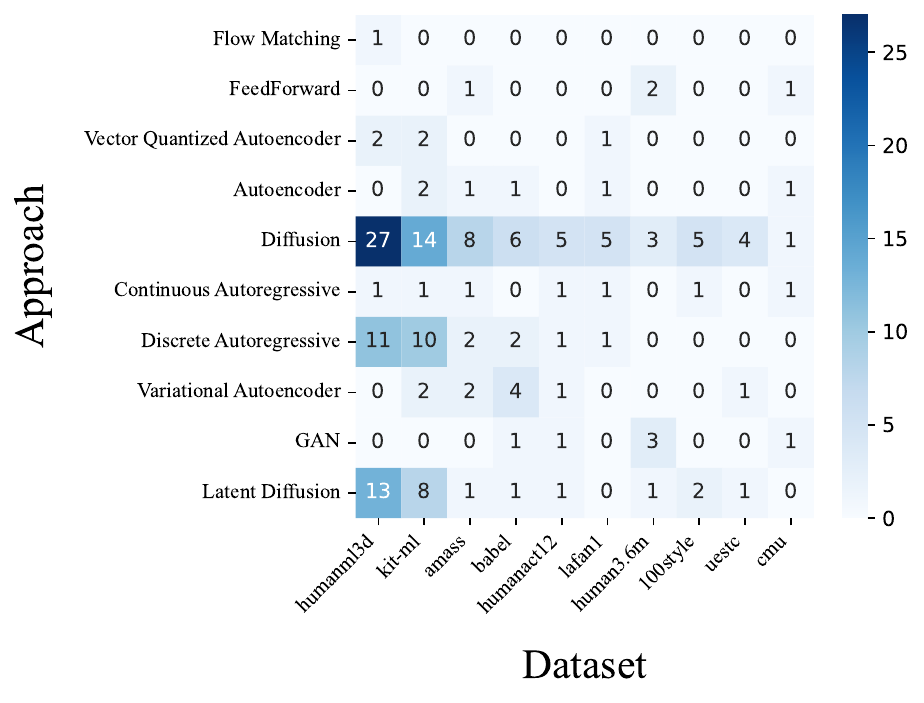}
\caption{Heatmap illustrating the relationship between research approaches and dataset usage for the top 10 most popular datasets. The pattern shows established evaluation practices within different methodological frameworks.}
\label{fig:approach_dataset_heatmap}
\end{figure}

The conditioning strategy analysis by approach (Fig.~\ref{fig:conditioning_approach_heatmap}) reveals how different methodological paradigms align with various control modalities. This relationship is particularly important for understanding the practical applications and use cases that different approaches are designed to address.
\begin{figure}[htbp]
\centering
\includegraphics[width=0.45\textwidth]{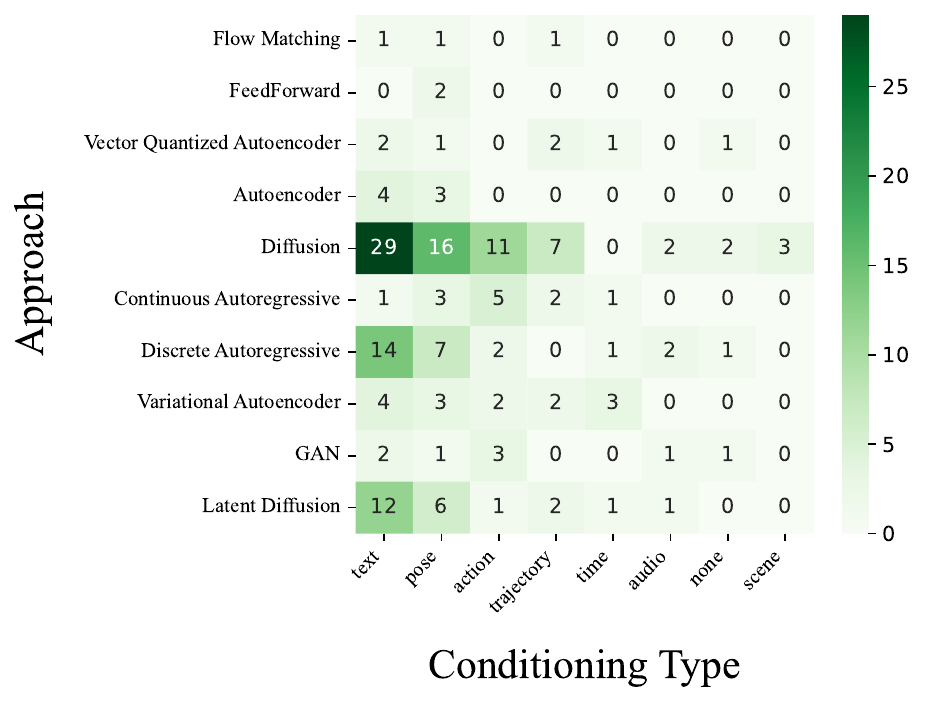}
\caption{Heatmap depicting the relationship between research approaches and conditioning strategies. The visualization shows how different methodological paradigms align with various control modalities.}
\label{fig:conditioning_approach_heatmap}
\end{figure}

\subsection{\textbf{Implications for Future Research}}

These statistical insights reveal several important trends that have implications for future research directions. The dominance of attention-based architectures and diffusion methods suggests a convergence toward approaches that can effectively model complex dependencies in motion data. The prevalence of text-based conditioning indicates the growing importance of natural language interfaces in motion generation systems.

The concentration of research around specific datasets, while beneficial for standardized comparison, also highlights the need for more diverse evaluation benchmarks to ensure broader applicability of proposed methods. Furthermore, the observed patterns in approach-architecture combinations suggest opportunities for exploring underrepresented methodological directions that might yield novel insights.

These findings provide a quantitative foundation for understanding the current state of motion generation research and can guide future investigations toward addressing gaps and exploring emerging opportunities in the field.
\section{\textbf{Conclusion}}
\label{sec:conclusion}

This survey provides a comprehensive review of recent advances in motion generation, with a particular focus on categorizing methods based on their underlying generative approaches. We systematically examine a wide range of models including autoregressive networks, autoencoders, variational autoencoders, GANs, diffusion-based models, and others, highlighting their architectural designs, conditioning mechanisms, and application scenarios. In addition, we compile a detailed overview of the evaluation metrics and datasets commonly used in the literature, enabling fairer comparison and supporting reproducibility.

Our survey primarily focuses on papers published in top-tier conferences since 2023, thereby reflecting the most up-to-date trends and innovations in the field. 
% \textcolor{red}{Through our analysis, we observe that the choice of generation strategy plays a significant role in determining the quality, efficiency, and controllability of the synthesized motion.}
In our review, we identified several common challenges across papers, including issues with physical plausibility (e.g., foot sliding and foot penetration), limitations to skeletal and human motion data, generalization, and evaluation standardization~\citep{tevethuman, zhao2024dartcontrol, li2024lamp, guo2023back, ahuja2019language2pose, petrovich2021action, wan2024tlcontrol, zhang2018mode, ahn2018text2action, zhou2023ude, cervantes2022implicit}.
We also noticed emerging directions such as real-time generation, multi-modal conditioning, and physics-informed modeling.

We hope this survey serves as a useful reference for both newcomers and experienced researchers by offering a structured and timely overview of the evolving landscape of motion generation. By consolidating recent developments and highlighting open problems, we aim to facilitate future research and guide the design of more effective and generalizable motion generation systems.

% \clearpage
\bibliographystyle{plain}  
\bibliography{bibliography}

\end{document}